\RequirePackage{fix-cm} 

\documentclass[twocolumn]{svjour3}          

\smartqed  

\usepackage{tikz}

\usepackage[round, sort, numbers, authoryear]{natbib}

\usepackage[switch]{lineno}

\usepackage{amsmath,amssymb}
\usepackage{color}
\usepackage{hyperref}
\usepackage{url}
\usepackage{breakurl}
\usepackage{rotating}
\usepackage{multirow}
\usepackage[lofdepth,lotdepth,caption=false]{subfig}
\definecolor{Gray}{gray}{0.8}
\usepackage{hhline}
\usepackage{colortbl}
\usepackage{footmisc}
\usepackage{comment}

\usepackage[ruled,vlined]{algorithm2e}

\makeatletter
\def\BState{\State\hskip-\ALG@thistlm}
\makeatother

\newcommand{\eg}{e.g.\ }
\newcommand{\ie}{i.e.\ }

\newcommand{\vertex}{{\bf v}}
\newcommand{\xClPoint}{{X}}
\newcommand{\xPrPoint}{{x}}
\newcommand{\ob}{{\bf o}}
\newcommand{\normal}{{\bf n}}
\newcommand{\ld}{{\bf d}}
\newcommand{\lm}{{\bf m}}
\newcommand{\para}{{\theta}}
\DeclareMathOperator*{\argmin}{arg\,\min}
\newcommand{\pluecker}{Pl{\"u}cker }

\newcommand{\twist}{\hat{\xi}}
\newcommand{\fingertip}{\phi}

\newcommand{\detectionIMG}{\delta}
\newcommand{\detectionPCL}{\delta'}

\newcommand{\w}{\omega}

\newcommand{\reviewChangeA}[1]{{\color{black}#1}}
\newcommand{\reviewChangeB}[1]{\textcolor{black}{#1}}

\begin{document}

\title{Capturing Hands in Action using Discriminative Salient Points and Physics Simulation}

\titlerunning{Capturing Hands in Action using Discriminative Salient Points and Physics Simulation}

\author{Dimitrios Tzionas \and Luca Ballan \and Abhilash Srikantha \and Pablo Aponte \and Marc Pollefeys \and Juergen Gall 
}

\authorrunning{D. Tzionas \and L. Ballan \and A. Srikantha \and P. Aponte \and M. Pollefeys \and J. Gall } 

\institute{D. Tzionas \and A. Srikantha \and P. Aponte \and J. Gall \at
			Institute of Computer Science III,
			University of Bonn \\
			R{\"o}merstra{\ss}e 164, 53117 Bonn, Germany \\
			\email{\{tzionas,srikanth,aponte,gall\}@iai.uni-bonn.de}
		\and
			D. Tzionas \and A. Srikantha \at
			Perceiving Systems Department \\
			Max Planck institute for Intelligent Systems \\
			Spemannstra{\ss}e 41, 72076 T{\"u}bingen, Germany 
		\and
			L. Ballan 
			\and M. Pollefeys \at
			Institute for Visual Computing,
			ETH Zurich  \\
			Universit{\"a}tstra{\ss}e 6, CH-8092 Zurich, Switzerland \\
			\email{\{luca.ballan,marc.pollefeys\}@inf.ethz.ch}
}

\date{Received: date / Accepted: date}

\maketitle

\begin{abstract}
Hand motion capture is a popular research field, recently gaining more attention due to the ubiquity of RGB-D sensors. 
However, even most recent approaches focus on the case of a single isolated hand. 
In this work, we focus on hands that interact with other hands or objects and present a framework that successfully captures motion in such interaction scenarios for both rigid and articulated objects. 
Our framework combines a generative model with discriminatively trained salient points to achieve a low tracking error and with collision detection and physics simulation to achieve physically plausible estimates even in case of occlusions and missing visual data. 
Since all components are unified in a single objective function which is almost everywhere differentiable, it can be optimized with standard optimization techniques. 
Our approach works for monocular RGB-D sequences as well as setups with multiple synchronized RGB cameras. 
For a qualitative and quantitative evaluation, we captured 29 sequences with a large variety of interactions and up to 150 degrees of freedom. 

\end{abstract}


\section{Introduction}\label{sec:introduction}

	Capturing 3d motion of human hands is an important research topic in computer vision since decades \citep{HoggHand96,Review_Erol_HandPose} due to its importance for numerous applications including, but not limited to, 
	computer graphics, animation, human computer interaction, rehabilitation and robotics. With recent technology advancements of consumer RGB-D sensors, the research interest in this topic has increased in the last few years~\citep{survey13,NYU_tracker_tompson14tog}. 	
	Despite being a special instance of full human body tracking, it can not be easily solved by applying 
	known techniques for human pose estimation like \citep{kinect_Paper} to human hands. While hands share some challenges with the full body like the high dimensionality of the underlying skeleton, they introduce additional difficulties. 
	The body parts of the hands are very similar in shape and appearance, palm and forearm are difficult to model, and severe self-occlusions are a frequent phenomenon.        
	
	Due to these difficulties, the research from the first efforts in the field \citep{HoggHand96} 
	even until very recent approaches \citep{NYU_tracker_tompson14tog} has mainly focused on a single isolated hand. While isolated hands are useful for a few applications like gesture control, humans use hands mainly for interacting with the environment and manipulating objects. In this work, we focus therefore on hands in action, \ie hands that interact with other hands or objects. This problem has been addressed so far only by a few works~\citep{Hamer_Hand_Manipulating,Hamer_ObjectPrior,Oikonomidis_1hand_object,Oikonomidis_2hands,Oikonomidis14,kyriazis2013,kyriazis2014}. While \cite{Hamer_Hand_Manipulating} considered objects only as occluders, \cite{Hamer_ObjectPrior} derive a pose prior from the manipulated objects to support the hand tracking. This approach, however, assumes that training data is available to learn the prior. A different approach to model interactions between objects and hands is based on a collision or physical model. Within a particle swarm optimization (PSO) framework, \cite{
	Oikonomidis_1hand_object,Oikonomidis_2hands} approximate the hand by spheres to detect and avoid collisions. In the same framework, \cite{kyriazis2013,kyriazis2014} enrich the set of particles by using a physical simulation for hypothesizing the state of one or several rigid objects.        
	
	Instead of employing a sampling based optimization approach like PSO, we propose in this work a single objective function that combines data terms, which align the model with the observed data, with a collision and physical model. The advantage of our objective function is that it can be optimized with standard optimization techniques. In our experiments, we use local optimization and enrich the objective function with discriminatively learned salient points to avoid pose estimation errors due to local minima. Salient points, like finger tips, have been used in the earlier work of~\cite{Kanade94}. 
	Differently from their scenario, however, 
	these salient points cannot be tracked continuously due to the huge amount of occlusions and the similarity in appearance of these features.
	Therefore we cannot rely on having a fixed association between the salient points and the respective fingers.
	To cope with this, we
	propose a novel approach that solves the salient point association jointly with the hand pose estimation problem.

	Preliminary versions of this paper appeared in \citep{LucaHands,GCPR_2014_Tzionas_Gall}. The present work unifies the pose estimation for multiple synchronized RGB cameras \citep{LucaHands} and a monocular RGB-D camera~\citep{GCPR_2014_Tzionas_Gall}. In addition, the objective function is extended by a physical model that increases the realism and physical plausibility of the hand poses. 
	In the experiments, we qualitatively and quantitatively evaluate our approach on 29 sequences and
	present for the first time successful tracking results of two hands strongly interacting with non-rigid objects.

	\begin{figure*}[t]
	\captionsetup[subfigure]{labelformat=empty}
	\centering
		\subfloat[subfigure 1 CompFORTH][]{	\includegraphics[width=0.15 \textwidth]{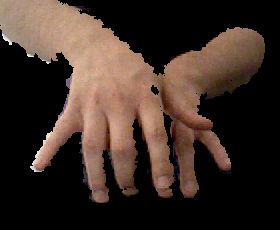}	\label{fig:muResult_102_rgbd}	}	\hspace*{-3.0mm}
		\subfloat[subfigure 2 CompFORTH][]{	\includegraphics[width=0.15 \textwidth]{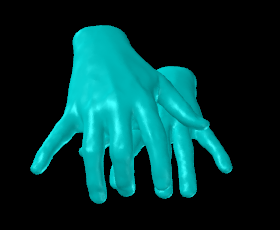}	\label{fig:muResult_102_synth}	}		
		\subfloat[subfigure 3 CompFORTH][]{	\includegraphics[width=0.15 \textwidth]{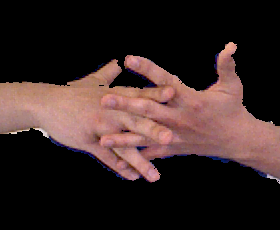}	\label{fig:muResult_109_rgbd}	}	\hspace*{-3.0mm}
		\subfloat[subfigure 4 CompFORTH][]{	\includegraphics[width=0.15 \textwidth]{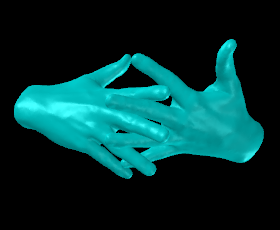}	\label{fig:muResult_109_synth}	}		
		\subfloat[subfigure 5 CompFORTH][]{	\includegraphics[width=0.15 \textwidth]{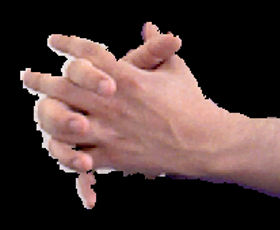}	\label{fig:muResult_122_rgbd}	}	\hspace*{-3.0mm}
		\subfloat[subfigure 6 CompFORTH][]{	\includegraphics[width=0.15 \textwidth]{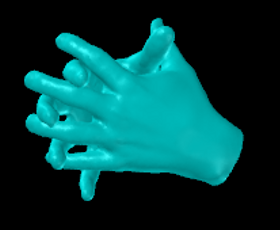}	\label{fig:muResult_122_synth}	}
		\vspace*{-4mm}
	\caption{	Qualitative results of our approach for the case of hand-hand interaction. 
			Each pair shows the aligned RGB and depth input images after depth thresholding along with the pose estimate
	}
	\label{fig:resultsQualitative}
	\end{figure*}


\section{Related Work}\label{sec:relatedWork}
	
	The study of hand motion tracking has its roots in the $90$s \citep{Kanade94,Kanade95}. 
	Although the problem can be simplified by means of data-gloves~\citep{Kragic_dataglove}, color-gloves \citep{glove_MIT}, markers \citep{colorGlove} or wearable sensors \citep{Oikonomidis_MSR_Digits}, 
	the ideal solution pursued is the unintrusive, marker-less capture of hand motion.

	In pursuit of this, one of the first hand tracking approaches \citep{Kanade94} introduced the use of local optimization in the field.
	Several filtering approaches have been presented \citep{Isard2000,Cipolla_ModelBased,Huang_capturingNaturalHandArtic2001,vanGool_smartParticle}, 
	while also belief-propagation proved to be suitable for articulated objects \citep{Hamer_ObjectPrior,Hamer_Hand_Manipulating,SudderthNonparamBeliefPropag}.
	\citet{OikonomidisBMVC} employ Particle Swarm Optimization (PSO) as a form of stochastic search, while later they present a novel evolutionary algorithm that capitalizes on quasi-random sampling \citep{Oikonomidis14}.
	\citet{Oikonomidis_MSR_Digits} and \citet{glove_MIT} use inverse-kinematics, 
	while \citet{HoggHand96} and \citet{Huang_capturingNaturalHandArtic2001} reduce the search space using linear subspaces.
	\citet{AthitsosCluttered2003} resort to probabilistic line matching, 
	while \citet{Cipolla_TORR} combine Bayesian filtering with Chamfer matching. 
	Recently, \citet{DART-Schmidt-RSS-14} extended the popular signed distance function (SDF) representation to articulated objects, 
	while \citet{MSR_ASIA_handTracking} combine a gradient based ICP approach with PSO, showing the complementary nature of the two approaches. 
	\reviewChangeB{
			\citet{srinath_iccv2013} explore the use of a Sum of Gaussians (SoG) model for hand tracking on RGB images, 
			which is later replaced by a Sum of Anisotropic Gaussians \citep{srinath_3dv2014}. 
	}

	All these approaches have in common that they are generative models. They use an explicit model to generate pose hypotheses, which are evaluated against the observed data. 
	The evaluation is based on an objective function which implicitly measures the likelihood
	by computing the discrepancy between the pose estimate (hypothesis) and the observed data in terms of an error metric.
	To keep the problem tractable, 
	each iteration is initialized by the pose estimate of the previous step, relying thus heavily on temporal continuity and being prone to accumulative error.
	The objective function is evaluated in the high-dimensional, continuous parameter space. 
	Recent approaches relax the assumption of a fixed predefined shape model, allowing for online non-rigid shape deformation \citep{MSR_handShapeAdaptation} 
	that enables 
	better data fitting 
	and 
	user-specific adaptation. 
	
	Discriminative methods learn a direct mapping from the observed image features to the discrete \citep{AthitsosCluttered2003,Romero09,JavierHandsInAction} or continuous \citep{Murray_handRegression2006,Athitsos_HandSpecializedMappings2001} target parameter space. 
	Some approaches also segment the body parts first and estimate the pose in a second step~\citep{NYU_tracker_tompson14tog,KeskinECCV12}. 
	Most methods operate on a single frame, 
	being thus immune to pose-drifting due to error accumulation.
	Generalization in terms of capturing illumination, articulation and view-point variation can be realized only through adequate representative training data. 
	Acquisition and annotation of realistic training data 
	is though a cumbersome and costly procedure.
	For this reason most approaches rely on synthetic rendered data \citep{KeskinECCV12,JavierHandsInAction} that has inherent ground-truth. 
	Special care is needed to avoid over-fitting to the training set, while the discrepancy between realistic and synthetic data is an important limiting factor. 
	Recent approaches \citep{TKKIM_ICCV13_Real_time_Articulated_Hand} tried to address the latter using transductive regression forests to 
	transfer knowledge from fully labeled synthetic 
	data to partially labeled realistic data. 
	Finally, the accuracy of discriminative methods heavily depends on the invariance, repeatability and discriminative properties of the features employed 
	and is lower in comparison to generative methods.
	
	A discriminative method can effectively complement a generative method, either 
	in terms of initialization or recovery, 
	driving the optimization framework away from local minima in the search space and aiding convergence to the global minimum. 
	\citet{srinath_iccv2013} combine in a real time system a Sum of Gaussians (SoG) generative model with a discriminatively trained fingertip detector in depth images using a linear SVM classifier. 
	\reviewChangeB{
	Alternatively, the model can also be combined with a part classifier based on random forests \citep{srinath_cvpr2015}.
	Recently, \citet{sharp_chi2015} combined a PSO optimizer with a robust, two-stage regression re-initializer that predicts a distribution over hand poses from a single RGB-D frame. 
	}
	
	Generative and discriminative methods have used various low level image cues for hand tracking that are often combined, namely silhouettes, edges, shading, color, optical flow \citep{metaxasSamarasHand_multipleCues,ParagiosHandMonocular2011},
	while depth \citep{Faugeras_PhysicalForces,vanGool_smartParticle,Hamer_Hand_Manipulating} has recently gained popularity with the ubiquity of 
	RGB-D sensors \citep{OikonomidisBMVC,Oikonomidis_2hands,srinath_iccv2013,NYU_tracker_tompson14tog,DART-Schmidt-RSS-14,MSR_ASIA_handTracking}.
	In this work, we combine in a single framework a generative model 
	with discriminative salient points detected by a Hough forest \citep{juergen_Hough}, \ie a finger nail detector on color images and finger tip detector on depth images, respectively. 
	As low level cues, we use edges and optical flow for the RGB sequences and depth for the RGB-D sequences.


\section{Pose Estimation}\label{sec:trackingMethod}
	
	\begin{sloppypar}
	Our approach for capturing the motion of hands and manipulated objects can be applied to RGB-D and multi-view RGB sequences. In both cases hands and objects are modeled in the same way as described in Section~\ref{sec:handModel}.	
	The main difference between RGB-D and RGB sequences is the used data term, which depends on depth or edges and optical flow, respectively. We therefore introduce first the objective function for a monocular RGB-D sequence in Section~\ref{sec:optimization} and describe the differences for RGB sequences in Section~\ref{sec:Differences_MultiRGB_2_SingRGBD}.
	\end{sloppypar}
	
		\subsection{Hand and Object Models}\label{sec:handModel}\label{sec:poseParameterization}
 
\begin{figure}[t]
		\captionsetup[subfigure]{}
		\centering
			\subfloat[subfigure 1 model][]	{	\includegraphics[width=0.173 \textwidth]{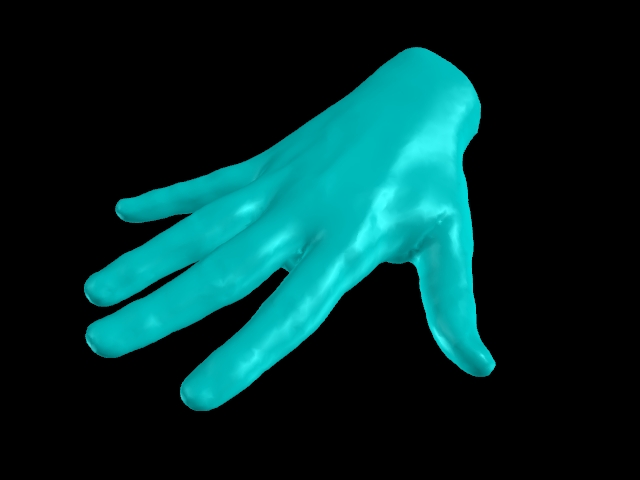}		\label{fig:model_mesh}		}		\hspace*{-2.1em}
			\subfloat[subfigure 2 model][]	{	\includegraphics[width=0.173 \textwidth]{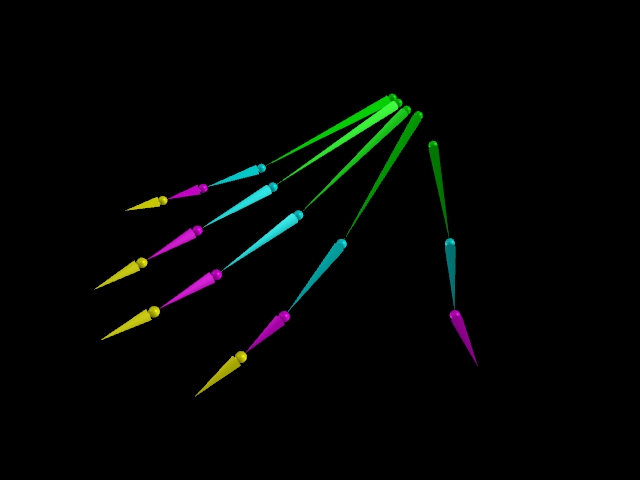}		\label{fig:model_skeleton}	}		\hspace*{-2.1em}
			\subfloat[subfigure 3 model][]	{	\includegraphics[width=0.173 \textwidth]{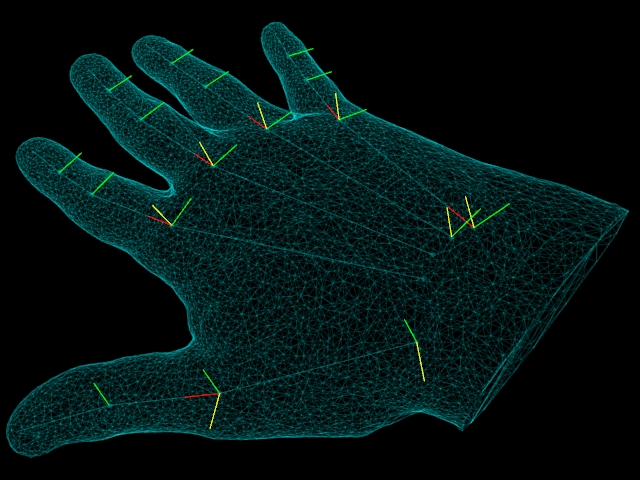}		\label{fig:model_DOFs}		}	\\	\vspace*{-3.1mm}
			\subfloat[subfigure 1 VOI][]	{	\includegraphics[width=0.173 \textwidth]{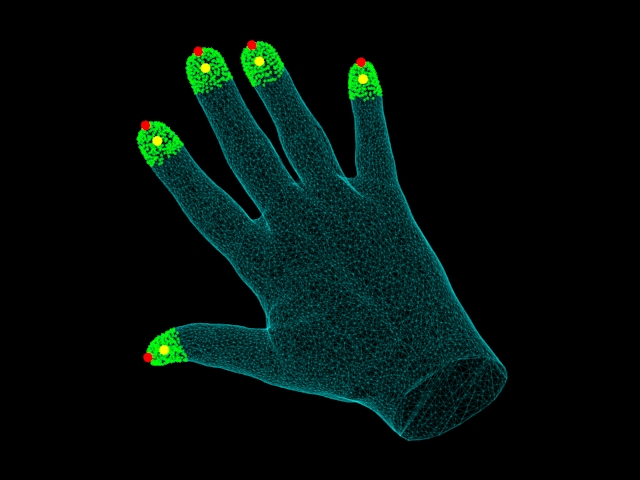}			\label{fig:VOI_zoomOut}		}		\hspace*{-2.1em}
			\subfloat[subfigure 2 VOI][]	{	\includegraphics[width=0.173 \textwidth]{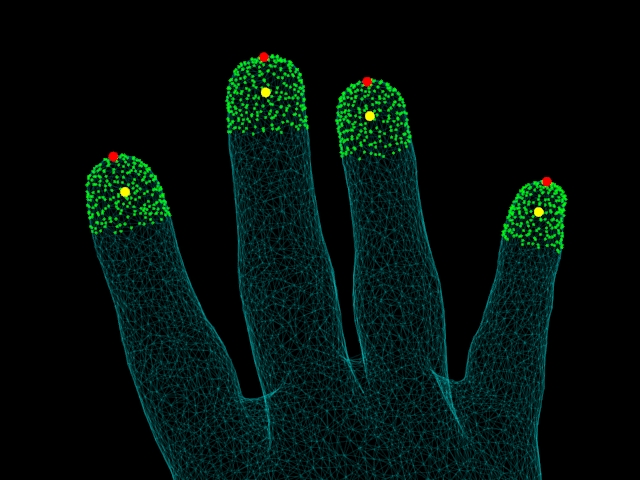}		\label{fig:VOI_zoomInn}		}		\hspace*{-2.1em}
			\subfloat[subfigure 3 VOI][]	{	\includegraphics[width=0.173 \textwidth]{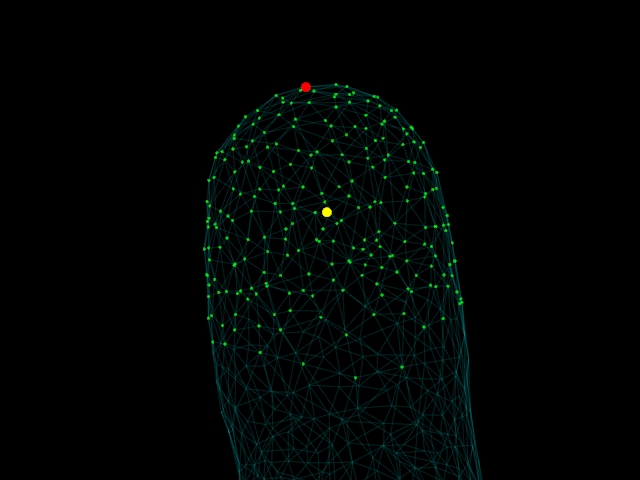}		\label{fig:VOI_zoomINN}		}
			\vspace*{-1mm}
		\caption{
			Hand model used for tracking. 
			(a) 	Mesh 
			(b) 	Kinematic Skeleton 
			(c) 	Degrees of Freedom (DoF)
			(d-f) 	Mesh fingertips (green) used for the salient point detector. 
			    	The vertices of the fingertips are found based on the manually annotated red vertices. 
			    	The centroid of the fingertips, as defined in Section~\ref{subsec:salientPointDetector}, is depicted with yellow color 
		}
		\label{fig:modelsAndVOIs_HAND}
	\end{figure}

\begin{figure}[t]
		\captionsetup[subfigure]{labelformat=empty}
		\centering								
			\subfloat[subfigure 1 ballMesh][]	{	\includegraphics[trim=000mm 023mm 000mm 009mm, clip=true, width=0.173 \textwidth]{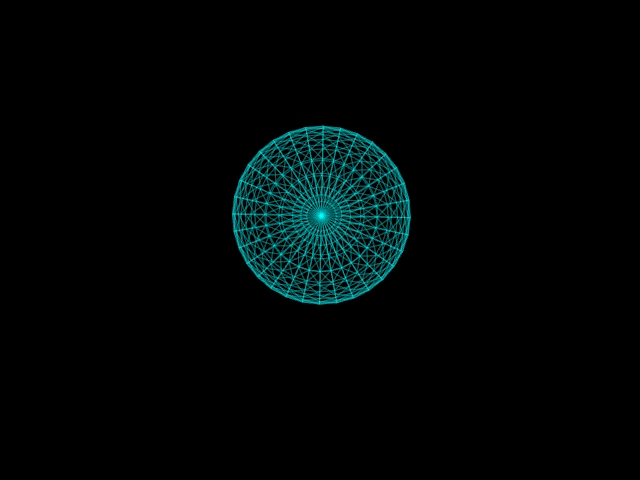}	\label{fig:model_ballMesh}	}		\hspace*{-02.4mm}
			\subfloat[subfigure 2 cubeMesh][]	{	\includegraphics[trim=000mm 021mm 000mm 011mm, clip=true, width=0.173 \textwidth]{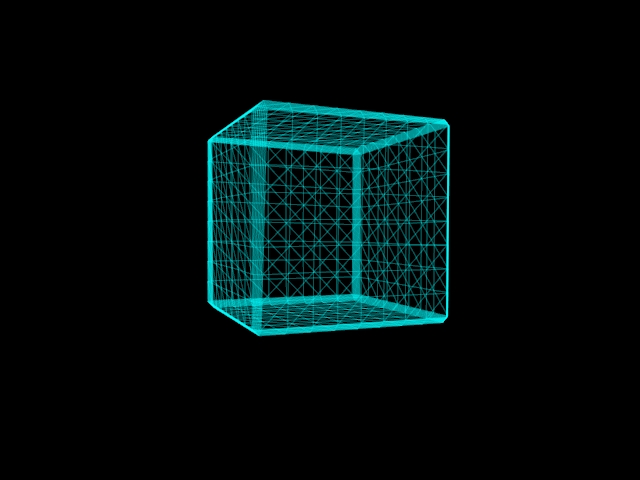}	\label{fig:model_cubeMesh}	}	\\	\vspace*{-08.0mm}
			\subfloat[subfigure 3 pipeDOF1][]	{	\includegraphics[trim=008mm 001mm 018mm 025mm, clip=true, width=0.176 \textwidth]{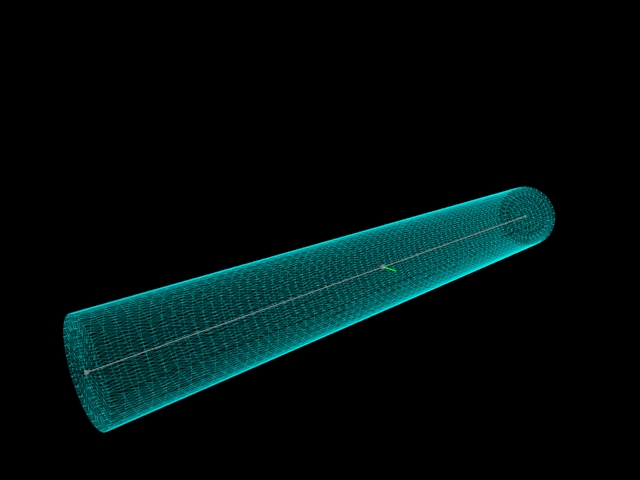}	\label{fig:model_pipeDOF1}	}		\hspace*{-03.5mm}
			\subfloat[subfigure 4 pipeDOF2][]	{	\includegraphics[trim=011mm 001mm 015mm 025mm, clip=true, width=0.176 \textwidth]{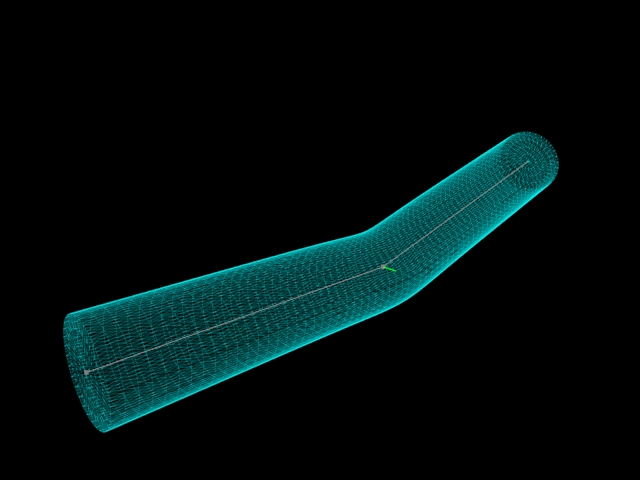}	\label{fig:model_pipeDOF2}	}	\\	\vspace*{-08.0mm}
			\subfloat[subfigure 5 ropeDOF1][]	{	\includegraphics[trim=013mm 012mm 015mm 066mm, clip=true, width=0.350 \textwidth]{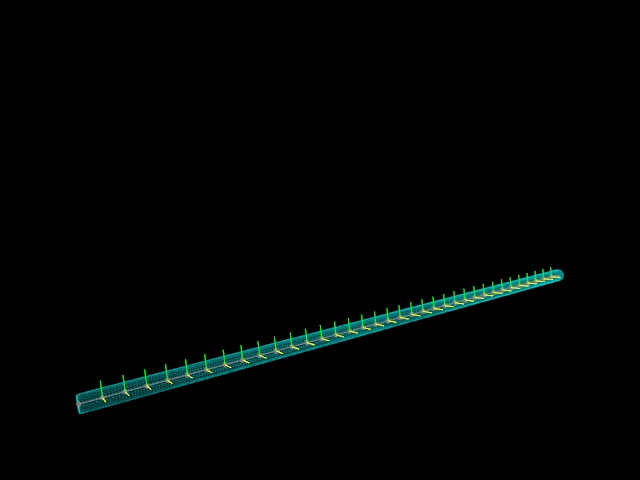}	\label{fig:model_ropeDOF1}	}	\\	\vspace*{-11.2mm}
			\subfloat[subfigure 6 ropeDOF2][]	{	\includegraphics[trim=013mm 012mm 015mm 051mm, clip=true, width=0.350 \textwidth]{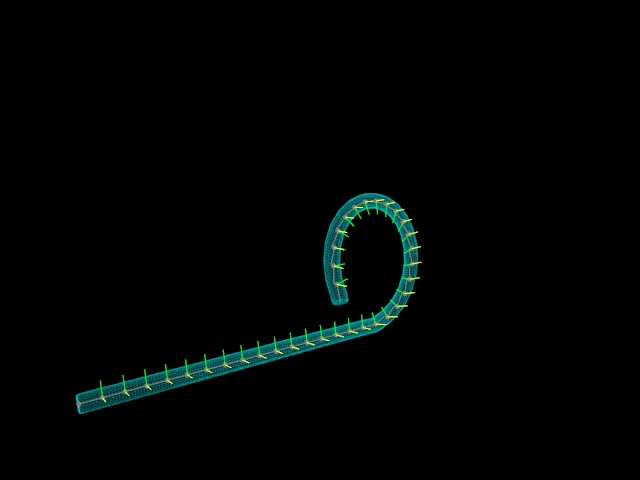}	\label{fig:VOI_ropeDOF2}	}	
			
			\vspace*{-5mm}
		\caption{
			\reviewChangeA
			{
				Object models used for tracking and their DoF: 
				(top-left)	a rigid ball with $6$ DoF;
				(top-right)	a rigid cube with $6$ DoF;
				(middle)	a pipe with $1$ revolute joint, \ie $7$ DoF; 
				(bottom)	a rope with $70$ revolute joints, \ie $76$ DoF
			}
		}
		\label{fig:modelsAndVOIs_OBJECTS}
	\end{figure}

\begin{figure}[ht]
	\vspace*{3mm}
		\begin{center}
		\tiny
		\begin{tabular}{c c c}
		\multirow{2}{*}[+12.50mm]{	\includegraphics[trim=005mm 000mm 005mm 000mm, clip=true, height=27.77mm]{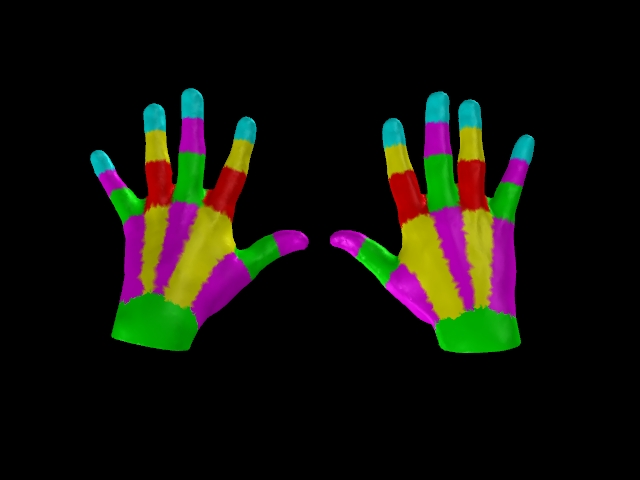}}	&	\hspace*{-06mm}		
					
						\includegraphics[trim=035mm 010mm 025mm 010mm, clip=true, height=14mm]{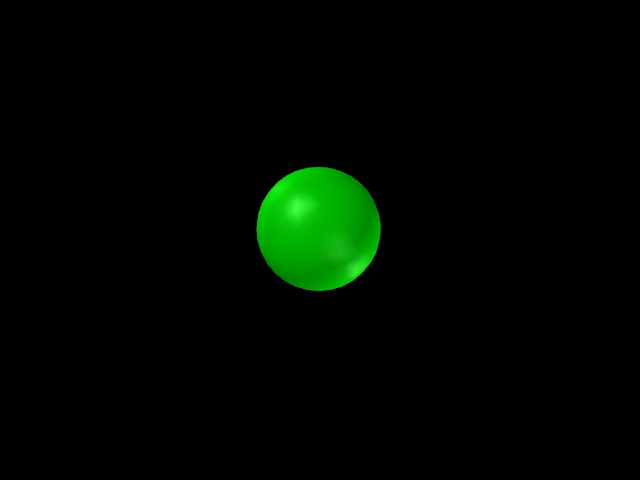}		&	\hspace*{-06mm}		
						\includegraphics[trim=000mm 031mm 000mm 023mm, clip=true, height=14mm]{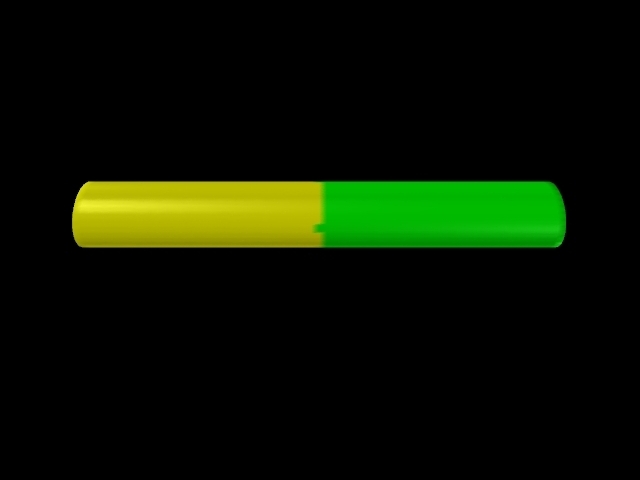}	\\		\vspace*{-03mm}		\\

						{}														&	\hspace*{-06mm}		
						\includegraphics[trim=035mm 010mm 025mm 010mm, clip=true, height=14mm]{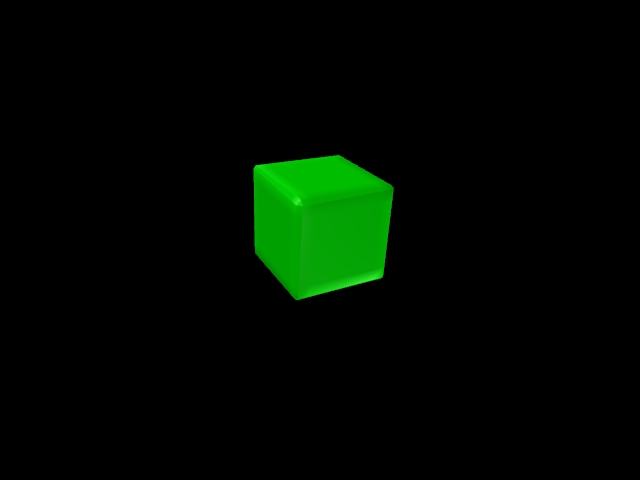}		&	\hspace*{-06mm}		
						\includegraphics[trim=000mm 029mm 000mm 025mm, clip=true, height=14mm]{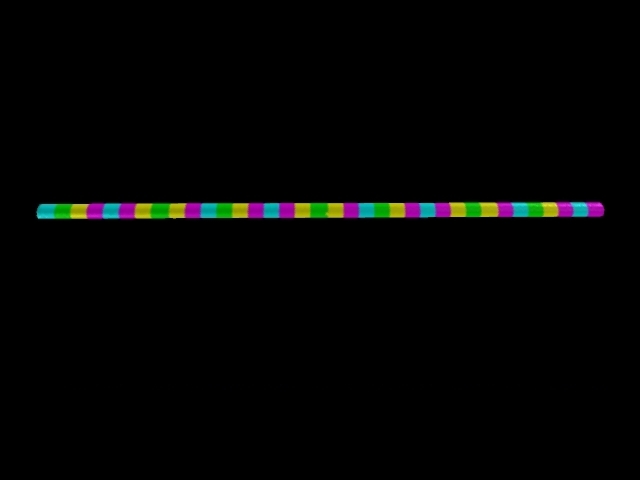}	
		\end{tabular}				  	
		\normalsize
		\caption{	Segmentation of the meshes based on the skinning weights.
				The ball and the cube are rigid objects while the pipe and rope are modeled as articulated objects. 
				Each hand has $20$ skinning bones, the pipe has $2$, while the rope has $36$ 
		}
		\label{fig:objectTemplateMeshes_skinSegmentation}
		\end{center}
	\end{figure}

		We resort to the popular linear blend skinning (LBS) model~\citep{LBS_PoseSpace}, consisting of a triangular mesh with an underlying kinematic skeleton, as depicted in Figure \ref{fig:modelsAndVOIs_HAND}a-c, and a set of skinning weights.	
		In our experiments, a 
		triangular mesh of a pair of hands
		was obtained by a 3D scanning solution, while meshes for several objects (ball, cube, pipe, rope) 
		were created manually with a 3D graphics software. \reviewChangeA{Some objects are shown in Figure~\ref{fig:modelsAndVOIs_OBJECTS}}. 
		A skeletal structure defining the kinematic chain was manually defined and fitted into the meshes. 
		The skinning weight $\reviewChangeA{\kappa}_{\vertex,j}$ defines the influence of bone $j$ on 3D vertex $\vertex$, where $\sum_j \reviewChangeA{\kappa}_{\vertex,j} = 1$. Figure~\ref{fig:objectTemplateMeshes_skinSegmentation} visualizes the mesh using the largest skinning weight for each vertex as bone association.   
		The deformation of each mesh is driven by its underlying skeleton with pose parameter vector $\para$ through the skinning weights and is expressed by the LBS operator: 
		\begin{linenomath} 
		\begin{equation}\label{eq:lbs}
		\vertex(\para) = \sum_j \reviewChangeA{\kappa}_{\vertex,j} T_j(\para) T_j(0)^{-1} \vertex(0)
		\end{equation}
		\end{linenomath} 
		where $T_j(0)$ and $\vertex(0)$ are the bone transformations and vertex positions at the known rigging pose.
		The skinning weights are computed using~\citep{pinocchio}.

	\clearpage

\reviewChangeA{
		The global rigid motion is represented by a $6$ DoF twist  
		$\vartheta\xi=\vartheta(u_{1},u_{2},u_{3},\w_{1},\w_{2},\w_{3})$ with $\Vert\w\Vert=1$~\citep{Malik_Twist,Murray_MathInRob,PonsModelBased}. 
		The twist action $\vartheta\twist \in se(3)$ has the form of a $4\times4$ matrix
		\begin{linenomath}
			\begin{equation}
				\vartheta \hat{\xi} = \vartheta  \begin{pmatrix}	 \hat{\w} 	& 	u	\\
										 0_{1\times3}	& 	0	\end{pmatrix}	= \vartheta
																	\begin{pmatrix}	 	0 	& 	-\w_3 	& 	 \w_2	&	u_1	\\
																				\w_3 	& 	 0	& 	-\w_1	&	u_2	\\
																				-\w_2	& 	 \w_1	& 	 0	&	u_3	\\
																				0	& 	 0	&	 0	&	0	\end{pmatrix} 
			\end{equation}
		\end{linenomath}
		and the exponential map operator $\exp(\vartheta \hat{\xi})$ defines the group action:
		\begin{linenomath}
			\begin{equation}
				T(\vartheta \hat{\xi}) = \begin{pmatrix}	 	R_{3\times3}	&	t_{3\times1} 	\\
											0_{1\times3}	& 	1		\end{pmatrix} = \exp(\vartheta \hat{\xi}) \in SE(3).
			\end{equation}
		\end{linenomath}
}

\reviewChangeA{
While $\para = \vartheta\xi$ for a rigid object, articulated objects have additional parameters. We model the joints by revolute joints. A joint with one DoF is modeled by a single revolute joint, \ie the transformation of the corresponding bone $j$ is given by $\exp(\vartheta_{p(j)}\xi_{p(j)})\exp(\vartheta_j\xi_j)$ where $p(j)$ denotes the parent bone. If a bone does not has a parent bone, it is the global rigid transformation. The transformation of an object with one revolute joint is thus described by $\para = (\vartheta\xi, \vartheta_1)$. Joints with two or three DoF are modeled by a combination of $K_j$ revolute joints, \ie $\prod_{k=1}^{K_j} \exp(\vartheta_{j,k}\xi_{j,k})$. For simplicity, we denote the relative transformation of a bone $j$ by  $\hat{T}_j(\para) = \prod_{k=1}^{K_j} \exp(\vartheta_{j,k}\xi_{j,k})$. The global transformation of a bone $j$ is then recursively defined by 
\begin{linenomath}
\begin{equation}\label{eq:motion_joint_j}
		T_j(\para) = T_{p(j)}(\para)\hat{T}_{j}(\para).
\end{equation}			
\end{linenomath}
}

		In our experiments, a single hand consists of $31$ revolute joints, \ie $37$ DoF, as shown in Figure \ref{fig:model_DOFs}. 
	\reviewChangeA
	{
		The rigid objects have $6$ DoF. The deformations of the non-rigid shapes shown in Figure~\ref{fig:modelsAndVOIs_OBJECTS} are approximated by a skeleton.
		The pipe has $1$ revolute joint, \ie $7$ DoF, while 
		the rope has $70$ revolute joints, \ie $76$ DoF.  		 
	}
		Thus, for sequences with two interacting hands we have to estimate all $74$ DoF and together with the rope $150$ DoF.


\subsection{Objective Function}\label{sec:optimization}
	
	\begin{sloppypar}
	\reviewChangeA
	{
	Our objective function for pose estimation consists of seven terms:
	\begin{linenomath}
	\begin{align}\label{eq:obj}
	\begin{split}
	E(\para, D) = \quad&E_{model \rightarrow data}(\para,D) + E_{data \rightarrow model}(\para,D) + 	\\  
		       &\gamma_c E_{collision}(\para) + E_{salient}(\para,D) + 				\\
		       &\gamma_{ph} E_{physics}(\para) + 							
		        \gamma_a E_{anatomy}(\para)		+						\\
		       &\gamma_r E_{regularization}(\para)
	\end{split}
	\end{align}
	\end{linenomath}
	where $\para$ are the pose parameters of the template meshes and $D$ is the current preprocessed depth image. 
	The preproccesing is explained in Section \ref{sec:Preprocessing}. 
	}
	The first two terms minimize the alignment error of the transformed mesh and the depth data. The alignment error is measured by $E_{model \rightarrow data}$, which measures how well the model fits the observed depth data, and $E_{data \rightarrow model}$, which measures how well the depth data is explained by the model. 
	$E_{salient}$ measures the consistency of the generative model with detected salient points in the image. The main purpose of the term in our framework is to recover from tracking errors of the generative model. 
	$E_{collision}$ penalizes intersections of fingers and 
	$E_{physics}$ enhances the realism of grasping poses during interaction with objects. 
	Both of the terms $E_{collision}$ and $E_{physics}$ ensure physically plausible poses and are complementary. 
	\reviewChangeA
	{The term $E_{anatomy}$ enforces amatomically inspired joint limits, while the last term is a simple regularization term that prefers the solution of the previous frame if there are insufficient oberservations to determine the pose. 
	} 
	\end{sloppypar}
	
	\reviewChangeA
	{
	In the following, we give details for the terms of the objective function \eqref{eq:obj} 
	as well as the optimization of it. 
}


\subsubsection{Preprocessing:}\label{sec:Preprocessing}

	For pose estimation, we first remove irrelevant parts of the RGB-D image by thresholding the depth values, 
	in order to avoid unnecessary processing like normal computation for points far away. 
	Segmentation of the hand from the arm is not necessary and is therefore not performed. 
	Subsequently we apply skin color segmentation on the RGB image~\citep{skinnColorGMM}. 
	As a result we get masked RGB-D images, denoted as $D$ in \eqref{eq:obj}. 	
	The skin color segmentation separates hands and non-skin colored objects, facilitating hand and object tracking accordingly.


	\subsubsection{Fitting the model to the data - $LO_{m2d}$:}\label{subsec:m2d}

	The first term in Equation \eqref{eq:obj} aims at fitting the mesh parameterized by pose parameters $\para$ to the preprocessed data $D$. 
	To this end, the depth values are converted into a 3D point cloud based on the calibration data of the sensor.	
	The point cloud is then smoothed by a bilateral filter \citep{bilateralFAST} and normals are computed \citep{normals_integralImages_Holzer}. 
	\reviewChangeA{For each visible vertex of the model $\vertex_{i}(\para)$, 
	with normal $\normal_{i}(\para)$, we search for the closest point $\xClPoint_i$ in the point cloud.} This gives a 3D-3D correspondence for each vertex. 
	We discard the correspondence if the angle between the normals of the vertex and the closest point is larger than 45$^\circ$ or the distance between the points is larger than 10 mm.  
	We can then write the term $E_{model \rightarrow data}$ as a least squared error of \emph{point-to-point} distances:
	\begin{linenomath} 
	\begin{equation}\label{eq:errorResidual_p2p}
	E_{model \rightarrow data}(\para,D) = \sum_i \Vert \vertex_{i}(\para) - \xClPoint_i \Vert^2\;
	\end{equation}
	\end{linenomath}
	An alternative to the \emph{point-to-point} distance is the \emph{point-to-plane} distance, which is commonly used for 3D reconstruction~\citep{p2pl,ICP_EfficientVariants,RusinkiewiczRealTimeINHAND}.
	\reviewChangeA
	{
		In this case: 
	}
	\begin{linenomath} 
	\begin{equation}\label{eq:errorResidual_p2pl}
	\reviewChangeA{E_{model \rightarrow data}(\para,D) = \sum_i \Vert \normal_{i}(\para)^T (\vertex_{i}(\para) - \xClPoint_i) \Vert^2}\;.
	\end{equation}
	\end{linenomath}
	The two distance metrics are evaluated in Section \ref{subsubsec:experimentDistanceMetrics}.


	\subsubsection{Fitting the data to the model - $LO_{d2m}$:}\label{subsec:d2m}
	Only fitting the model to the data is not sufficient as we will show in our experiments. 
	In particular, poses with self-occlusions can have a very low error since the measure only evaluates how well the visible part of the model fits the point cloud. 
	The second term $E_{data \rightarrow model}(\para,D)$ matches the data to the model to make sure that the solution is not degenerate and explains the data as well as possible. 
	However, matching the data to the model is more expensive since after each iteration the pose changes, which would require to update the data structure for matching, \eg distance fields or kd-trees, after each iteration.
	We therefore reduce the matching to depth discontinuities~\citep{functionalCategorization}.
	To this end, we extract depth discontinuities from the depth map and the projected depth profile of the model using an edge detector~\citep{canny}. 
	Correspondences are again established by searching for the closest points, but now in the depth image using a 2D distance transform~\citep{DT_Felzenszwalb}. 
	Similar to $E_{model \rightarrow data}(\para,D)$, we discard correspondences with a large distance. The depth values at the depth discontinuities in $D$, 
	however, are less reliable not only due to the depth ambiguities between foreground and background, but also due to the noise of consumer sensors. 
	The depth of the point in $D$ is therefore computed as average in a local $3\times3$ pixels neighborhood and the outlier distance threshold is increased to 30 mm. 
	The approximation is sufficient for discarding outliers, but insufficient for minimization. 
	For each matched point in $D$ we therefore compute the projection ray uniquely expressed as a \pluecker line~\citep{PonsModelBased,Rosenh_Plucker_IJCV,Stolfi91} 
	with direction $\ld_i$ and moment $\lm_i$ 
	and minimize the least square error between the projection ray and the vertex $\vertex_{i}(\para)$ for each correspondence:  
	\begin{linenomath}
	\begin{equation}\label{eq:errorResidual_Pluecker}
	E_{data \rightarrow model}(\para,D) = \sum_{i} \Vert \vertex_{i}(\para) \times \ld_i - \lm_i \Vert^2\;
	\end{equation}
	\end{linenomath}
	
	\reviewChangeB{We compared the matching based on depth discontinuities with a direct matching of the point cloud to the model using a kd-tree. The direct matching increases the runtime by $40\%$ or more without reducing the error. 
	}

	\begin{figure}[t]
	    \begin{center}	
	      \includegraphics[trim=0mm 0mm 0mm 0mm, clip=true,width=0.50 \textwidth]{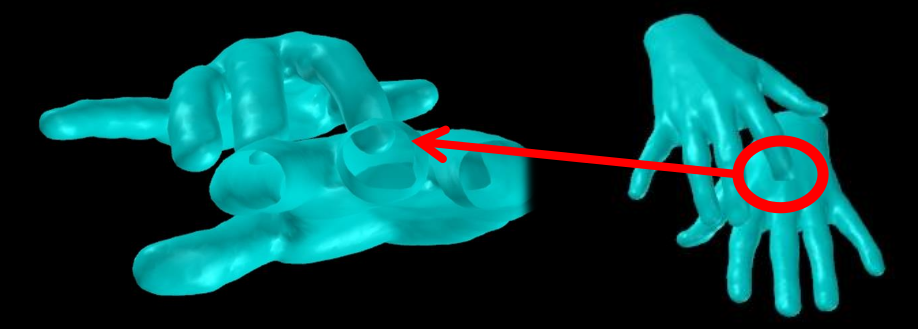}
	    \end{center}
	    \caption{``Walking'' sequence. Without the collision term unrealistic mesh intersections are observed during interactions}
	    \vspace*{-2mm}
	\label{fig:CORR_col_composition}
	\end{figure}

	\begin{figure*}[t]
	\begin{center}
	    \includegraphics[height=4.0cm]{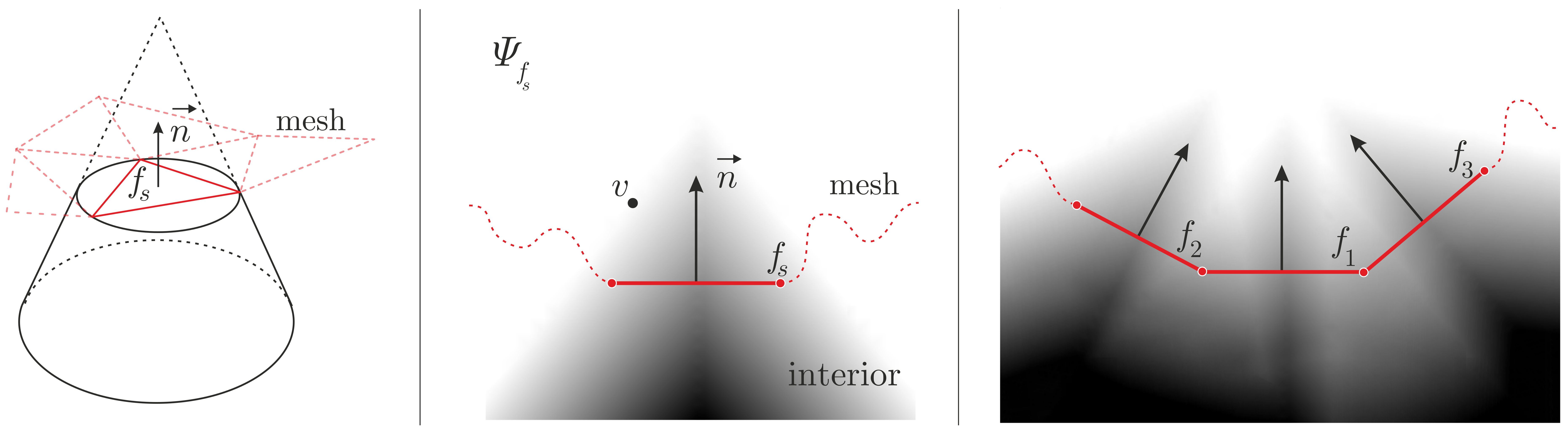}
	    \end{center}
	    \vspace*{-4mm}
	    \caption{(Left) Domain of the distance field $\Psi_{f_{s}}$ generated by the face $f_{s}$.
	    (Middle) Longitudinal section of the distance field $\Psi_{f_{s}}$: darker areas correspond
	    to higher penalties. (Right) Distance fields add up in case of multiple collisions}
	\label{fig:CollisionCone}
	\end{figure*}

\vspace*{-2mm}
	\subsubsection{Collision detection - $\mathcal{C}$}\label{subsec:collisionDetection}
\vspace*{-1mm}

		Collision detection is based on the observation that two objects cannot share the same space and
		is of high importance in case of self-penetration, inter-finger penetration or general intensive interaction, 
		as in the case depicted in Figure \ref{fig:CORR_col_composition}. 
		
		Collisions between meshes are detected by efficiently finding the set of colliding trianges $\mathcal{C}$ using bounding volume hierarchies (BVH)~\citep{collisionDeformableObjects}. 
		In order to penalize collisions and penetrations, we avoid using a signed 
		3D distance field for the whole mesh due to its high computational complexity and the fact that it has to be recomputed at every iteration of the optimization framework. 
		Instead, we resort to a more efficient approach with local 3D distance fields defined by the set of colliding triangles $\mathcal{C}$ that have the form of a cone as depicted in Figure \ref{fig:CollisionCone}. 
		In case of multiple collisions the defined conic distance fields are sumed up as shown in the same figure. 
		Having found a collision between two triangles $f_t$ and $f_s$, the amount of penetration can be computed by the position inside the conic distance fields. 
		The value of the distance field represents the intention of the repulsion that is needed to penalize the intrusion. 
		
		Let us consider the case where the vertices of $f_t$ are the \emph{intruders} and the triangle $f_s$ is the \emph{receiver} of the penetration. The opposite case is then similar. 
		The cone for computing the 3D distance field $\Psi_{f_s}: \mathbb{R}^3 \rightarrow \mathbb{R}_+$ is defined by the circumcenter of the triangle $f_s$. 
		\reviewChangeA
		{
		Letting $\normal_{f_s} \in \mathbb{R}^3$ denote the normal of the triangle, $\ob_{f_s} \in \mathbb{R}^3$ the circumcenter, and $r_{f_s} \in \mathbb{R}_{>0}$ the radius of the circumcircle, we have 
		}
		\begin{linenomath}
		\begin{equation}\label{eq:collision_PSI}
		\Psi_{f_s}(\vertex_t) = 
		\begin{cases} 
		~\vert ( 1-\Phi(\vertex_t) ) \Upsilon( \normal_{f_s} \cdot( \vertex_t-\ob_{f_s} ) ) \vert^2	&\Phi(\vertex_t)<1 	\\
		~0												&\Phi(\vertex_t)\ge1
		\end{cases}
		\end{equation}
		\end{linenomath}
		\begin{linenomath}
		\begin{equation}\label{eq:collision_FFF}
		\Phi(\vertex_t) = \frac{  \Vert (\vertex_t-\ob_{f_s})-(\normal_{f_s}\cdot(\vertex_t-\ob_{f_s}))\normal_{f_s} \Vert  }{  -\frac{r_{f_s}}{\sigma} (\normal_{f_s}\cdot(\vertex_t-\ob_{f_s}))+r_{f_s}  }
		\end{equation}
		\end{linenomath}
		\begin{linenomath}
		\begin{equation}\label{eq:collision_YYY}
		\Upsilon(x) = 
		\begin{cases} 
		-x+1-\sigma										&~~ x \le -\sigma		\\
		-\frac{1-2\sigma}{4\sigma^2}x^2 - \frac{1}{2\sigma}x + \frac{1}{4}(3-2\sigma)		&~~ x \in(-\sigma,+\sigma)	\\
		0											&~~ x \ge +\sigma.
		\end{cases}
		\end{equation}
		\end{linenomath}
		\reviewChangeA
		{
		The term $\Phi$ projects the vertex $\vertex$ onto the axis of the right circular cone defined by the triangle normal $\normal$ going through the circumcenter $\ob$ and measures the distance to it as illustrated in Figure~\ref{fig:CollisionCone}. The distance is scaled by the radius of the cone at this point. If $\Phi(\vertex)<1$ the vertex is inside the cone and if $\Phi(\vertex)=0$ the vertex is on the axis. The term $\Upsilon$ measures how far the projected point is from the circumcenter and defines the intensity of the repulsion. If $\Upsilon<0$, the projected point is behind the triangle. Within the range $(-\sigma,+\sigma)$, the penalizer term is quadratic with values between zero and one. If the penetration is larger than $\vert\sigma\vert$ the penalizer term becomes linear. The parameter $\sigma$ also defines the field of view of the cone and is fixed to $0.5$. 
		}
		
		\reviewChangeA
		{For each vertex penetrating a triangle, a repulsion term
		in the form of a 3D-3D correspondence that pushes the vertex back is computed. The direction of the
		repulsion 
		is given by the inverse normal direction of the vertex and its intensity by $\Psi$.  
		Using \emph{point-to-point} distances, the
		repulsion correspondences are computed for the set of colliding triangles $\mathcal{C}$:    
		}
		\begin{linenomath}
		\begin{equation}\label{eq:collisionDetection_p2p}
		\begin{split}
		E_{collision}(\para) = \sum_{(f_s(\para), f_t(\para))\in \mathcal{C}} \biggl\{ &\sum_{\vertex_s \in f_s} \Vert -\Psi_{f_t}(\vertex_s)\normal_s  \Vert^2 + \\ &\sum_{\vertex_t \in f_t} \Vert -\Psi_{f_s}(\vertex_t)\normal_t \Vert^2 \biggr\} 
		\end{split}
		\end{equation}
		\end{linenomath}
		Though not explicitly denoted, $f_s$ and $f_t$ depend on $\para$ and therefore also $\Psi$, $\vertex$ and $\normal$. For \emph{point-to-plane} distances, the equation gets simplified since $\normal^T\normal=1$:   
		\begin{linenomath}
		\begin{equation}\label{eq:collisionDetection_p2plane}
		\begin{split}
		E_{collision}(\para) = \sum_{(f_s(\para), f_t(\para))\in \mathcal{C}} \biggl\{	&	\sum_{\vertex_s \in f_s} \Vert -\Psi_{f_t}(\vertex_s)  \Vert^2 + 		\\ 
													&	\sum_{\vertex_t \in f_t} \Vert -\Psi_{f_s}(\vertex_t) \Vert^2 \biggr\}
		\end{split}
		\end{equation}
		\end{linenomath}

		This term takes part in the objective function \eqref{eq:obj} regulated by weight $\gamma_c$. An evaluation of different $\gamma_c$ values is presented in Section \ref{subsubsec:experimentCollisionDetection}.


	\subsubsection{Salient point detection - $\mathcal{S}$}\label{subsec:salientPointDetector}
	Our approach is so far based on a generative model, which provides accurate solutions in principle, but recovers only slowly from ambiguities and tracking errors. 
	However, this can be compensated by integrating a discriminatively trained salient point detector into a generative model. 
	
	To this end, we train a fingertip detector 
	on raw depth data. 
	\reviewChangeA
	{
	We manually 
	annotate\footnote{
		\reviewChangeA
		{
		All annotated sequences are available at \url{http://files.is.tue.mpg.de/dtzionas/hand-object-capture.html}} 
		}
	the fingertips of $56$ sequences consisting of approximately $2000$ frames, with $32$ of the sequences forming the training and $24$ forming the testing set. 
	We use a Hough forest~\citep{juergen_Hough} with $10$ trees, each trained with $100000$ positive and $100000$ negative patches. The negative patches are uniformly sampled from the background. The trees have a maximal depth of $25$ and a minimum leaf size of $20$ samples. 
	Each patch is sized $16\times16$ and consists of $11$ feature channels: $2$ channels obtained by a $5\times5$ min- and max-filtered depth channel and $9$ gradient features obtained by $9$ HOG bins using a $5\times5$ cell and soft binning.
	As for the pool of split functions at a test node, we randomly generate a set of $20000$ binary tests. 
	Testing is performed at multiple scales and non-maximum suppression is used to retain the most confident detections that do not overlap by more than $50\%$. 
	}
	
			\begin{table*}[t]
			\footnotesize 
			\begin{center}
				\caption{	\reviewChangeA{The graph contains $T$ mesh fingertips $\fingertip_t$ and $S$ fingertip detections $\detectionIMG_s$. 
						The cost of assigning a detection $\detectionIMG_s$ to a fingertip $\fingertip_t$ is given by $w_{st}$ as shown in table (a). 
						The cost of declaring a detection as false positive is $\lambda w_s$, where $w_s$ is the detection confidence. 
						The cost of not assigning any detection to fingertip $\fingertip_t$ is given by $\lambda$. 
						The binary solution of table (b) is constrained to sum up to one for each row and column}
				}
				\label{table:bipartiteTable}
				\setlength{\tabcolsep}{1pt}
				\hspace*{-06mm}
				\begin{tabular}{|c|c|c|c|c|c|c|c|c|}
					\cline{4-7}\cline{9-9}
					\multicolumn{2}{c}{\multirow{2}{*}{(a)}} & \multicolumn{1}{c}{} & \multicolumn{4}{|c|}{Fingertips \reviewChangeA{$\fingertip_t$}} & \multicolumn{1}{c}{} & \multicolumn{1}{|c|}{V} \\
					\cline{4-7}\cline{9-9}
					\multicolumn{2}{c}{} & \multicolumn{1}{c}{} & \multicolumn{1}{|c|}{\reviewChangeA{$\fingertip_1$}} & \reviewChangeA{$\fingertip_2$} & {\dots} & \reviewChangeA{$\fingertip_T$} & \multicolumn{1}{c}{} & \multicolumn{1}{|c|}{$\alpha$} \\
					\cline{4-7}\cline{9-9}
					\noalign{\smallskip}
					\cline{1-2}\cline{4-7}\cline{9-9}
					\multirow{5}{*}{\centering\begin{turn}{90}Detections $\detectionIMG_s$\end{turn} }  & {	\textit{$\detectionIMG_1$}	} & {} & { $w_{11}$ } & { $w_{12}$ } & { $\dots$  } & { $w_{1T}$ } & {} & { $\lambda w_1$ }                \\ \cline{2-2}\cline{4-7}\cline{9-9}
															  {} & 	{	\textit{$\detectionIMG_2$}	} & {} & { $w_{21}$ } & { $w_{22}$ } & { $\dots$  } & { $w_{2T}$ } & {} & { $\lambda w_2$ }                \\ \cline{2-2}\cline{4-7}\cline{9-9}
															  {} & 	{	\textit{$\detectionIMG_3$}	} & {} & { $w_{31}$ } & { $w_{32}$ } & { $\dots$  } & { $w_{3T}$ } & {} & { $\lambda w_3$ }                \\ \cline{2-2}\cline{4-7}\cline{9-9}
															  {} & 	{	         \vdots  	  	} & {} & { $\vdots$ } & { $\vdots$ } & { $\ddots$ } & { $\vdots$ } & {} & { $\vdots                      $ }                \\ \cline{2-2}\cline{4-7}\cline{9-9}
															  {} & 	{	\textit{$\detectionIMG_S$}	} & {} & { $w_{S1}$ } & { $w_{S2}$ } & { $\dots$  } & { $w_{ST}$ } & {} & { $\lambda w_S$ }                \\ 
					\cline{1-2}\cline{4-7}\cline{9-9}
					\noalign{\smallskip}
					\cline{1-2}\cline{4-7}\cline{9-9}
					\multirow{1}{*}{\centering\begin{turn}{90}V\end{turn}                     	   } & 	{        \textit{$\beta$}		} & {} & { $\lambda$ } & { $\lambda$ } & {\dots} & { $\lambda$ } & {} & { $\infty$ } \\ 
					\cline{1-2}\cline{4-7}\cline{9-9}
				\end{tabular}
				\hspace*{+05mm}
				\begin{tabular}{|c|c|c|c|c|c|c|c|c|}
					\cline{4-7}\cline{9-9}
					\multicolumn{2}{c}{\multirow{2}{*}{(b)}} & \multicolumn{1}{c}{} & \multicolumn{4}{|c|}{Fingertips \reviewChangeA{$\fingertip_t$}} & \multicolumn{1}{c}{} & \multicolumn{1}{|c|}{V} \\
					\cline{4-7}\cline{9-9}
					\multicolumn{2}{c}{} & \multicolumn{1}{c}{} & \multicolumn{1}{|c|}{\reviewChangeA{$\fingertip_1$}} & \reviewChangeA{$\fingertip_2$} & {\dots} & \reviewChangeA{$\fingertip_T$} & \multicolumn{1}{c}{} & \multicolumn{1}{|c|}{$\alpha$} \\
					\cline{4-7}\cline{9-9}
					\noalign{\smallskip}
					\cline{1-2}\cline{4-7}\cline{9-9}
					\multirow{5}{*}{\centering\begin{turn}{90}Detections $\detectionIMG_s$\end{turn} } & 	{	\textit{$\detectionIMG_1$}	} & {} & { $e_{11}$ } & { $e_{12}$ } & { $\dots$  } & { $e_{1T}$ } & {} & { $\alpha_1$ }                \\ \cline{2-2}\cline{4-7}\cline{9-9}
															  {} & 	{	\textit{$\detectionIMG_2$}	} & {} & { $e_{21}$ } & { $e_{22}$ } & { $\dots$  } & { $e_{2T}$ } & {} & { $\alpha_2$ }                \\ \cline{2-2}\cline{4-7}\cline{9-9}
															  {} & 	{	\textit{$\detectionIMG_3$}	} & {} & { $e_{31}$ } & { $e_{32}$ } & { $\dots$  } & { $e_{3T}$ } & {} & { $\alpha_3$ }                \\ \cline{2-2}\cline{4-7}\cline{9-9}
															  {} & 	{	         \vdots  	  	} & {} & { $\vdots$ } & { $\vdots$ } & { $\ddots$ } & { $\vdots$ } & {} & { $\vdots                      $ }                \\ \cline{2-2}\cline{4-7}\cline{9-9}
															  {} & 	{	\textit{$\detectionIMG_S$}	} & {} & { $e_{S1}$ } & { $e_{S2}$ } & { $\dots$  } & { $e_{ST}$ } & {} & { $\alpha_S$ }                \\ 
					\cline{1-2}\cline{4-7}\cline{9-9}
					\noalign{\smallskip}
					\cline{1-2}\cline{4-7}\cline{9-9}
					\multirow{1}{*}{\centering\begin{turn}{90}V\end{turn}                     	   } & 	{        \textit{$\beta$   }		} & {} & { $\beta_1$ } & { $\beta_2$ } & {\dots} & { $\beta_T$ } & {} & { $0$ } \\ 
					\cline{1-2}\cline{4-7}\cline{9-9}
				\end{tabular}
			\end{center}
		\end{table*}
		
	Since we resort to salient points only for additional robustness, it is usually sufficient to have only sparse fingertip detections. 
	We therefore collect detections with a high confidence, 
	choosing a threshold of $c_{thr}=3.0$ for our experiments. 
	\reviewChangeA
	{
	The association between the
		$T$
		fingertips 
		$\fingertip_t$
		of the model depicted in Figure \ref{fig:modelsAndVOIs_HAND} (d-f) and the 
		$S$
		detections 
		$\detectionIMG_s$ is solved by integer programming~\citep{Malik_Bipartite}:
	\begin{linenomath}
	\begin{equation}\label{eq:bipartite_MIP}
	\begin{split}
	\argmin_{e_{st},\alpha_s,\beta_t} 		\qquad 		& \sum_{s,t}e_{st}w_{st} + \lambda\sum_{s}\alpha_s w_s + \lambda\sum_{t}\beta_t			\\
	\text{subject to} 				\qquad		& \sum_{s}e_{st} + \beta_t = 1 		\qquad \reviewChangeA{\forall t \in \{1,...,T\}}		\\
									& \sum_{t}e_{st} + \alpha_s  = 1 	\qquad \reviewChangeA{\forall s \in \{1,...,S\}}		\\
									& e_{st}, \alpha_s, \beta_t \in \{0,1\}
	\end{split}
	\end{equation}
	\end{linenomath}
	As illustrated in Table~\ref{table:bipartiteTable}, $e_{st}=1$ defines an assignment of a detection $\detectionIMG_s$ to a fingertip $\fingertip_t$. The assignment cost is defined by $w_{st}$. If $\alpha_s=1$, the detection  $\detectionIMG_s$ is declared as a false positive with cost $\lambda w_s$ and if $\beta_t=1$, the fingertip $\fingertip_t$ is not assigned to any detection with cost $\lambda$. 
		}

	\reviewChangeA
	{
	The weights $w_{st}$ are given by the 3D distance between the detection $\delta_s$ and the finger of the model $\fingertip_t$. 
	For each finger $\fingertip_t$, a set of vertices are marked in the model. 
	The distance is then computed between the 3D centroid of the visible vertices of $\fingertip_t$ (Figure \ref{fig:modelsAndVOIs_HAND}d-f) and the centroid of the detected region $\delta_s$. The latter is computed based on the 3D point cloud $\detectionPCL_s$ corresponding to the detection bounding box.}
	For the weights $w_s$, we investigate two approaches. 
	The first approach uses $w_s=1$. 
	The second approach takes the confidences $c_s$ of the detections into account by setting $w_s=\frac{c_s}{c_{thr}}$. 
	The weighting parameter $\lambda$ is evaluated in Section \ref{subsubsec:experimentSalientPointDetector}.

\begin{figure}[t]
	\captionsetup[subfigure]{}
	\centering						         
	\subfloat[subfigure 3 CORR_det_p2p][]{		\includegraphics[trim=00mm 00mm 00mm 00mm, clip=true, height=0.177 \textwidth]{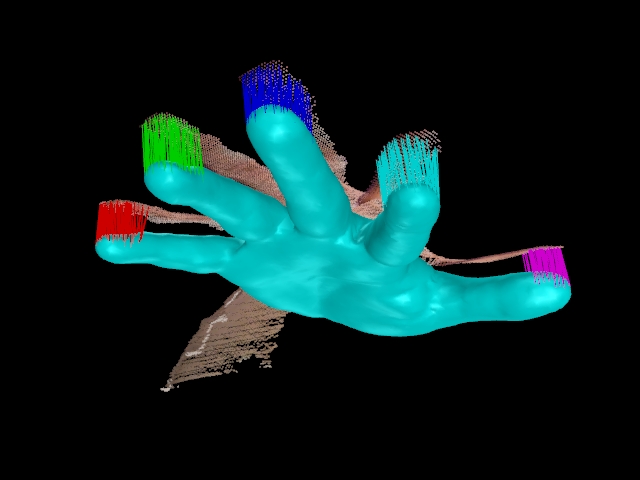}	\label{fig:CORR_det_p2p_p2centr___p2p}		}	\hspace*{-1.1em}
	\subfloat[subfigure 3 CORR_Det_p2centr][]{	\includegraphics[trim=00mm 00mm 00mm 00mm, clip=true, height=0.177 \textwidth]{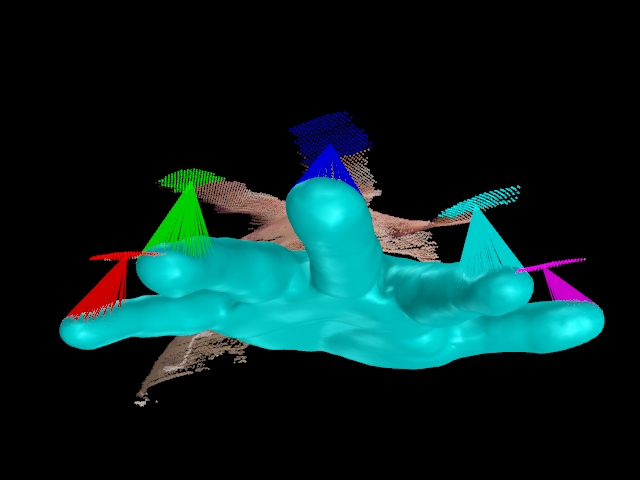}	\label{fig:CORR_det_p2p_p2centr___p2centr}	}
	\vspace*{-1mm}
	\caption{\reviewChangeA{	Correspondences between the fingertips $\fingertip_t$ of the model and
				    (a) the closest points of the associated detections $\detectionPCL_s$ (b) the centroids of the associated detections $\detectionPCL_s$
				} 
	}
	\label{fig:CORR_det_p2p_p2centr}
	\end{figure}

	\reviewChangeA
	{
	If a detection $\detectionIMG_s$ has been associated to a fingertip $\fingertip_t$, we have to define correspondences between the set of visible vertices of $\fingertip_t$ 
	and the detection point cloud $\detectionPCL_s$. If the fingertip $\fingertip_t$ is already very close to $\detectionPCL_s$, \ie 
	$w_{st} < 10 mm$, 
	we do not compute any correspondences since the localization accuracy of the detector is not higher. 
	In this case just the close proximity of the fingertip $\fingertip_t$ to the data 
	suffices for a good alignment. 
	Otherwise, we compute the closest points between the vertices $\vertex_i$ and the points $\xClPoint_i$ of the detection $\detectionPCL_s$ as illustrated in Figure~\ref{fig:CORR_det_p2p_p2centr___p2p}:    
	}
	\reviewChangeA
	{
	\begin{linenomath}
	\begin{equation}\label{eq:salientPoint_residual_p2p}
	 E_{salient}(\para,D) = \sum_{s,t}e_{st} \reviewChangeB{\biggl(} \sum_{(\xClPoint_i,\vertex_i) \in \detectionPCL_s \times \fingertip_t} \Vert \vertex_i(\para) - \xClPoint_i \Vert^2 \reviewChangeB{\biggr)}
	\end{equation}
	\end{linenomath}
	}
	As in \eqref{eq:errorResidual_p2pl}, a \emph{point-to-plane} distance metric can replace the \emph{point-to-point} metric.
 	\reviewChangeA
 	{
	When less than 50\% of the vertices of $\fingertip_t$ project inside the detection bounding box, we even avoid the additional step of computing correspondences between the vertices and the detection point cloud. Instead we associate all vertices with the centroid of the detection point cloud as shown in Figure~\ref{fig:CORR_det_p2p_p2centr___p2centr}. 
 	}
	
\vspace*{2mm}

	\subsubsection{Physics Simulation - $\mathcal{P}$}\label{subsec:physicsComponent}

\begin{figure}[t]
	\captionsetup[subfigure]{}
	\centering								
		\subfloat[subfigure 1 physicsConcept][]{	\includegraphics[trim=020mm 022mm 025mm 012mm, clip=true, width=0.23\textwidth]{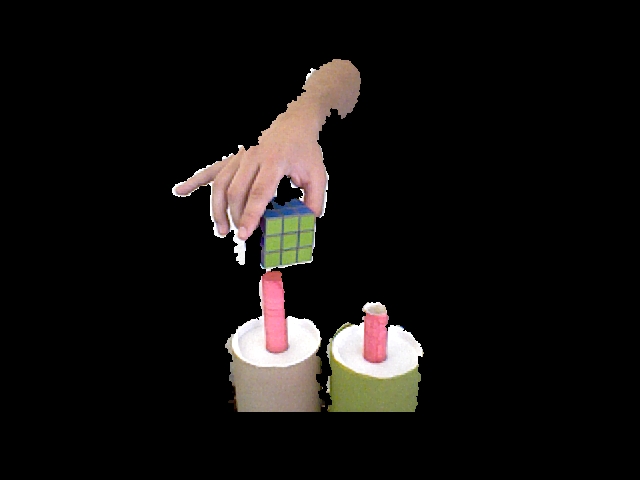}				\label{fig:physicsComponent_Concept___rgbd}		}		\hspace*{-5mm}
		\subfloat[subfigure 2 physicsConcept][]{	\includegraphics[trim=020mm 022mm 025mm 012mm, clip=true, width=0.23\textwidth]{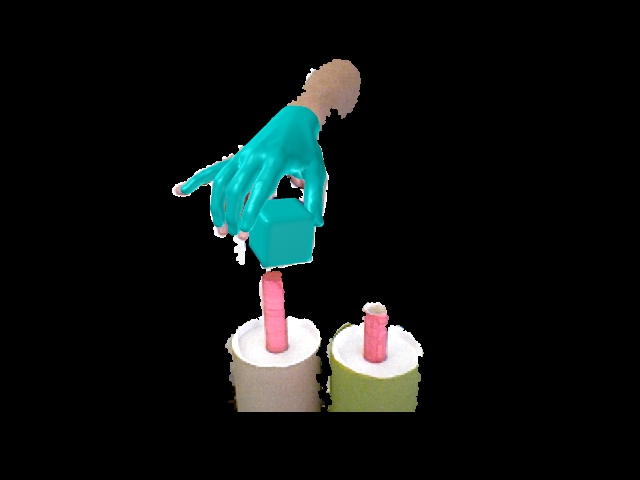}			\label{fig:physicsComponent_Concept___synth}		}	\\	\vspace*{-2mm}
		\subfloat[subfigure 3 physicsConcept][]{	\includegraphics[trim=020mm 022mm 025mm 012mm, clip=true, width=0.23\textwidth]{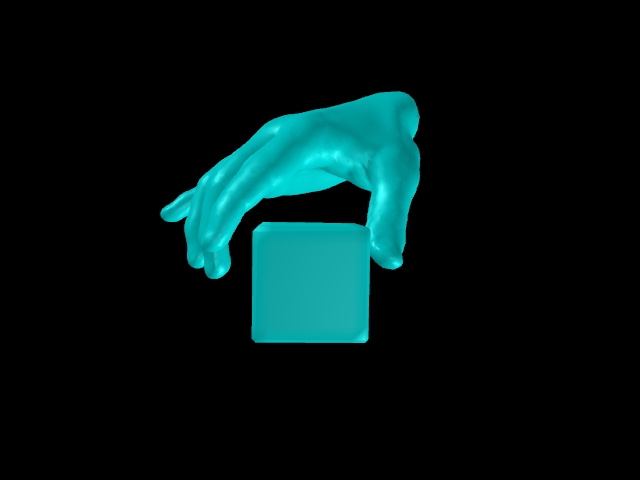}		\label{fig:physicsComponent_Concept___synthZoomBAD}	}		\hspace*{-5mm}
		\subfloat[subfigure 4 physicsConcept][]{	\includegraphics[trim=020mm 022mm 025mm 012mm, clip=true, width=0.23\textwidth]{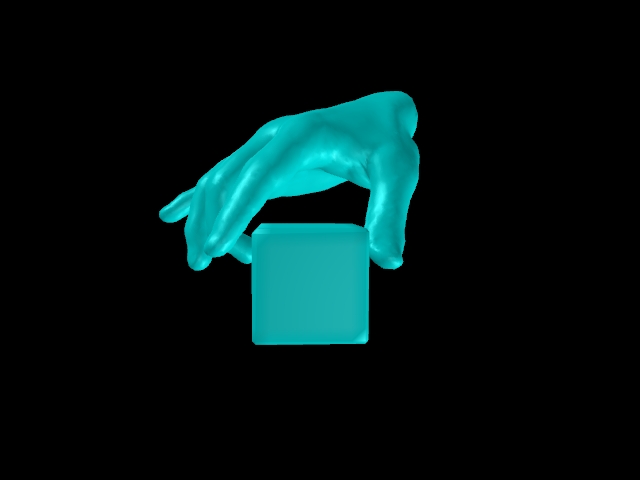}		\label{fig:physicsComponent_Concept___synthZoomGOOD}	}
		\vspace*{-2mm}
	\caption{	Physical plausibility during hand-object interaction. (a) Input RGB-D image. (b-c) Obtained results without the physics component. 
			(d) Obtained results with the physics component, ensuring a more realistic pose during interaction 
	}
	\label{fig:physicsComponent_Concept}
	\end{figure}

	\reviewChangeA{A phenomenon that frequently occurs in the context of hand-object interaction are physically unrealistic poses due to occlusions or missing data.} 
	Such an example is illustrated in Figure \ref{fig:physicsComponent_Concept}, where a cube is grasped and moved by two fingers. Since one of the fingers that is in contact with the cube is occluded, the estimated pose is physical unrealistic. Due to gravity, the cube would fall down.

\begin{figure}[t]
	\captionsetup[subfigure]{labelformat=empty}
	\centering								         
		\subfloat[subfigure 3 physicsHulls_dummyFront][]{	\includegraphics[trim=10mm 05mm 10mm 03mm, clip=true, width=0.23 \textwidth]{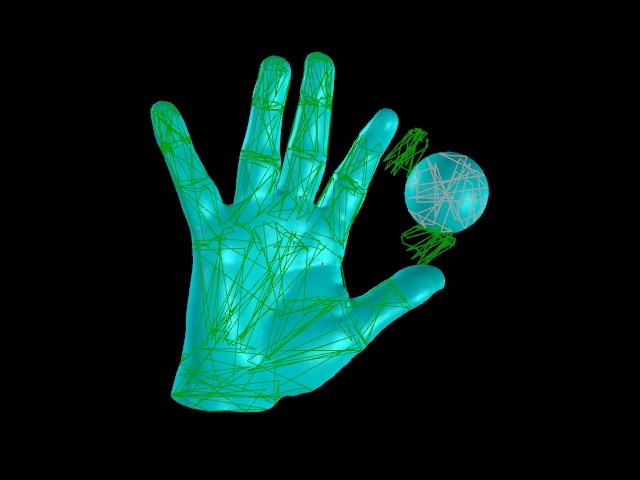}	}		\hspace*{-1.1em}
		\subfloat[subfigure 3 physicsHulls_dummyFront][]{	\includegraphics[trim=10mm 05mm 10mm 03mm, clip=true, width=0.23 \textwidth]{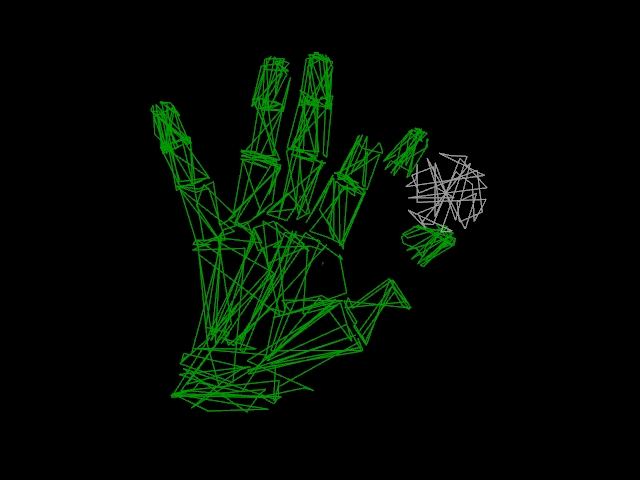}		}	
		\vspace*{-4mm}
	\caption{	
			Low resolution representation of the hands and objects for the 
			physics simulation. In order to predict the finger parts (green) that give the physically most stable results if they were in contact with the object (white), all combinations of finger parts close to the object are evaluated.     
			The image shows how two finger parts are moved to the object for examining the contribution to the stability of the object. The stability is measured by a physics simulation where all green parts are static
	}
	\label{fig:physicsHulls_dummyFront}
	\end{figure}

	In order to compensate for this during hand-object interaction scenarios, we resort to physics simulation \citep{BulletPhysics} 
	for additional realism and physical plausibility. To this end, we model the static scene as well and based on gravity and the parameters friction, restitution, and mass for each object we can run a physics simulation.
	To speed up the simulation, we represent each body or object part defined by the skinning weights as shown in Figure~\ref{fig:objectTemplateMeshes_skinSegmentation} as convex hulls. This is visualized in Figure~\ref{fig:physicsHulls_dummyFront}.

\begin{figure}[t]
	\captionsetup[subfigure]{labelformat=empty}
	\centering								
		\subfloat[subfigure 1 jointsErrorMetric][]{ \includegraphics[trim=15mm 28mm 18mm 18mm, clip=true, width=0.25 \textwidth]{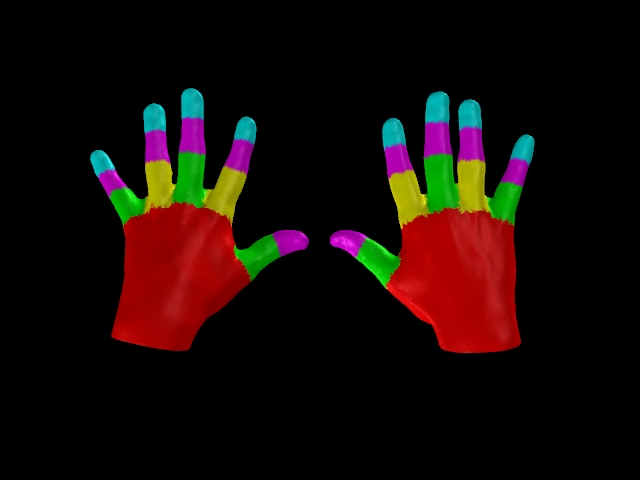} 
		}
		\vspace*{-4mm}
	\caption{	Finger parts 
			that form all possible supporting combinations in the physics simulation component.
			Parts with red color do not take part in this process 
	}
	\label{fig:proximity_bonesIntoAccount}
	\end{figure}

	\reviewChangeA{Given current pose estimates of the hands and the manipulated object, we first evaluate if the current solution is physically plausible.} 
	To this end, we run the simulation for $35$ iterations with a time-step of $0.1$ seconds. If the centroid of the object moves by less than $3$mm we consider the solution as stable. Otherwise, we have to search for the hand pose which results in a more stable estimate. Since it is intractable to evaluate all possible hand poses, we search only for configurations which require a minor change of the hand pose. This is a reasonable assumption for our tracking scenario. To this end, we first compute the distances between all parts of the fingers, as depicted in Figure~\ref{fig:proximity_bonesIntoAccount}, and the object~\citep{CGAL_distance1,CGAL_distance2}. Each finger part with distance less than $10$mm is then considered as candidate for being in contact with the object and each combination of at least two and maximum four candidate parts is taken into account.
 	
	The contribution of each combination to the stability of the object is examined through the physics simulation after rigidly moving the corresponding finger parts towards the closest surface point of the object. 
	Figure~\ref{fig:physicsHulls_dummyFront} illustrates the case for a combination of two finger parts. 
	The simulation is repeated for all combinations and we select the combination with the lowest movement of the object, \ie the smallest displacement of its centroid from the initial position. 
	Based on the selected combination, we define an energy that forces the corresponding finger parts to get in contact with the object by minimizing the closest distance between the parts $i$ and the object:  
	\begin{linenomath}
	\begin{equation}\label{eq:physiceResidual_p2p}
	 E_{physics}(\para) = \sum_i  \Vert \vertex_i(\theta) - \xClPoint_i \Vert^2 
	\end{equation}
	\end{linenomath}
	The vertices $\vertex_i$ and $\xClPoint_i$ correspond to the closest point of a finger part and the object, respectively.  
	As in \eqref{eq:errorResidual_p2pl}, a \emph{point-to-plane} distance metric can replace the \emph{point-to-point} metric.

	\subsubsection{Anatomical limits}\label{subsec:angleLimits}

	\begin{figure}[t]
	\captionsetup[subfigure]{}
	\centering																							\hspace*{-01.5mm}
		\subfloat[subfigure 1 angleLimits][]{	\includegraphics[trim=10mm 00mm 00mm 00mm, clip=true, height=0.200 \textwidth]{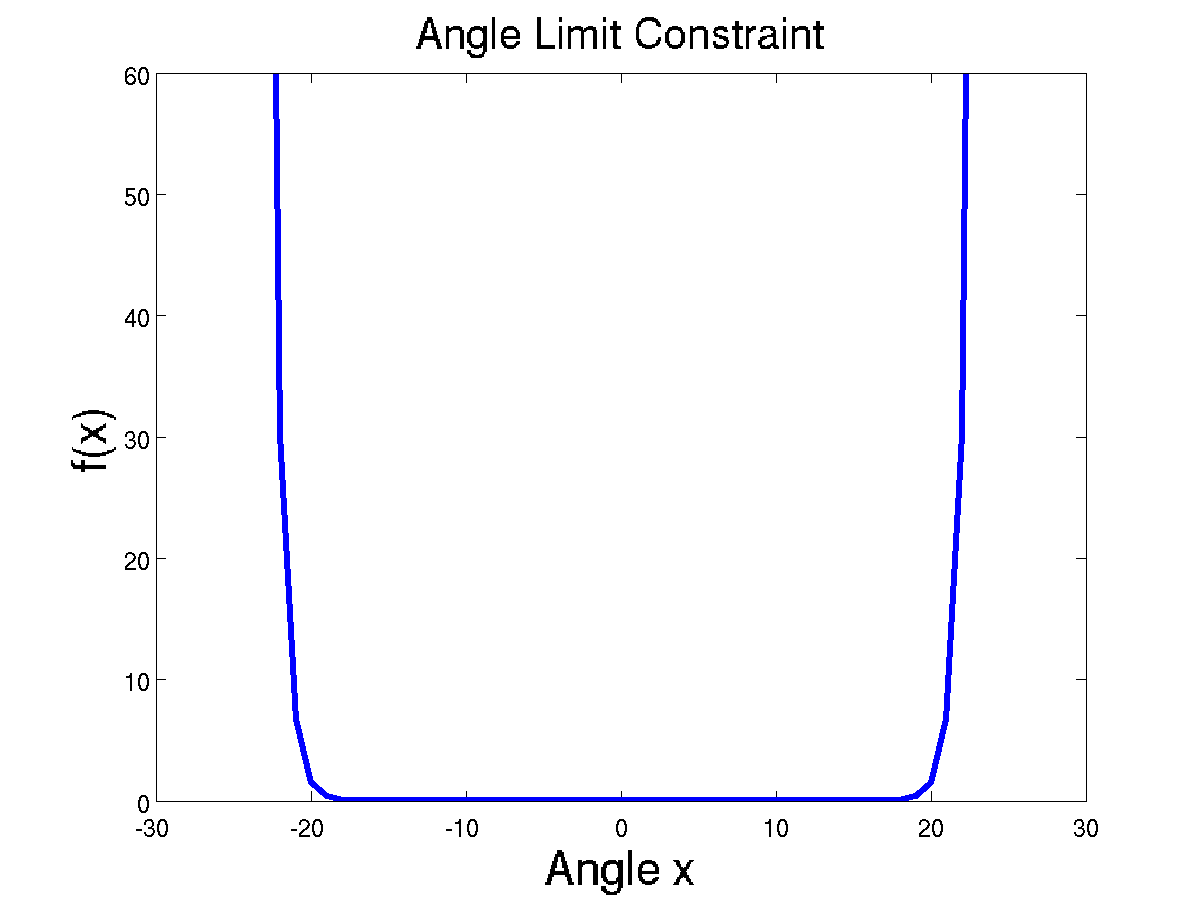}	\label{fig:angleLimits_f}	}	\hspace*{-05.5mm}
		\subfloat[subfigure 2 angleLimits][]{	\includegraphics[trim=10mm 00mm 00mm 00mm, clip=true, height=0.200 \textwidth]{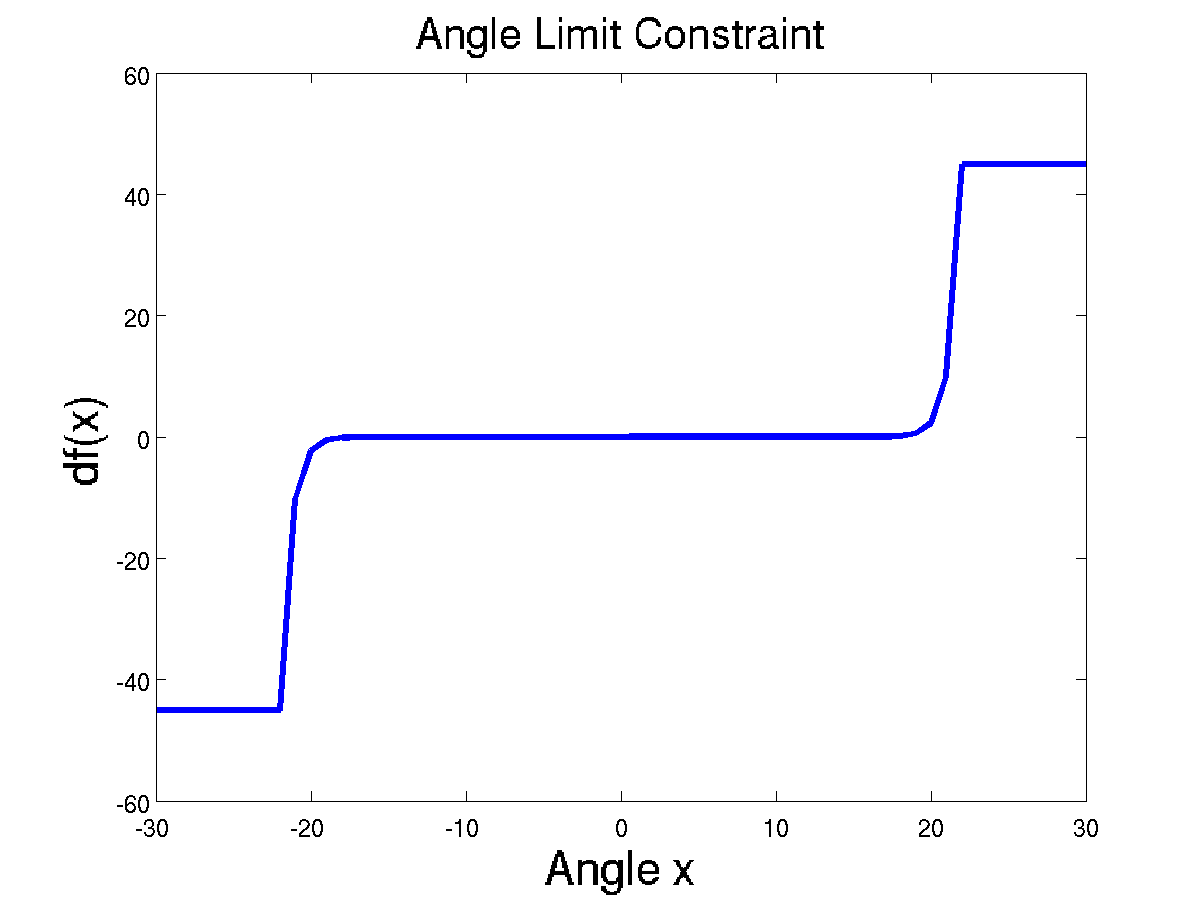}	\label{fig:angleLimits_df}	}
	\caption{	
		\reviewChangeA
		{
			Angle limits are independently defined for each revolute joint. 
			The plot visualizes the function (a), and its truncated derivative (b), that penalizes the deviation from an allowed range of $\pm20.0$ degree 
		}
	}
	\label{fig:angleLimits}
	\end{figure}
	
	\reviewChangeA
	{
	Anatomically inspired joint-angle limits \citep{AnatomicalModel} are enforced as soft constraints by the term: 
				\begin{linenomath}
					\begin{equation}\label{eq:angleLimits_f}
						  E_{anatomy}(\para) = \sum_k \left( \exp\left(p(l_k-\para_k)\right) + \exp\left(p(\para_k-u_k)\right) \right)
					\end{equation}
				\end{linenomath}
	where $p=10$. The index $k$ goes over all revolute joints and $[u_k, l_k]$ is the allowed range for each of them. The term is illustrated for a single revolute joint in Figure \ref{fig:angleLimits}. We use $\gamma_a = 0.0015~C_{all}$, where $C_{all}$ is the total number of correspondences.  
	}


	\subsubsection{Regularization}\label{subsec:regulizer}

	\reviewChangeA
	{
	In case of occlusions or due to missing depth data, the objective function \eqref{eq:obj} based solely on the previous terms can be ill-posed. We therefore add a term that penalizes deviations from the previous estimated joint angles $\tilde{\para}$:  
	\begin{linenomath}
		\begin{equation}\label{eq:physiceResidual_p2pl}
		E_{regularization}(\para) = \sum_k ( \para_k - \tilde{\para}_k )^2.
		\end{equation}
	\end{linenomath}
	We use $\gamma_r = 0.02~C_{all}$. 
	}


	\subsubsection{Optimization}\label{subsec:optimization}

	\newcommand{\algHSPACE}{\hspace*{-8.5mm}}
	
	\begin{algorithm}[t]
		\caption{
				\reviewChangeA{Pose estimation for RGB-D data with \emph{point-to-point} distances}
		}
		\label{pseudo:trackingOURs}
		\footnotesize
		\SetKwFor{Repeat}{Repeat until convergence or max $i_{thr}$ iterations}{}{end}
		$\tilde{\para} = $ pose estimate of the previous frame														\\
		$i=0$, $\para_0 = \tilde{\para}$																\\
		\Repeat{} {
		    - Render meshes at pose $\theta$																\\
		    - Find corresp. $LO_{m2d}$		\hspace*{+0.00mm}	Section \ref{subsec:m2d}			-~Eq. \eqref{eq:errorResidual_p2p}		\\
		    - Find corresp. $LO_{d2m}$		\hspace*{+0.00mm}	Section \ref{subsec:d2m}			-~Eq. \eqref{eq:errorResidual_Pluecker}	\\
		    - Find corresp. $\mathcal{C}$ 	\hspace*{+8.40mm}	Section \ref{subsec:collisionDetection}	-~Eq. \eqref{eq:collisionDetection_p2p}	\\
		    - Find corresp. $\mathcal{S}$ 	\hspace*{+8.20mm}	Section \ref{subsec:salientPointDetector}	-~Eq. \eqref{eq:salientPoint_residual_p2p}	\\
		    - Find corresp. $\mathcal{P}$ 	\hspace*{+7.90mm}	Section \ref{subsec:physicsComponent}		-~Eq. \eqref{eq:physiceResidual_p2p}		\\
		    \normalsize
		    $\theta _{i+1}=\arg \min_{\theta}E(\theta,D)$					\\
		    \footnotesize
		    $i=i+1$																			\\
		}
		\label{alg:algorithm}
		\normalsize
	\end{algorithm}

	\reviewChangeA
	{For pose estimation, we alternate between computing the correspondences 
		$LO_{m2d}$		(Section \ref{subsec:m2d}), 
		$LO_{d2m}$		(Section \ref{subsec:d2m}), 
		$\mathcal{C}$		(Section \ref{subsec:collisionDetection}), 
		$\mathcal{S}$		(Section \ref{subsec:salientPointDetector}),	and 
		$\mathcal{P}$		(Section \ref{subsec:physicsComponent})
		according to the current pose estimate and optimizing the objective function \eqref{eq:obj} based on them as summarized in Algorithm \ref{pseudo:trackingOURs}. 
		This process is repeated until convergence or until a maximum number of iterations $i_{thr}$ is reached. 
		It should be noted that the objective function $E(\para,D)$ is only differentiable for a given set of correspondences.  
		We optimize $E(\para,D)$, which is a non-linear least squares problem, with the Gauss-Newton method as in \citep{Bro10}.   
	}


\subsection{Multicamera RGB}\label{sec:Differences_MultiRGB_2_SingRGBD}

	The previously described approach can also be applied to multiple synchronized RGB videos. To this end, the objective function \eqref{eq:obj} needs to be changed only slightly due to the differences of depth and RGB data. While the error is directly minimized in 3D for RGB-D data, we minimize the error for RGB images in 2D since all our observations are 2D. \reviewChangeA{Instead of using 
	a 3D \emph{point-to-point} \eqref{eq:errorResidual_p2p} or \emph{point-to-plane} \eqref{eq:errorResidual_p2pl} measure, the error is therefore given by
	\begin{linenomath}
	\begin{equation}\label{eq:errorResidual_LUCA}
	\sum_c\sum_{i} \Vert \Pi_{c}(\vertex_{i}(\para)) - \xPrPoint_{i,c} \Vert^2\;
	\end{equation}
	\end{linenomath}
	where $\Pi _{c}:\mathbb{R}^{3}\longrightarrow\mathbb{R}^{2}$	
	are the known projection functions, mapping 3D points into the image plane of each static camera $c$, and $(\vertex_{i},x_{i,c})$ is a correspondence between a 3D vertex and a 2D point.} 
	Furthermore, the salient point detector, introduced in Section~\ref{subsec:salientPointDetector}, is not applied to the depth data but to all camera views. 
	Since multiple high resolution views allow to detect more distinctive image features, we do not detect finger tips but finger nails in this case. 
	
	The only major change is required for the data terms $E_{model \rightarrow data}(\para,D)$ and $E_{data \rightarrow model}(\para,D)$ in \eqref{eq:obj}.  
	The term $E_{data \rightarrow model}(\para,D)$ is replaced by an edge term that matches edge pixels in all camera views to the edges of the projected model in the current pose $\para$.  
	As in the RGB-D case, the orientation of the edges is taken into account for matching and mismatches are removed by thresholding. 
	Working with 2D distances though has the disadvantage of not being able to apply intuitive 3D distance thresholds, as presented in 
	Section \ref{subsec:d2m}. 
	In order to have an alternative rejection mechanism of noisy correspondences, 
	we compute for each bone the standard deviation of the 2D error that is suggested by all of its correspondences. 
	Subsequently, correspondences that suggest an error bigger than twice this standard deviation are rejected as outliers. 
	The second term $E_{model \rightarrow data}(\para,D)$ is replaced by a term based on optical flow as in \citep{Luca3DPVT08}. 
	The term introduces temporal consistency and harness the higher resolution and frame rates of the RGB data in comparison to the RGB-D data.

\section{Experimental Evaluation}\label{sec:experimentalEvaluation}
	
	Benchmarking in the context of 3D hand tracking remains an open problem \citep{Review_Erol_HandPose} 
	despite recent contributions \citep{srinath_iccv2013,GCPR_2013_Tzionas_Gall,GCPR_2014_Tzionas_Gall, TKKIM_CVPR14_LatentRegressionForest, TKKIM_ICCV13_Real_time_Articulated_Hand, MSR_ASIA_handTracking, NYU_tracker_tompson14tog}. 
	The vast majority of them focuses on the problem of single hand tracking, especially in the context of real-time human computer interaction, neglecting challenges occurring during the interaction between two hands or between hands and objects. 
	For this reason we captured 29 sequences in the context of hand-hand and hand-object interaction. 
	\reviewChangeA{The sequences were captured either with a single RGB-D camera or with $8$ 
	synchronized RGB cameras.} 
	While 20 sequences have been used in the preliminary works~\citep{LucaHands,GCPR_2014_Tzionas_Gall}, which include interactions with rigid objects, the 9 newly captured sequences also include interactions with non-rigid objects.

	We first evaluate our approach on RGB-D sequences with hand-hand interactions in Section~\ref{sec:experimentalEvaluation___Monocular_RGBD__hand2hand}. Sequences with hand-object interactions are used in Section~\ref{sec:experimentalEvaluation___Monocular_RGBD__hand2object} for evaluation and finally our approach is evaluated on sequences captured with several RGB cameras in Section~\ref{sec:experimentalEvaluation___Multicamera_RGB}.
	
		\begin{figure}[t]
	\captionsetup[subfigure]{labelformat=empty}
	\centering
		\subfloat[subfigure 1 jointsErrorMetric][]{
			\includegraphics[width=0.30 \textwidth]{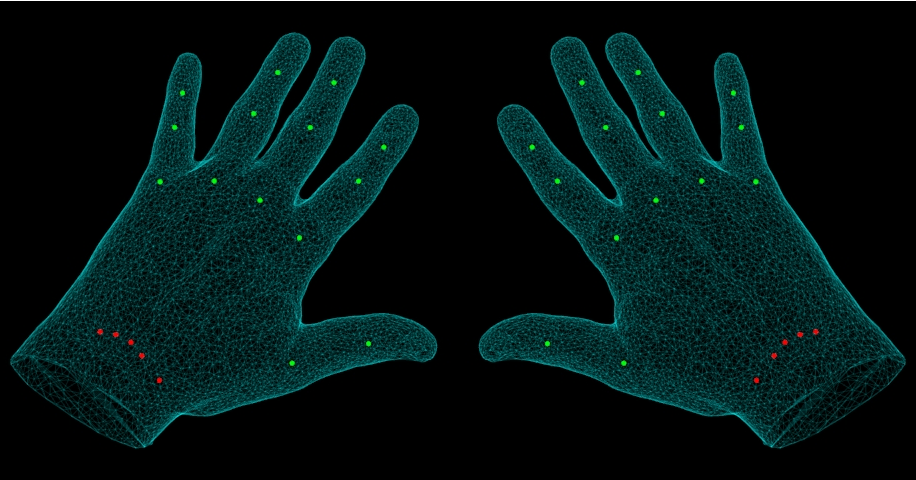} 
		}
		\vspace*{-5mm}
	\caption{	Hand joints used for quantitative evaluation. Only the green joints of our hand model are used for measuring the pose estimation error  
	}
	\label{fig:jointsErrorMetric}
	\end{figure}

\subsection{Monocular RGB-D - Hand-Hand Interactions}\label{sec:experimentalEvaluation___Monocular_RGBD__hand2hand}
	
	Related RGB-D methods \citep{OikonomidisBMVC} usually report quantitative results only on synthetic sequences, which inherently include ground-truth, while for realistic conditions they resort to qualitative results.
	
	\begin{sloppypar}
	Although qualitative results are informative, 
	quantitative evaluation based on ground-truth is of high importance. 
	We therefore manually annotated $14$ sequences,  
	$11$ of which are used to evaluate the components of our pipeline and $3$ 
	for comparison with the state-of-the-art method \citep{OikonomidisBMVC}. 
	These sequences contain motions of a single hand and two interacting hands with $37$ and $74$ DoF, respectively. 
	They vary from $100$ to $270$ frames and contain several actions, like 
	\textit{``Walking''}, 
	\textit{``Crossing''}, 
	\textit{``Crossing and Twisting''}, 
	\textit{``Tips Touching''}, 
	\textit{``Dancing''}, 
	\textit{``Tips Blending''}, 
	\textit{``Hugging''}, 
	\textit{``Grasping''}, 
	\textit{``Flying''}, as well as performing the 
	\textit{``Rock''} and 
	\textit{``Bunny''} gestures. 
	As indicator for the accuracy of the annotations, we measured the standard deviation of 4 annotators, which is $1.46$ pixels. 
	All sequences were captured in $640x480$ resolution at $30$ fps with a Primesense Carmine $1.09$ camera. 
	\end{sloppypar}

	The error metric for our experiments is the 2D distance (pixels) between the projection of the 3D joints and the corresponding 2D annotations. 
	The joints taken into account in the metric are depicted in Figure \ref{fig:jointsErrorMetric}. 
	Unless explicitly stated, we report the average over all frames of all relevant sequences.


	Our system is based on an objective function consisting of five terms, described in Section \ref{sec:optimization}. 
	Two of them minimize the error between the posed mesh and the depth data by fitting the \emph{model to the data} and the \emph{data to the model}. 
	\reviewChangeA{A \emph{salient point} detector further constrains the pose using fingertip detections in the depth image, 
	while a \emph{collision detection} method contributes to realistic pose estimates that are physically plausible.}
	The \reviewChangeB{function} is complemented by the \emph{physics simulation} component, that contributes towards more realistic interaction of hands with objects.
	However, this component is only relevant for hand-object interactions and thus it will be studied in detail in Section~\ref{sec:experimentalEvaluation___Monocular_RGBD__hand2object}. 
	In the following, we evaluate each component and the parameters of the objective function \eqref{eq:obj}.
	
	\begin{table}[t]
		\footnotesize 
		\begin{center}
			\caption{	Evaluation of \emph{point-to-point} ($p2p$) and \emph{point-to-plane} ($p2plane$) distance metrics, along with iterations number of the optimization framework, 
					using a 2D distance error metric (px). The highlighted setting is used for all other experiments
				}
			\label{table:evaluation_p2p_p2plane}
			\setlength{\tabcolsep}{1pt}	
			\begin{tabular}{|c|c|c|c|c|c|c|}
				\hhline{-|~|-----}
				\multicolumn{1}{|c|}{\textit{~Iterations~~}} & \multicolumn{1}{c}{} & \multicolumn{1}{|c|}{$5$} & \multicolumn{1}{c|}{\cellcolor[gray]{0.8}{$10$}} & \multicolumn{1}{c|}{$15$} & \multicolumn{1}{c|}{$20$} & \multicolumn{1}{c|}{$30$} \\
				\hhline{-|~|-----}
				\noalign{\smallskip}																		   \hhline{-|~|-----}
				\multirow{1}{*}{ $p2p$} 				& {} & {~$7.33$~} & 				{~$5.25$~} & {~$5.05$~} & {~$4.98$~} & {~$4.91$~}	\\ \hhline{-|~|-----}
				\multirow{1}{*}{\cellcolor[gray]{0.8}{~$p2plane$~}} 	& {} & { $5.33$ } & \cellcolor[gray]{0.8}	 {$5.12$~} & { $5.08$ } & { $5.07$ } & { $5.05$ }	\\ \hhline{-|~|-----}
			\end{tabular}
		\end{center}
	\end{table}

		\subsubsection{Distance Metrics}\label{subsubsec:experimentDistanceMetrics}
	
		Table \ref{table:evaluation_p2p_p2plane} presents an evaluation of the two distance metrics presented in Section \ref{subsec:m2d}, 
		namely \emph{point-to-point} \eqref{eq:errorResidual_p2p} 
		and \emph{point-to-plane} \eqref{eq:errorResidual_p2pl}, 
		along with the number of iterations of the minimization framework.
		The \emph{point-to-plane} metric leads to an adequate pose estimation error with only $10$ iterations, 
		providing a significant speed gain compared to \emph{point-to-point}. 
		\reviewChangeB{
		If the number of iterations does not matter, the \emph{point-to-point} metric is preferable since it results in a lower error and does not suffer from wrongly estimated normals. 
		}
		
		For the first frame, we perform $50$ iterations in order to ensure an accurate refinement of the manually initialized pose. 
		\reviewChangeA{  
		For the chosen setup, we measure the runtime for the sequence \textit{``Bunny''} that contains one hand and for the sequence \textit{``Crossing and Twisting''} that contains two hands. 
		For the first sequence,	the runtime is $2.82$ seconds per frame, of which $0.12$ seconds are attributed to the salient point component $\mathcal{S}$ and $0.65$ to the collision component $\mathcal{C}$. 
		For the second sequence, the runtime is $4.96$ seconds per frame, of which $0.05$ seconds are attributed to the component $\mathcal{S}$ and $0.36$ to the component $\mathcal{C}$. 
		}

	\begin{table}[t]
		\footnotesize 
		\begin{center}
			\caption{	Evaluation of the weighting parameter $\lambda$ in \eqref{eq:bipartite_MIP}, using a 2D distance error metric (px). 
					Weight $\lambda=0$ corresponds to the objective function without salient points, noted as ``$LO+\mathcal{C}$'' in Table \ref{table:switching_ON_OFF___GCPR}. 
					Both versions of $w_{s}$ described in Section \ref{subsec:salientPointDetector} are evaluated. The highlighted setting is used for all other experiments
			}
			\vspace*{-0mm}
			\label{table:bipartiteTable_VIRTUAL_mean}
			\setlength{\tabcolsep}{1pt}	
			\begin{tabular}{|c|c|c|c|c|c|c|c|c|}
				\hhline{-~-------} 
				\multicolumn{1}{|c|}{$\lambda$} & \multicolumn{1}{c}{} & \multicolumn{1}{|c}{\textit{0}} & \multicolumn{1}{|c|}{$0.3$} & $0.6$ & $0.9$ & \cellcolor[gray]{0.8}{$1.2$} & $1.5$ & $1.8$ \\				      \hhline{-~-------} 
				\noalign{\smallskip}																									      \hhline{-~-------} 
				{ $w_{s}=1$}      						& {} & \multirow{2}{*}{\centering~$5.17$~}	& {$5.17$}	& {~$5.15$~} & {~$5.14$~} & 			  {~$5.12$~} 	& {~$5.12$~} & {~$5.23$~} \\ \hhline{-|~||~|------}  
				\cellcolor[gray]{0.8}{~$w_{s}=\frac{c_{s}}{c_{thr}}$~}   	& {} & 				  {}		& {$5.14$}	& { $5.12$ } & { $5.12$ } & \cellcolor[gray]{0.8}{ $5.12$ } 	& { $5.22$ } & { $5.61$ } \\ \hhline{-~-------} 
				  
			\end{tabular}
		\end{center}
	\end{table}

	\subsubsection{Salient Point Detection - $\mathcal{S}$}\label{subsubsec:experimentSalientPointDetector}
	
		The salient point detection component 
		depends on the parameters $w_{s}$ and $\lambda$, as described in Section \ref{subsec:salientPointDetector}. 
		Table \ref{table:bipartiteTable_VIRTUAL_mean} summarizes our evaluation of the parameter $\lambda$ spanning
		a range of possible values for both cases $w_{s}=1$ and $w_{s}=\frac{c_{s}}{c_{thr}}$. 
		The differences between the two versions of $w_{s}$ is minor although the optimal range of $\lambda$ varies for the two versions. The latter is expected since $\frac{c_{s}}{c_{thr}} \geq 1$ and smaller values of $\lambda$ compensate for the mean difference to $w_{s}=1$ in \eqref{eq:bipartite_MIP}. If $\lambda=0$ all detections are classified as false positives and the salient points are not used in the objective function \eqref{eq:obj}.   
		
\begin{figure}[t]
	\captionsetup[subfigure]{}
	\centering						         
																										\hspace*{-2.5mm}
		\subfloat[subfigure 3 prcOUR][]{	\includegraphics[trim=06mm 00mm 00mm 09mm, clip=true, height=0.195 \textwidth]{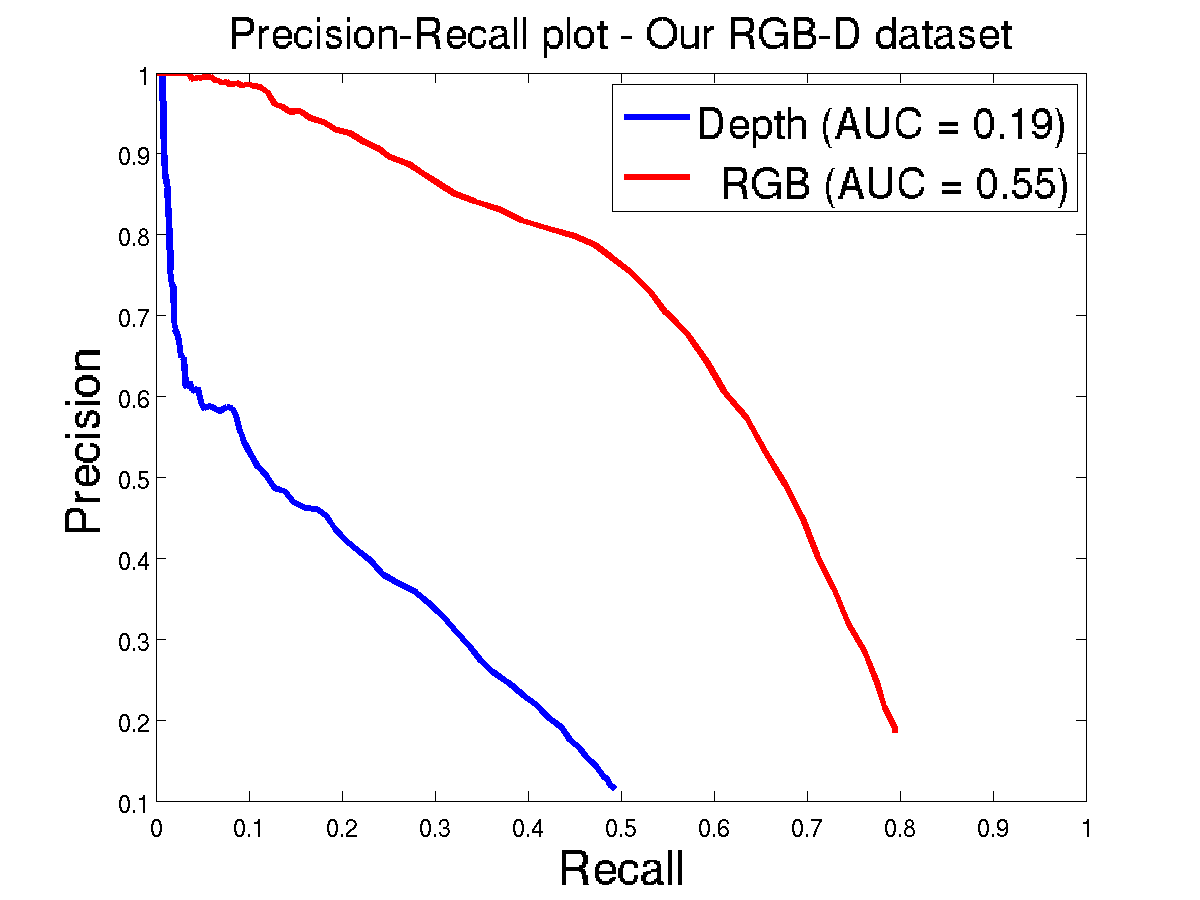}	\label{fig:prc_OUR_only}	}	\hspace*{-6.0mm}
		\subfloat[subfigure 3 prcDexter][]{	\includegraphics[trim=18mm 00mm 00mm 09mm, clip=true, height=0.195 \textwidth]{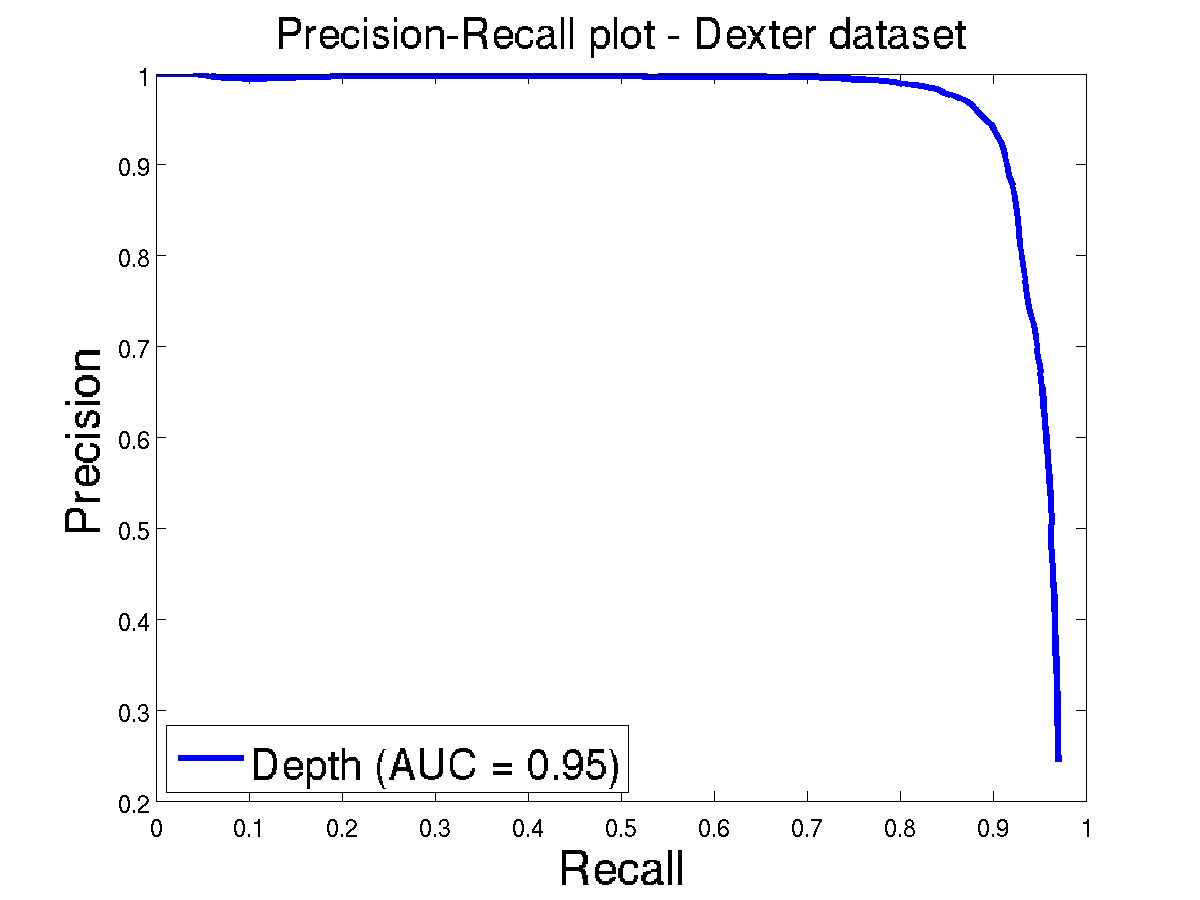}	\label{fig:prc_DEXTER_only}	}
	\caption{	
		\reviewChangeA
		{
			Precision-recall plot for (a) our RGB-D dataset and (b) the Dexter dataset. 
			We show the performance of a fingertip detector trained only on depth (blue) and only rgb (red) images. 
			The area under the curve (AUC) for our dataset (a) is $0.19$ and $0.55$ respectively.
			The AUC for the Dexter dataset (b) is $0.95$
		}
	}
	\label{fig:prc_ourRGBD_dexter}
	\end{figure}

		\reviewChangeA
		{
		To evaluate the performance of the detector, we follow the PASCAL-VOC protocol \citep{pascalChallengeVOC}. 
		Figure~\ref{fig:prc_OUR_only} shows the precision-recall plot for our RGB-D dataset including all hand-hand and hand-object sequences. The plot shows that the detector does not perform well on this dataset and suffers from the noisy raw depth data. This also explains why the salient term improves the pose estimation only slightly. We therefore trained and evaluated the detector also on the RGB data. In this case, the detection accuracy is much higher. We also evaluated the detector on the Dexter dataset~\citep{srinath_iccv2013}. On this dataset, the detector is very accurate. Our experiments on Dexter in Section~\ref{sec:evaluation_DEXTER} and a multi-camera RGB dataset in Section~\ref{sec:experimentalEvaluation___Multicamera_RGB} will show that the salient points reduce the error more if the detector performs better.  
		}

	\begin{table}[t]
		\footnotesize 
		\begin{center}
			\caption{	Evaluation of collision weights $\gamma_c$, using a 2D distance error metric (px). 
					Weight $0$ corresponds to the objective function without collision term, noted as ``$LO+\mathcal{S}$'' in Table~\ref{table:switching_ON_OFF___GCPR}. 
					Sequences are grouped in $3$ categories: 
					``\emph{severe}''	 for intense, 
					``\emph{some}''		 for light and 
					``\emph{no apparent}''	 for imperceptible collision. ``$\geq$ \emph{some}'' is the union of ``\emph{severe}'' and ``\emph{some}''. The highlighted value is the default value we use for all other experiments
				}
			\label{table:evaluationCollisionWeights}
			\setlength{\tabcolsep}{1pt}	
			\vspace*{-0mm}
			\begin{tabular}{|c|c|c|c|c|c|c|c|c|c|}
				\hhline{-|~|--------}
				\multicolumn{1}{|c|}{$\gamma_c$} & \multicolumn{1}{c}{} & \multicolumn{1}{|c|}{$0$} & \multicolumn{1}{c|}{$1$} & \multicolumn{1}{c|}{$2$} & \multicolumn{1}{c|}{$3$} & \multicolumn{1}{c|}{$5$} & \multicolumn{1}{c}{$7.5$} & \multicolumn{1}{|c|}{\cellcolor[gray]{0.8} {$10$}} & \multicolumn{1}{c|}{$12.5$} \\
				\hhline{-|~|--------}
				\noalign{\smallskip}																							   \hhline{-|~|--------}
				\multirow{1}{*}{\textit{All}} 			& {} & {$5.34$} & {~$5.44$~} & {~$5.57$~} & {~$5.16$~} & {~$5.12$~} & {~$5.12$~} & \cellcolor[gray]{0.8}	{~$5.12$~} & 	{~$5.14$~}		\\ \hhline{-|~|--------}
				\multirow{1}{*}{\textit{Severe}} 		& {} & {$5.90$} & { $6.07$ } & { $6.27$ } & { $5.62$ } & { $5.56$ } & { $5.57$ } & \cellcolor[gray]{0.8}	{ $5.55$ } & 	{~$5.61$~}		\\ \hhline{-|~|--------}
				\multirow{1}{*}{\textit{~$\geq$ Some~}} 	& {} & {$5.44$} & { $5.57$ } & { $5.72$ } & { $5.23$ } & { $5.18$ } & { $5.19$ } & \cellcolor[gray]{0.8}	{ $5.18$ } & 	{~$5.22$~}		\\ \hhline{-|~|--------}
				\multirow{1}{*}{\textit{Some}} 			& {} & {$3.99$} & { $3.98$ } & { $3.98$ } & { $3.98$ } & { $3.98$ } & { $3.99$ } & \cellcolor[gray]{0.8}	{ $3.99$ } & 	{~$3.98$~}		\\ \hhline{-|~|--------}

			\end{tabular}
		\end{center}
	\end{table}

	\begin{table}[t]
		\footnotesize 
		\begin{center}
			\caption{\reviewChangeB{Comparison of the proposed collision term based on 3D distance fields with correspondences between vertices of colliding triangles
					}
				}
			\label{table:TableDummyCollision}
			\setlength{\tabcolsep}{1pt}	
			\begin{tabular}{|l|c|c|c|}
				\hhline{-~--}
				\multicolumn{1}{|c|}{} & \multicolumn{1}{c}{} & \multicolumn{1}{|c|}{~Corresponding vertices~~} 	& \multicolumn{1}{c|}{~Distance fields~~} 	\\ 
														   \hhline{-~--}
				\noalign{\smallskip}								   \hhline{-~--}
				\multicolumn{1}{|c|}{~All}		& {} & $6.66$	&	$5.12$		\\ \hhline{-~--}
				\multicolumn{1}{|c|}{~Severe}		& {} & $7.96$	&	$5.55$		\\ \hhline{-~--}
				\multicolumn{1}{|c|}{$\ge$ Some}	& {} & $7.04$	&	$5.18$		\\ \hhline{-~--}
				\multicolumn{1}{|c|}{~Some}		& {} & $4.12$	&	$3.99$		\\ \hhline{-~--}
			\end{tabular}
		\end{center}
	\end{table}

	\subsubsection{Collision Detection - $\mathcal{C}$}\label{subsubsec:experimentCollisionDetection}
	
	The impact of the collision detection component is regulated in the objective function \eqref{eq:obj} by the weight $\gamma_c$. 
	For the evaluation, we split the sequences in three sets depending on the amount of observed collision: \emph{severe}, \emph{some}, and \emph{no apparent} collision. The set with \emph{severe} collisions comprises
	\textit{``Walking''}, \textit{``Crossing''}, \textit{``Crossing and Twisting''}, \textit{``Dancing''}, \textit{``Hugging''}, \emph{some} collisions are present in \textit{``Tips Touching''}, \textit{``Rock''}, \textit{``Bunny''}, and no collisions are apparent in \textit{``Grasping''}, \textit{``Tips Blending''}, \textit{``Flying''}.
	Table~\ref{table:evaluationCollisionWeights} summarizes our evaluation experiments for the values of $\gamma_c$. The results show that over all sequences, the collision term reduces the error and that a weight $\gamma_c \geq 3$ gives similar results. For small weights $0 < \gamma_c < 3$, the error is even slightly increased compared to $\gamma_c = 0$. In this case, the impact is too small to avoid collisions and the term only adds noise to the pose estimation. As expected, the impact of the collision term is only observed for the sequences with \emph{severe} collision.     
	
	\reviewChangeB{
	The proposed collision term is based on a fast approximation of the distance field of an object. It is continuous and less sensitive to a change of the mesh resolution than a repulsion term based on 3D-3D correspondences between vertices of colliding triangles. To show this, we replaced the collision term by correspondences that move vertices of colliding triangles towards the counterpart. The results in Table~\ref{table:TableDummyCollision} show that such a simple repulsion term performs poorly.     
	}	

	\begin{table*}[t]
		\footnotesize 
		\begin{center}
			\caption{	Evaluation of the components of our pipeline. ``$LO$'' stands for local optimization and includes fitting both \emph{data-to-model} ($d2m$) and \emph{model-to-data} ($m2d$), unless otherwise specified. 
					Collision detection is noted as ``$\mathcal{C}$'', while salient point detector is noted as ``$\mathcal{S}$''.  
					The number of sequences where the optimization framework collapses is noted in the last row, while the mean error is reported only for the rest
			}
			\label{table:switching_ON_OFF___GCPR}
			\setlength{\tabcolsep}{1pt}	
			\begin{tabular}{|c|c|c|c|c|c|c|c|c|}
					   \cline{3-4}\cline{6-9}
				\cline{1-1}\cline{3-4}\cline{6-9}
				\multicolumn{1}{|c|}{\textit{Components}} & \multicolumn{1}{c}{} & \multicolumn{1}{|c}{~~$LO_{m2d}$~~} & \multicolumn{1}{|c|}{~$LO_{d2m}$~~} & \multicolumn{1}{c}{} & \multicolumn{1}{|c|}{~~~~$LO$~~~~} & ~~$LO+\mathcal{C}$~~ & ~~$LO+\mathcal{S}$~~ & ~$LO+\mathcal{C}\mathcal{S}$~ \\
				\cline{1-1}\cline{3-4}\cline{6-9}
				\noalign{\smallskip}
				\hhline{-|~|--|~|----}  
				{\textit{Mean Error (px)}}	& \multicolumn{1}{c|}{}	& 	  { $27.17$ } 	&			     	{ $   -   $ }	& 		     	{} 	& { $5.53$ } 	& { $5.17$ } 	& { $5.34$ } 	& \cellcolor[gray]{0.8}	{ $5.12$ }		\\ \cline{1-1}\cline{3-4}\cline{6-9}
				{\textit{~Improvement (\%)~}  }	& \multicolumn{1}{c}{} 	& \multicolumn{1}{c}{} 		& \multicolumn{1}{c}	{           }	& \multicolumn{1}{c|}	{} 	& { $  -  $ } 	& { $6.46$ } 	& { $3.44$ } 	& 			{ $7.44$ }		\\ \cline{1-1}		  \cline{6-9}   
				\cline{1-1}
				\cline{6-9}
				\noalign{\smallskip}
				\cline{1-1}\cline{3-4}\cline{6-9}
			        {\textit{Failed Sequences}}	& {} & { $1/11$ } & { $11/11$ } & {} & { $  0/11 $ } & { $  0/11 $ } & { $ 0/11 $ } & 			{ ~$ 0/11 $ }\\ \cline{1-1}\cline{3-4}\cline{6-9}
				\cline{1-1}\cline{3-4}\cline{6-9}
			\end{tabular}
		\end{center}
	\end{table*}

	\begin{table*}[t]
		\footnotesize 
		\begin{center}
			\caption{Pose estimation error for each sequence
			}
		\vspace*{-3mm}
			\label{table:finalPipeline_errorPerSequence_HORIZONTAL}
			\setlength{\tabcolsep}{1pt}	
			\begin{tabular}{|c|c|c|c|c|c|c|c|c|c|c|c|c|}																																																\hhline{~~-----------}
				\multicolumn{1}{c}{} & \multicolumn{1}{c}{} & \multicolumn{1}{|c|}{\scriptsize{~Walking~~}} 	& \scriptsize{~Crossing~}	& \scriptsize{~Crossing~} 	& \scriptsize{Tips} 		& \scriptsize{~Dancing~}	& \scriptsize{Tips} 		& \scriptsize{~Hugging~}	& \scriptsize{~Grasping~}	& \scriptsize{~Flying~} 	& \scriptsize{~Rock~} 	& \scriptsize{~Bunny~}		\\ 	
				\multicolumn{1}{c}{} & \multicolumn{1}{c}{} & \multicolumn{1}{|c|}{\scriptsize{}} 		& \scriptsize{} 		& \scriptsize{Twisting} 	& \scriptsize{~Touching~} 	& \scriptsize{} 		& \scriptsize{~Blending~}	& \scriptsize{} 		& \scriptsize{} 		& \scriptsize{} 		& \scriptsize{} 	& \scriptsize{}			\\ 	\hhline{~~-----------}
				\noalign{\smallskip}																																																		\hhline{-~-----------}
				{\textit{Mean Error (px)}}				& {}	&  $5.99$	&  $4.53$	&  $4.76$	&  $3.65$	&  $6.49$	&  $4.87$	&  $5.22$	&  $4.37$	&  $5.11$	&  $4.44$	&  $4.50$																						\\ 	\hhline{-|~|-----------}
				{\textit{~Standard Deviation (px)~}}			& {}	&  $3.65$	&  $2.99$	&  $3.51$	&  $2.21$	&  $3.70$	&  $2.97$	&  $3.42$	&  $2.06$	&  $2.77$	&  $2.63$	&  $2.61$																						\\ 	\hhline{-|~|-----------}
				{\textit{Max Error (px)}}				& {}	&  ~$24.19$~	& ~$18.03$~	& ~$22.80$~	& ~$13.60$~	& ~$20.25$~	& ~$18.36$~	& ~$20.03$~	& ~$11.05$~	& ~$15.03$~	& ~$14.76$~	& ~$10.63$~																						\\ 	\hhline{-|~|-----------}

			\end{tabular}
		\end{center}
	\end{table*}

	\subsubsection{Component Evaluation}\label{subsubsec:evaluation_GCPR_ComponentEv}
	
		\begin{sloppypar}
		Table \ref{table:switching_ON_OFF___GCPR} presents the evaluation of each component and 
		the combination thereof. 
		Simplified versions of the pipeline, fitting either just the \emph{model to the data} ($LO_{m2d}$) or the \emph{data to the model} ($LO_{d2m}$) can lead to a 
		collapse of the pose estimation, due to unconstrained optimization.
		Our experiments quantitatively show the notable contribution 
		of both the collision detection and the salient point detector.
		The best overall system performance is achieved with all four components of the objective function \eqref{eq:obj}. The fifth term $E_{physics}$ is only relevant for hand-object interactions and will be evaluated in Section~\ref{sec:experimentalEvaluation___Monocular_RGBD__hand2object}. Table \ref{table:finalPipeline_errorPerSequence_HORIZONTAL} shows the error for each sequence. 
		Figure \ref{fig:My_Results_NoObject}, which is at the end of the article, depicts qualitative results for 8 out of the 11 sequences. 
		It shows that the hand motion is accurately captured even in cases of close interaction and severe occlusions. 
		The data and videos are available.\footnote{All annotated sequences are available at \url{http://files.is.tue.mpg.de/dtzionas/hand-object-capture.html}.\label{websiteFootnote}}
		\end{sloppypar}


	\subsubsection{Comparison to State-of-the-Art}\label{sec:evaluation_FORTH}

		\begin{figure}[t]
		\captionsetup[subfigure]{labelformat=empty}
		\centering
			\subfloat[subfigure 1 CompFORTH][]{	\includegraphics[width=0.15 \textwidth]{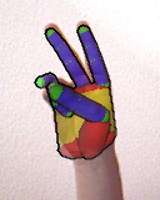}	\label{fig:comp202forth}	}		\hspace*{-1mm}
			\subfloat[subfigure 2 CompFORTH][]{	\includegraphics[width=0.15 \textwidth]{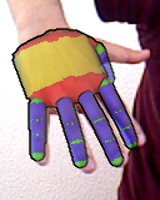}	\label{fig:comp204forth}	}		\hspace*{-1mm}
			\subfloat[subfigure 3 CompFORTH][]{	\includegraphics[width=0.15 \textwidth]{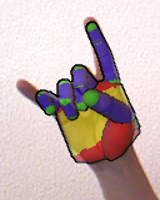}	\label{fig:comp210forth}	}	\\	\vspace*{-9mm}
			\subfloat[subfigure 4 CompFORTH][]{	\includegraphics[width=0.15 \textwidth]{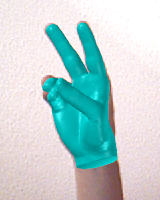}	\label{fig:comp202mine}		}		\hspace*{-1mm}
			\subfloat[subfigure 5 CompFORTH][]{	\includegraphics[width=0.15 \textwidth]{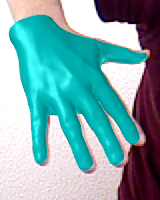}	\label{fig:comp204mine}		}		\hspace*{-1mm}
			\subfloat[subfigure 6 CompFORTH][]{	\includegraphics[width=0.15 \textwidth]{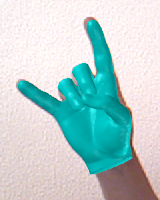}	\label{fig:comp210mine}		}
			\vspace*{-4mm}
		\caption{	Qualitative comparison with \citep{OikonomidisBMVC}. 
				Each image pair corresponds to the pose estimate of the FORTH tracker (up) and our tracker (down)
		}
		\label{fig:comparisonFORTH}
		\end{figure}

		Recently, \citet{OikonomidisBMVC,Oikonomidis_1hand_object,Oikonomidis_2hands} used particle swarm optimization (PSO) for a real-time hand tracker. 
		For comparison we use the software released for tracking a single hand \citep{OikonomidisBMVC}, with the parameter setups used also in the other works.
		Each setup is evaluated three times 
		in order to compensate for the manual initialization and the inherent randomness of PSO. 
		Qualitative results depict the best result of all three runs, while quantitative results report the average error. 
		Table~\ref{table:ComparisonFORTH} shows that our system outperforms \citep{OikonomidisBMVC} in terms of tracking accuracy. Figure \ref{fig:comparisonFORTH} shows a visual comparison.  		
		However, it should be noted that the GPU implementation of \citep{OikonomidisBMVC} runs in real time using 25 generations and 64 particles, in contrast to our single-threaded CPU implementation.

		\begin{table*}[t]
			\footnotesize 
			\begin{center}
				\caption{	Comparison with \citep{OikonomidisBMVC}. 
						We evaluate the FORTH tracker with 4 parameter settings, 
						3 of which were used in the referenced literature of the last column
					}
				\label{table:ComparisonFORTH}
				\setlength{\tabcolsep}{1pt}
				\begin{tabular}{|c|c|c|c|c|c|c|c|c|c|c|l|}
					\cline{5-7}\cline{9-10}\cline{12-12}
					\multicolumn{1}{c}{} & \multicolumn{1}{c}{} & \multicolumn{1}{c}{} & \multicolumn{1}{c|}{} & \textit{~Mean (px)~} & \textit{~St.Dev (px)~} & \multicolumn{1}{c|}{\textit{~Max (px)~}} & \multicolumn{1}{c}{} & \multicolumn{1}{|c|}{\textit{~Generations~~}} & \multicolumn{1}{c|}{\textit{~Particles~~}} & \multicolumn{1}{c}{} & \multicolumn{1}{|c|}{\textit{~Reference~}} \\
					\cline{5-7}\cline{9-10}\cline{12-12}
					\noalign{\smallskip}
					\cline{1-1}\cline{3-3}\cline{5-7}\cline{9-10}\cline{12-12}
					\multirow{4}{*}{\centering\begin{turn}{90}\textit{FORTH}\end{turn} 	 } & {} & \textit{~set 1~}	& {} & {$8.58$} & {$5.74$} & {$61.81$} & {} & {$25$} & {$64$}  & \multicolumn{1}{c|}{} & {~\cite{OikonomidisBMVC}~} 			\\ \cline{3-3}	\cline{5-7}\cline{9-10}\cline{12-12}
														{} & {} & \textit{~set 2~}	& {} & {$8.32$} & {$5.42$} & {$57.97$} & {} & {$40$} & {$64$}  & \multicolumn{1}{c|}{} & {~\cite{Oikonomidis_1hand_object}~} 	\\ \cline{3-3}	\cline{5-7}\cline{9-10}\cline{12-12}
														{} & {} & \textit{~set 3~}	& {} & {$8.09$} & {$5.00$} & {$38.90$} & {} & {$40$} & {$128$} & \multicolumn{1}{c}{}  & \multicolumn{1}{c}{} 			\\ \cline{3-3}	\cline{5-7}\cline{9-10}\cline{12-12}
														{} & {} & \textit{~set 4~}	& {} & {$8.16$} & {$5.18$} & {$39.85$} & {} & {$45$} & {$64$}  & \multicolumn{1}{c|}{} & {~\cite{Oikonomidis_2hands}~} 		\\ \cline{5-7}		   \cline{9-10}\cline{12-12}
					\cline{1-1}\cline{3-3}\cline{5-7}\cline{9-10}\cline{12-12}
					\noalign{\smallskip}
					\cline{1-3}\cline{5-7}\hhline{---|~|---}  
					\multicolumn{3}{|c|}{\textit{Proposed} } 		      			   & {} & \cellcolor[gray]{0.8}{ $3.76$ } & {$2.22$} & { $19.92$ } & \multicolumn{1}{c}{} & \multicolumn{1}{c}{} & \multicolumn{1}{c}{} & \multicolumn{1}{c}{} & \multicolumn{1}{c}{} 	\\ \cline{3-3}
					\cline{1-3}\cline{5-7}
					
				\end{tabular}
			\end{center}
		\end{table*}


\subsubsection{Dexter dataset}\label{sec:evaluation_DEXTER}

\reviewChangeA
{
	We further evaluate our approach on the recently introduced Dexter dataset \citep{srinath_iccv2013}. As suggested in \citep{srinath_iccv2013}, we use the first part of the sequences for evaluation and the second part for training.   
	More specifically, the evaluation set contains 
	the frames $018-158$ of the sequence ``adbadd'', 
	$061-185$ of ``fingercount'',
	$020-173$ of ``fingerwave'',
	$025-224$ of ``flexex1'',
	$024-148$ of ``pinch'', 
	$024-123$ of ``random'', and
	$016-166$ of ``tigergrasp''. 
	\reviewChangeB{		We use only the depth of the Time-of-Flight camera. 
	}
}

\reviewChangeA
{
	The performance of our tracker is summarized in Tables \ref{table:dexter_persequence} and \ref{table:evaluation_DEXTER__AllCombination_TurnOnOff}. Since the dataset does not provide a hand model, we simply scaled our hand model in $(x,y,z)$ direction by $(0.95,0.95,1)$. Since the annotations in the dataset do not correspond to anatomical landmarks but are close to the finger tips, we compare the annotations with the endpoints of our skeleton.      
	Table \ref{table:dexter_persequence} shows the error of our tracker for each of the sequences, reporting the mean, the maximum, and the standard deviation of the error over all the tested frames. 
	Despite of the differences of our hand model and the data, the average error is for most sequences only around $1cm$.
	\reviewChangeB{
	Our approach, however, fails for the sequence ``random'' due to the very fast motion in the sequence.}   
}

\reviewChangeA
{
	Table \ref{table:evaluation_DEXTER__AllCombination_TurnOnOff} presents the evaluation of each component of our pipeline and the combination of them. On this dataset, both the collision term as well as the salient point detector reduce the error. 
	Compared to Table~\ref{table:switching_ON_OFF___GCPR}, the error is more reduced. 
	In particular, the salient point detector reduces the error more since the detector performs well on this dataset as shown in Figure \ref{fig:prc_DEXTER_only}.
	Compared to ``$LO$'', the average error of ``$LO+SC$'' is by more than $3.5$mm lower. 
	\reviewChangeB{		The average error reported by \cite{srinath_iccv2013} on the slow part of the Dexter dataset is $13.1$ mm.
	}
}

	\begin{table}[t]
		\footnotesize 
		\begin{center}
			\caption{
				\reviewChangeA
				{
					Pose estimation error of our tracker for each sequence of the Dexter dataset. 
				}
			}
			\label{table:dexter_persequence}
			\setlength{\tabcolsep}{1pt}	
			\begin{tabular}{|c|c|c|c|c|c|c|}																									\hhline{-~---~~}
				\multicolumn{1}{|c|}{~LO + $\mathcal{S}$$\mathcal{C}$} & \multicolumn{1}{c}{} & \multicolumn{1}{|c|}{\scriptsize{Mean Error}} 	& \scriptsize{St. Deviation}	& \scriptsize{Max Error} 			\\ 	\hhline{-~---~~}
				\noalign{\smallskip}																										\hhline{-~---~-}
				{\textit{Adbadd}}			& {}	&                 $17.34$	&                $15.35$	&                $69.73$	& {} & \multirow{7}{*}{\centering\begin{turn}{90}[mm]\end{turn}}	\\ 	\hhline{-~---~~}
				{\textit{Fingercount}}			& {}	&                 $11.94$	&                $7.18$		&                $47.77$	& {} & {}								\\ 	\hhline{-~---~~}
				{\textit{Fingerwave}}			& {}	&                 $10.88$	&                $5.47$		&                $49.62$	& {} & {}								\\ 	\hhline{-~---~~}
				{\textit{Flexex1}}			& {}	&                 $11.87$	&                $12.86$	&                $91.70$	& {} & {}								\\ 	\hhline{-~---~~}
				{\textit{Pinch}}			& {}	&                 $24.19$	&                $28.34$	&                $131.97$	& {} & {}								\\ 	\hhline{-~---~~}
				{\reviewChangeB{\textit{Random}}}	& {}	&  \reviewChangeB{$96.93$}	& \reviewChangeB{$122.34$}	& \reviewChangeB{$559.37$}	& {} & {}								\\ 	\hhline{-~---~~}
				{\textit{Tigergrasp}}			& {}	&                 $11.77$	&                $5.36$		&                $30.18$	& {} & {}								\\ 	\hhline{-~---~-}
				\noalign{\smallskip}
				\noalign{\smallskip}																										\hhline{-~---~-}
				{\textit{Adbadd}}			& {}	&                 $7.79$	&                $8.38$		&                $42.54$	& {} & \multirow{7}{*}{\centering\begin{turn}{90}[px]\end{turn}}	\\ 	\hhline{-~---~~}
				{\textit{Fingercount}}			& {}	&                 $6.03$	&                $5.39$		&                $38.28$	& {} & {}								\\ 	\hhline{-~---~~}
				{\textit{Fingerwave}}			& {}	&                 $4.45$	&                $2.80$		&                $15.26$	& {} & {}								\\ 	\hhline{-~---~~}
				{\textit{Flexex1}}			& {}	&                 $5.24$	&                $8.37$		&                $61.40$	& {} & {}								\\ 	\hhline{-~---~~}
				{\textit{Pinch}}			& {}	&                 $12.56$	&                $16.48$	&                $73.16$	& {} & {}								\\ 	\hhline{-~---~~}
				{\reviewChangeB{\textit{Random}}}	& {}	&  \reviewChangeB{$59.93$}	& \reviewChangeB{$77.77$}	& \reviewChangeB{$307.00$}	& {} & {}								\\ 	\hhline{-~---~~}
				{\textit{Tigergrasp}}			& {}	&                 $6.84$	&                $4.22$		&                $21.21$	& {} & {}								\\ 	\hhline{-~---~-}
			\end{tabular}
		\end{center}
	\end{table}

	\begin{table}[t]
		\footnotesize 
		\begin{center}
			\caption{
				\reviewChangeA
				{
					Evaluation of the components of the objective function \eqref{eq:obj} on the Dexter dataset. ``$LO$'' stands for local optimization and includes fitting both \emph{data-to-model} and \emph{model-to-data}. 
					Collision detection is noted as ``$\mathcal{C}$'' and salient point detector as ``$\mathcal{S}$''. 
				}
				\reviewChangeB{		The ``random'' sequence is excluded because our approach fails due to very fast motion 
				}
			}
			\label{table:evaluation_DEXTER__AllCombination_TurnOnOff}
			\setlength{\tabcolsep}{1pt}	
			\begin{tabular}{|l|c|c|c|c|c|}
																											\hhline{-~--~~}
				\multicolumn{1}{|c|}{~Components~~} & \multicolumn{1}{c}{} & \multicolumn{1}{|c|}{~~Mean Error~~~} & \multicolumn{1}{c|}{~~St. Dev.~~~} 					\\ 	\hhline{-~--~~}
				\noalign{\smallskip}																					\hhline{-~--~-}
	\cellcolor[gray]{0.8}	 {LO + $\mathcal{S}$$\mathcal{C}$}			& {} &		$14.26$		&	$14.91$	& {} & \multirow{4}{*}{\centering\begin{turn}{90}[mm]\end{turn}}	\\ 	\hhline{-~--~~}
				~{LO + $\mathcal{S}$}					& {} &		$15.51$		&	$16.67$	& {} & {}								\\ 	\hhline{-~--~~}
				~{LO + \hspace{1.7mm}$\mathcal{C}$}			& {} &		$16.97$		&	$16.60$	& {} & {}								\\ 	\hhline{-~--~~}
				~{LO}							& {} &		$17.86$		&	$18.80$	& {} & {}								\\ 	\hhline{-~--~-}
				\noalign{\smallskip}																					\hhline{-~--~-}
	\cellcolor[gray]{0.8}	 {LO + $\mathcal{S}$$\mathcal{C}$}			& {} &		$6.90$		&	$8.88$	& {} & \multirow{4}{*}{\centering\begin{turn}{90}[px]\end{turn}}	\\ 	\hhline{-~--~~}
				~{LO + $\mathcal{S}$}					& {} &		$7.64$		&	$9.87$	& {} & {}								\\ 	\hhline{-~--~~}
				~{LO + \hspace{1.7mm}$\mathcal{C}$}			& {} &		$8.98$		&	$10.29$	& {} & {}								\\ 	\hhline{-~--~~}
				~{LO}							& {} &		$9.33$		&	$10.73$	& {} & {}								\\ 	\hhline{-~--~-}
			\end{tabular}
		\end{center}
	\end{table}

	\begin{figure}[t]
	\captionsetup[subfigure]{labelformat=empty}
	\centering								 
		\subfloat[subfigure 1 failureDetectorNoo1][]{	\includegraphics[trim=034mm 024mm 000mm 006mm, clip=true, width=0.23\textwidth]{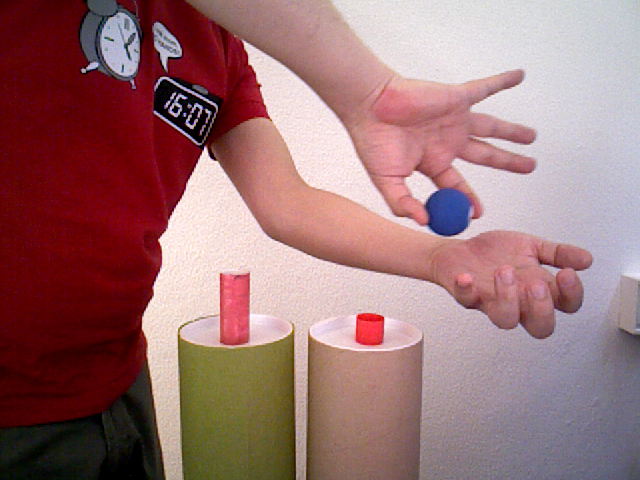}	\label{fig:failure_detectionXXX_inputRGB}	}		\hspace*{-01.0mm}
		\subfloat[subfigure 2 failureDetectorNoo2][]{	\includegraphics[trim=034mm 024mm 000mm 006mm, clip=true, width=0.23\textwidth]{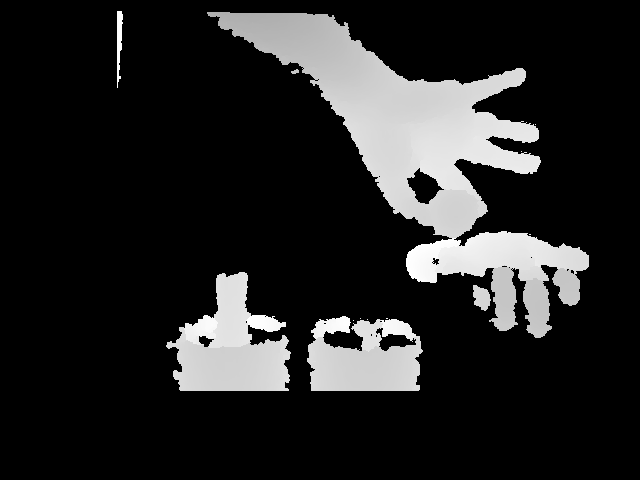}	\label{fig:failure_detectionXXX_inputDDD}	}	\\	\vspace*{-06.7mm}
		\subfloat[subfigure 4 failureDetectorNoo2][]{	\includegraphics[trim=034mm 024mm 000mm 006mm, clip=true, width=0.23\textwidth]{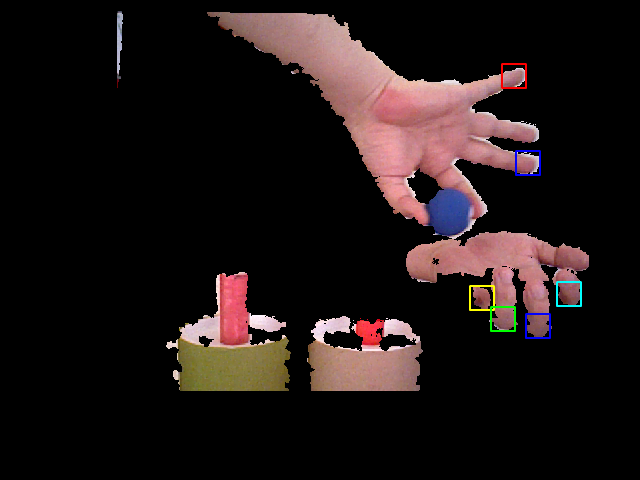}	\label{fig:failure_detectionYES_inputRGBD}	}
		\subfloat[subfigure 6 failureDetectorYes2][]{	\includegraphics[trim=034mm 024mm 000mm 006mm, clip=true, width=0.23\textwidth]{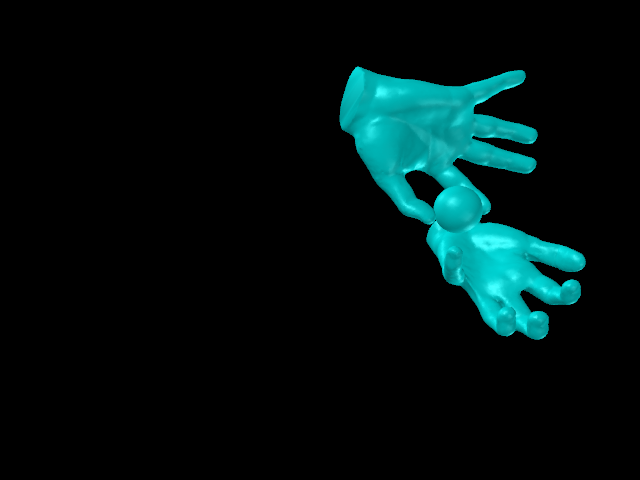}	\label{fig:failure_detectionYES_outputSYNTH}	}
		\vspace*{-0mm}
	\caption{	
		\reviewChangeA
		{
			Failure case due to missing data and detection errors. 
			The images show RGB image (top-left), input depth image (top-right), 
			fingertip detections (bottom-left), and estimated pose (bottom-right). 
			The detector operates on the raw depth image, while the RGB image is used just for visualization
		}
	}
	\label{fig:failure_case}
	\end{figure}

	\subsection{Monocular RGBD - Hand-Object Interactions}\label{sec:experimentalEvaluation___Monocular_RGBD__hand2object}

	\begin{sloppypar}
	For the evaluation of the complete energy function \eqref{eq:obj} for hand-object interactions, we captured $7$ new sequences\textsuperscript{\ref{websiteFootnote}} 
	of hands interacting with several objects, either rigid (\emph{ball}, \emph{cube)} or articulated (\emph{pipe}, \emph{rope}). The DoF of the objects varies a lot. The rigid objects have $6$ DoF, the \textit{pipe} $7$ DoF, and the \textit{rope} $76$ DoF.	
	The sequences vary from $180$ to $400$ frames and contain several actions, like:
	\emph{``Moving a Ball''} with one ($43$ DoF) or two hands ($80$ DoF), 
	\emph{``Moving a Cube''} with one hand ($43$ DoF), 
	\emph{``Bending a Pipe''} with two hands ($81$ DoF), and 
	\emph{``Bending a Rope''} with two hands ($150$ DoF). 
	In addition, the sequences \emph{``Moving a Ball''} with one hand and \emph{``Moving a Cube''} were captured twice, one with occlusion of a manipulating finger and one without.  
	Manual ground-truth annotation was performed by a single subject. 
	\end{sloppypar}
	
	For the salient point ($\mathcal{S}$) and the collision detection component ($\mathcal{C}$), we use the parameter setup presented in Section \ref{sec:experimentalEvaluation___Monocular_RGBD__hand2hand}. 
	The influence of the physics simulation component ($\mathcal{P}$) and its parameters are evaluated in the following section. 
	The error metric used is the 2D distance (pixel units) between the projection of the 3D joints and the 
	2D annotations as in Section \ref{sec:experimentalEvaluation___Monocular_RGBD__hand2hand} and visualized in Figure \ref{fig:jointsErrorMetric}.  
	Unless otherwise stated, we report the average over all frames of all seven sequences. 
	
	\subsubsection{Physics Simulation - $\mathcal{P}$}\label{subsec:physicsSimulation}
	
	For the physics simulation, we model the entire scene, which includes the hands as well as manipulated and static objects, with a low resolution representation as described in Section \ref{subsec:physicsComponent} and visualized in Figure \ref{fig:physicsHulls_dummyFront}. 
	Each component of the scene is characterized by three properties: friction, restitution, and mass. Since in each simulation step we consider each component except of the manipulated object as static, only the mass of the object is relevant, which we set to $1$ kg. We set the restitution of the static scene and hands to $0$ and of the object to $0.5$. 
	For the static scene, we use a friction value of $3$. The friction for both the hand and the object are assumed to be equal. Since the main purpose of the physics simulation is to evaluate if the current pose estimates are physical stable, the exact values for friction, restitution, and mass are not crucial.      
	To demonstrate this, we evaluate the impact of the friction value for hands and manipulated objects. 	
	For this experiment, we set the weight $\gamma_{ph}$ equal to $10.0$, being the same as the weight $\gamma_{c}$ of the complementary collision detection component. 
	The results presented in Table \ref{table:evaluation_newSeq_AllSeqAllFrames_FRICTION} show that the actual value of friction has no significant impact on the pose estimation error as long as it is in a reasonable range.

	\begin{table}[t]
		\footnotesize 
		\begin{center}
			\caption{	Evaluation of the friction value of both the hands and the object. 
					We report the error over all the frames of all seven sequences with hand-object interactions using a 2d error metric (px). 
					Value $3.0$ is the same as the friction value of the static scene. The highlighted value is the default value we use for all other experiments
				}
			\label{table:evaluation_newSeq_AllSeqAllFrames_FRICTION}
			\setlength{\tabcolsep}{1pt}
			\vspace*{-0mm}	
			\begin{tabular}{|c|c|c|c|c|c|c|c|c|}
				\hhline{-~-----~~}
				\multicolumn{1}{|c|}{Friction} & \multicolumn{1}{c}{} & \multicolumn{1}{|c|}{~~$0.6$~~} & ~~$0.9$~~ & \cellcolor[gray]{0.8}~~$1.2$~~ & ~~$1.5$~~ & ~~$3.0$~~ & \multicolumn{1}{c}{} & \multicolumn{1}{c}{} \\
				\hhline{-~-----~~}
				\noalign{\smallskip}																															   \hhline{-~-----~-}
				{~Mean~}		& {} &		$6.19$		&	$6.18$		&	\cellcolor[gray]{0.8}$6.19$		&	$6.17$		&	$6.17$	& \multicolumn{1}{c|}{}    & \multirow{2}{*}{\centering\begin{turn}{90}[px]~~\end{turn}}{}	\\ \hhline{-~-----~~}
				{~St. Dev.~}		& {} &		$3.82$		&	$3.81$		&	\cellcolor[gray]{0.8}$3.81$		&	$3.81$		&	$3.81$	& \multicolumn{1}{c|}{}    & {}									\\ \hhline{-~-----~-}
			\end{tabular}
			
		\end{center}
	\end{table}

	\begin{table}[t]
		\footnotesize 
		\begin{center}
			\caption{	Evaluation of collision weights $\gamma_{ph}$ for ``$LO+\mathcal{S}\mathcal{C}\mathcal{P}$'', using a 2D distance error metric (px). 
					Weight $0$ corresponds to the objective function without physics term, noted as ``$LO+\mathcal{S}\mathcal{C}$'' in Table~\ref{table:evaluation_newSeq_IJCV__AllCombination_TurnOnOff}. 
					Sequences are grouped in $3$ categories: 
					``\emph{severe}''	 for intense, 
					``\emph{some}''		 for light and 
					``\emph{no apparent}''	 for imperceptible occlusion of manipulating fingers. ``$\geq$ \emph{some}'' is the union of ``\emph{severe}'' and ``\emph{some}''. The highlighted value is the default value we use for all other experiments
			}
			\label{table:evaluationPhysicsWeights}
			\setlength{\tabcolsep}{1pt}	
			\begin{tabular}{|c|c|c|c|c|c|c|c|c|c|}
				\hhline{-~--------}
				\multicolumn{1}{|c|}{$\gamma_{ph}$} & \multicolumn{1}{c}{}  & \multicolumn{1}{|c|}{$0$} & \multicolumn{1}{c|}{$1$} & \multicolumn{1}{c|}{$2$} & \multicolumn{1}{c|}{$3$} & \multicolumn{1}{c|}{$5$} & \multicolumn{1}{c}{$7.5$} & \multicolumn{1}{|c|}{\cellcolor[gray]{0.8}$10$} & \multicolumn{1}{c|}{$12.5$} \\
				\hhline{-~--------}
				\noalign{\smallskip}																					   \hhline{-~--------}
				\multirow{1}{*}{\textit{All}} 		& {} & {~$6.21$~} & {~$6.20$~} & {~$6.21$~} & {~$6.19$~} & {~$6.19$~} & {~$6.18$~} & \cellcolor[gray]{0.8} {~$6.19$~}	& {~$6.17$~}	\\ \hhline{-~--------}
				\multirow{1}{*}{\textit{Severe}} 	& {} & { $5.68$ } & { $5.66$ } & { $5.65$ } & { $5.63$ } & { $5.63$ } & { $5.63$ } & \cellcolor[gray]{0.8} {~$5.62$~}	& { $5.61$ }	\\ \hhline{-~--------}
				\multirow{1}{*}{\textit{$\geq$ Some}}	& {} & { $6.02$ } & { $6.00$ } & { $6.00$ } & { $5.98$ } & { $5.98$ } & { $5.97$ } & \cellcolor[gray]{0.8} {~$5.96$~}	& { $5.94$ }	\\ \hhline{-~--------}
			\end{tabular}
		\end{center}
	\end{table}

	The impact of the physics simulation component $E_{physics}$ in the objective function \eqref{eq:obj} is regulated by the weight $\gamma_{ph}$. The term penalizes implausible manipulation or grasping poses. 
	For the evaluation, we split the sequences in three sets depending on the amount of occlusions of the manipulating fingers: ``\emph{severe}'' for intense (``Moving a Cube'' with one hand and occlusion), 
	``\emph{some}'' for light (``Moving a Ball'' with one hand and occlusion, ``Moving a Cube'' with one hand) and 
	``\emph{no apparent}'' for imperceptible occlusions (\emph{``Moving a Ball''} with one  and two hands, \emph{``Bending a Pipe''}, \emph{``Bending a Rope''}). 
	Table \ref{table:evaluationPhysicsWeights} summarizes the pose estimation error for various values of $\gamma_{ph}$ for the three subsets. 
	Although the pose estimation error is only slightly reduced by $E_{physics}$, the results are physically more plausible. 
	This is shown in Figure~\ref{fig:physicsDIFF} at the end of the article, which provides a qualitative comparison between the setups ``{LO + $\mathcal{S}$$\mathcal{C}$}'' and ``{LO + $\mathcal{S}$$\mathcal{C}$$\mathcal{P}$}''.
	The images show the notable contribution of component $\mathcal{P}$ towards more realistic, physically plausible poses, 
	especially in cases of missing or ambiguous visual data, as in sequences with an occluded manipulating finger. 	
	\reviewChangeA{
	To quantify this, we run the simulation for $35$ iterations with a time-step of $0.1$ seconds after the pose estimation and measured the displacement of the centroid of the object for each frame. While the average displacement is $9.26$mm for the setup ``{LO + $\mathcal{S}$$\mathcal{C}$}'', the displacement is reduced to $9.05$mm            
	by the setup ``{LO + $\mathcal{S}$$\mathcal{C}$$\mathcal{P}$}''. 
	}
	The tracking runtime for the aforementioned sequences for the setup ``{LO + $\mathcal{S}$$\mathcal{C}$}'' ranges from $4$ to $8$ seconds per frame. 
	The addition of $\mathcal{P}$ in the setup ``{LO + $\mathcal{S}$$\mathcal{C}$$\mathcal{P}$}'' increases the runtime for most sequences for about $1$ second. 
	However, this increase might reach up to more than $1$ minute depending on the complexity of the object and tightness of interaction, 
	as in the case of \emph{``Bending a Pipe''} with two hands ($150$ DoF), 
	\reviewChangeA{
		with the main bottleneck being the computation of the closest finger vertices to the manipulated object. 
	}
	Figure \ref{fig:My_Results_WithObject} depicts qualitative results for the full setup ``{LO + $\mathcal{S}$$\mathcal{C}$$\mathcal{P}$}'' of the objective function \eqref{eq:obj} 
	for all seven sequences. 
	The results show successful tracking of interacting hands with both rigid and articulated objects, whose articulation is described from $1$ to as many as $71$ DoF.

	\begin{table}[t]
		\footnotesize 
		\begin{center}
			\caption{	Evaluation of the components of the objective function \eqref{eq:obj}. ``$LO$'' stands for local optimization and includes fitting both \emph{data-to-model} and \emph{model-to-data}. 
					Collision detection is noted as ``$\mathcal{C}$'', salient point detector as ``$\mathcal{S}$'' and physics simulation as ``$\mathcal{P}$''.
					\reviewChangeB{We report the error for fixed $10$ iterations and for the stopping criterion $\epsilon < 0.2 mm$ 
					}
				}
			\label{table:evaluation_newSeq_IJCV__AllCombination_TurnOnOff}
			\setlength{\tabcolsep}{1pt}	
			\begin{tabular}{|l|c|c|c|c|c|c|}
				\hhline{~~--~--}
				\multicolumn{1}{c}{}					& {} &		\multicolumn{2}{c|}{Fixed $10$}	& {} &	\multicolumn{2}{c|}{\reviewChangeB{Stopping Thresh.}}	\\ 
				\multicolumn{1}{c}{}					& {} &		\multicolumn{2}{c|}{Iterations}	& {} &	\multicolumn{2}{c|}{\reviewChangeB{$0.2~mm$}}		\\ \hhline{~~--~--}
				\noalign{\smallskip}
				\hhline{-~--~--}
				\multicolumn{1}{|c|}{~Components~~} & \multicolumn{1}{c}{} & \multicolumn{1}{|c|}{~Mean~~} & \multicolumn{1}{c|}{~St. Dev.~~} & \multicolumn{1}{c}{} & \multicolumn{1}{|c|}{\reviewChangeB{~Mean~~}} & \multicolumn{1}{c|}{\reviewChangeB{~St. Dev.~~}} \\ 
				\hhline{-~--~--}
				\noalign{\smallskip}												   										   \hhline{-~--~--}
				{~LO + $\mathcal{S}$$\mathcal{C}$$\mathcal{P}$}	& {} &		$6.19$		&	$3.81$	& {} &		\reviewChangeB{$6.25$}		&	\reviewChangeB{$3.86$}		\\ \hhline{-~--~--}
				{~LO + $\mathcal{S}$$\mathcal{C}$}			& {} &		$6.21$		&	$3.82$	& {} &		\reviewChangeB{$6.31$}		&	\reviewChangeB{$3.89$}		\\ \hhline{-~--~--}
				\noalign{\smallskip}												   										   \hhline{-~--~--}
				{~LO + $\mathcal{S}$}					& {} &		$6.05$		&	$3.76$	& {} &		\reviewChangeB{$6.09$}		&	\reviewChangeB{$3.77$}		\\ \hhline{-~--~--}
				{~LO + \hspace{2mm}$\mathcal{C}$$\mathcal{P}$}		& {} &		$6.19$		&	$3.83$	& {} &		\reviewChangeB{$6.31$}		&	\reviewChangeB{$3.90$}		\\ \hhline{-~--~--}
				{~LO + \hspace{2mm}$\mathcal{C}$}			& {} &		$6.24$		&	$3.84$	& {} &		\reviewChangeB{$6.38$}		&	\reviewChangeB{$3.94$}		\\ \hhline{-~--~--}
				{~LO}							& {} &		$6.07$		&	$3.77$	& {} &		\reviewChangeB{$6.15$}		&	\reviewChangeB{$3.83$}		\\ \hhline{-~--~--}
				\noalign{\smallskip}												   										   \hhline{~~--~--}
				\multicolumn{1}{c}{}					& {} &		\multicolumn{2}{c|}{px}	& {} &		\multicolumn{2}{c|}{\reviewChangeB{px}}					\\ \hhline{~~--~--}
				
			\end{tabular}
		\end{center}
	\end{table}

	\begin{figure}[ht!]
	\captionsetup[subfigure]{}
	\centering						         
																																                                       \hspace*{-5.5mm}
		\subfloat[subfigure stopping1 bars][]{	\includegraphics[trim=05mm 00mm 05mm 09mm, clip=true, height=0.196 \textwidth]{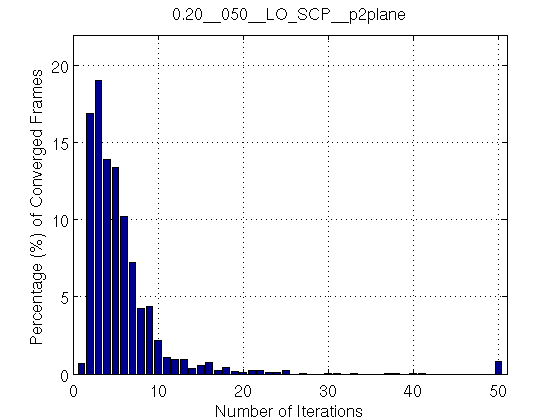}	\label{fig:stoppingCriterionBars_LO_SCP}	}	\hspace*{-4.5mm}
		\subfloat[subfigure stopping2 accc][]{	\includegraphics[trim=05mm 00mm 05mm 09mm, clip=true, height=0.196 \textwidth]{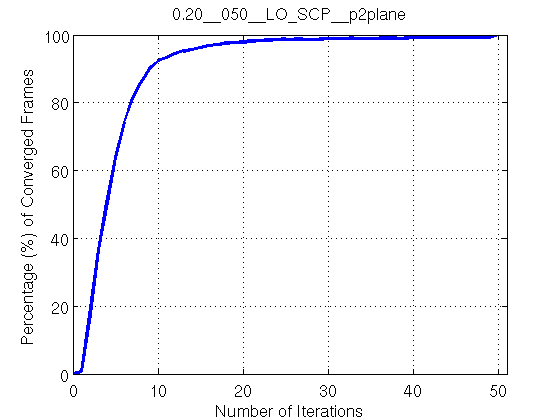}	\label{fig:stoppingCriterionAccum_LO_SCP}	}
																															      \\       \vspace*{-2mm}                  \hspace*{-5.5mm}
		\subfloat[subfigure stopping3 bars][]{	\includegraphics[trim=05mm 00mm 05mm 09mm, clip=true, height=0.196 \textwidth]{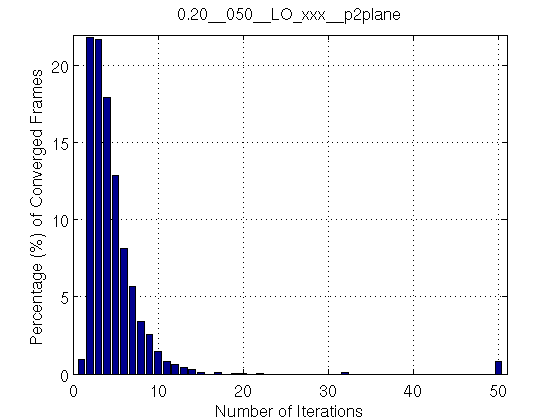}	\label{fig:stoppingCriterionBars_LO_xxx}	}	\hspace*{-4.5mm}
		\subfloat[subfigure stopping4 accc][]{	\includegraphics[trim=05mm 00mm 05mm 09mm, clip=true, height=0.196 \textwidth]{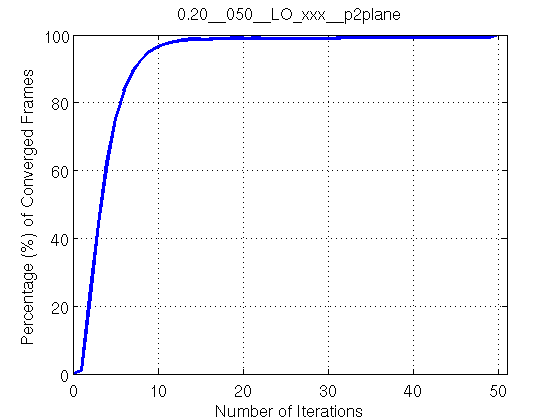}	\label{fig:stoppingCriterionAccum_LO_xxx}	}
		\caption{	
			\reviewChangeB
			{
			Number of iterations that are required to converge for {LO + $\mathcal{S}$$\mathcal{C}$$\mathcal{P}$} (top) and LO (bottom). (a,c) Distribution of frames where the pose estimation converged after a given number of iterations. (b,d) Cumulative distribution
			}\label{fig:iteration}
		}
	\end{figure}

		\subsubsection{Component Evaluation}\label{subsubsec:evaluation_IJCV_ComponentEv}
		
		Table \ref{table:evaluation_newSeq_IJCV__AllCombination_TurnOnOff} presents the evaluation 
		of each component and their combinations for the seven sequences with hand-object interaction. 
		Since the physical simulations $\mathcal{P}$ assumes that there are no severe intersections, it is meaningful only as a complement to the collision component $\mathcal{C}$. 
		One can observe that the differences between the components are relatively small since the hand poses in the hand-object sequences are in general simpler than the poses in the sequences with tight hand-hand interactions as considered in Section \ref{sec:experimentalEvaluation___Monocular_RGBD__hand2hand}.
		The collision term $\mathcal{C}$ slightly increases the error, but without the term the hand poses are often physically implausible and intersect with the object.  
		When comparing {LO + $\mathcal{S}$$\mathcal{C}$} and {LO + $\mathcal{S}$$\mathcal{C}$$\mathcal{P}$}, we see that the error is slightly reduced by  
		the physics simulation component $\mathcal{P}$. The pose estimation errors for each sequence using {LO + $\mathcal{S}$$\mathcal{C}$$\mathcal{P}$} are summarized in Table~\ref{table:evaluation_newSeq__LO_SCP__AllSew_combinedAndSeparate}.
		
		\reviewChangeB{
		Instead of using a fixed number of iterations per frame, a stopping criterion can be used. We use the average change of the joint positions after each iteration. As threshold, we use $0.2 mm$ and a maximum of $50$ iterations. Table~\ref{table:evaluation_newSeq_IJCV__AllCombination_TurnOnOff} shows that for the stopping criterion the impact of the terms is slightly more prominent, but it also shows that the error is slightly higher for all approaches. To analyze this more in detail, we report the distribution of required iterations until the stopping criterion is reached in Figure~\ref{fig:iteration}. Although {LO + $\mathcal{S}$$\mathcal{C}$$\mathcal{P}$} requires a few more iterations until convergence compared to LO, it converges in $10$ or less iterations in $92\%$ of the frames, which supports our previous results. There are, however, very few frames where the approach has not converged after 50 iterations. In most of these cases, the local optimum of the energy is far away from the true pose and the 
		error is increased with more iterations. These outliers are also the reason for the slight increase of the error in Table~\ref{table:evaluation_newSeq_IJCV__AllCombination_TurnOnOff}. For all combinations from LO to {LO + $\mathcal{S}$$\mathcal{C}$$\mathcal{P}$} we observed this behavior, which shows that the energy can be further improved.}

	\begin{table*}[t]
		\footnotesize 
		\begin{center}
			\caption{Pose estimation error for each sequence
			}
			\label{table:evaluation_newSeq__LO_SCP__AllSew_combinedAndSeparate}
			\setlength{\tabcolsep}{1pt}	
			\begin{tabular}{|c|c|c|c|c|c|c|c|c|c|c|}																																	\hhline{~~-------~~}
				\multicolumn{1}{c}{} & \multicolumn{1}{c}{} & \multicolumn{1}{|c|}{\scriptsize{~\textit{Moving Ball}~}} 	& \scriptsize{~\textit{Moving Ball}~} 	& \scriptsize{~\textit{Bending Pipe}~} & \scriptsize{~\textit{Bending Rope}~} & \scriptsize{~\textit{Moving Ball}~} 	& \scriptsize{~\textit{Moving Cube}~} 	& \scriptsize{~\textit{Moving Cube}~}	& \multicolumn{1}{c}{}	& \multicolumn{1}{c}{} 	\\ 
				\multicolumn{1}{c}{} & \multicolumn{1}{c}{} & \multicolumn{1}{|c|}{\scriptsize{$1$ hand}} 			& \scriptsize{$2$ hands} 		&  					& 					& \scriptsize{~$1$ hand, occlusion~} 	& \scriptsize{$1$ hand} 		& \scriptsize{~$1$ hand, occlusion~}	& \multicolumn{1}{c}{}	& \multicolumn{1}{c}{} 	\\ 
				\hhline{~~-------}
				\noalign{\smallskip}																																			\hhline{-~-------~-}
				{\textit{\scriptsize{Mean Error}}}		& {}	& $6.10$	&	$7.15$	&	$6.09$		&	$5.65$	&	$8.03$		&	$4.68$	&	$5.55$	& \multicolumn{1}{c|}{}	& \multirow{2}{*}{\centering\begin{turn}{90}[px]~~\end{turn}}{}		\\ 	\hhline{-~-------~~}
				{\textit{\scriptsize{Standard Deviation}}}	& {}	&  $3.90$	&	$4.82$	&	$3.07$		&	$3.04$	&	$5.47$		&	$2.61$	&	$3.28$	& {}			& {}										\\ 	\hhline{-~-------~-}
			\end{tabular}   
		\end{center}
	\end{table*}


\subsection{Limitations}\label{sec:experimentalEvaluation___Monocular_RGBD__failureCases}

	\reviewChangeA
	{
	As shown in Sections \ref{sec:experimentalEvaluation___Monocular_RGBD__hand2hand} and \ref{sec:experimentalEvaluation___Monocular_RGBD__hand2object}, our approach captures accurately the motion of hands tightly interacting either with each other or with a rigid or articulated object. However, for very fast motion like the ``random'' sequence of the Dexter dataset our approach fails. Furthermore, we assume that a hand model is given or can be acquired by an approach like \citep{MSR_handShapeAdaptation}. Figure~\ref{fig:failure_case} also visualizes an inaccurate hand pose of the lower hand due to missing depth data and two detections, which are not at the finger tips but located at other bones.       
	}

	\subsection{Multicamera RGB}\label{sec:experimentalEvaluation___Multicamera_RGB}

		\begin{sloppypar}
		We finally evaluated the approach for sequences captured using a setup of $8$ synchronized cameras recording FullHD footage at $50$ fps.
		To this end, we recorded $9$ sequences that
		span a variety of hand-hand and hand-object interactions, namely:
		\textit{``Praying''},
		\textit{``Fingertips Touching''},
		\textit{``Fingertips Crossing''},
		\textit{``Fingers Crossing and Twisting''},
		\textit{``Fingers Folding},
		\textit{``Fingers Walking''} on the back of the hand,
		\textit{``Holding and Passing a Ball''},
		\textit{``Paper Folding''} and 
		\textit{``Rope Folding''}.
		The length of the sequences varies from $180$ to $1500$ frames.
		\end{sloppypar}

		\begin{sloppypar}
		Figure~\ref{fig:Luca_Results} shows one frame from each of the tested sequences and
		the obtained results overlayed on the original frames from two different cameras.
		Visual inspection reveals that the proposed algorithm works also quite well for multiple RGB cameras even
		in challenging scenarios of very closely interacting hands with multiple occlusions. The data and videos are available.\footnote{\url{http://files.is.tue.mpg.de/dtzionas/hand-object-capture.html}}
		\end{sloppypar}

\subsubsection{Component Evaluation}

		As for the RGB-D sequences, we also evaluate the components of our approach. To this end,     
		we synthesized two sequences: first, fingers crossing and folding,
		and second, holding and passing a ball, both similar to the ones captured
		in the real scenario. Videos were generated using a commercial rendering software.
		The pose estimation accuracy was then evaluated both in terms of error in the
		joints position, and in terms of error in the bones orientation.
		
				\begin{table}[t]
			\footnotesize 
			\begin{center}
				\caption{	Quantitative evaluation of the algorithm performance with respect to the used visual features: edges $\mathcal{E}$,
						collisions $\mathcal{C}$, optical flow $\mathcal{O}$, and salient points $\mathcal{S}$.
						LO stands for our local optimization approach, while HOPE64 and HOPE128 stand for
						our implementation of~\citep{OikonomidisBMVC} with $64$ and $128$ particles respectively,
						evaluated over $40$ generations.
				}
				\label{fig:QResults}
				\setlength{\tabcolsep}{1pt}	
				\begin{tabular}{|l|c|c|c|c|c|c|}																														\hhline{-~---~~}
				
					\multicolumn{1}{|c|}{~~Used features~~} & 
					\multicolumn{1}{ c|}{} 	& 
					~~Mean~	& 
					~St.Dev.~& 
					~~Max~~	& 
					\multicolumn{1}{c}{}& 
					\multicolumn{1}{c}{} 																														\\	\hhline{-~---~~}
					\noalign{\smallskip}																															\hhline{-~---~-}
								{~~~LO      + $\mathcal{E}$~~}						& {}	& 				 {~$3.11$ }	& { $4.52$ }	& {~$49.86$~} & {} & \multirow{6}{*}{\centering\begin{turn}{90}[mm]\end{turn}}	\\	\hhline{-~---~~}
								{~~~LO      + $\mathcal{E}\mathcal{C}$~~}				& {}	& 				 {~$2.50$~}	& {~$2.89$~}	& {~$52.94$~} & {} & {} 								\\	\hhline{-~---~~}
								{~~~LO      + $\mathcal{E}\mathcal{C}\mathcal{O}$~~}			& {}	& 				 {~$2.38$ }	& { $2.25$ }	& { $16.84$ } & {} & {} 								\\	\hhline{-~---~~}
					\cellcolor[gray]{0.8}	{~~LO       + $\mathcal{E}\mathcal{C}\mathcal{O}\mathcal{S}$~~}	& {}	& 	 	 		 { $1.49$ }	& { $1.44$ }	& {~$13.27$~} & {} & {} 								\\	\hhline{-~---~~}
								{~~~HOPE64  + $\mathcal{E}\mathcal{C}\mathcal{O}\mathcal{S}$~~}	& {}	& 				 {~$4.86$~}	& {~$3.69$~}	& {~$31.05$~} & {} & {} 								\\	\hhline{-~---~~}
								{~~~HOPE128 + $\mathcal{E}\mathcal{C}\mathcal{O}\mathcal{S}$~~}	& {}	& 				 {~$4.67$ }	& { $3.28$ }	& { $41.11$ } & {} & {}  								\\	\hhline{-~---~-}
					\noalign{\smallskip}\noalign{\smallskip}\noalign{\smallskip}																										\hhline{-~---~~}
					\multicolumn{1}{|c|}{~~Used features~~} & 
					\multicolumn{1}{ c|}{} 	& 
					~~Mean~	& 
					~St.Dev.~& 
					~~Max~~	& 
					\multicolumn{1}{c}{}& 
					\multicolumn{1}{c}{} 																														\\	\hhline{-~---~~}
					\noalign{\smallskip}																															\hhline{-~---~-}
								{~~~LO      + $\mathcal{E}$~~}						& {}	& 				 {~$2.36$~}	& { $6.84$ }	& { $94.58$ } & {} & \multirow{6}{*}{\centering\begin{turn}{90}[deg]\end{turn}}	\\	\hhline{-~---~~}
								{~~~LO      + $\mathcal{E}\mathcal{C}$~~}				& {}	& 				 {~$1.98$~}	& { $4.57$ }	& { $91.89$ } & {} & {}									\\	\hhline{-~---~~}
								{~~~LO      + $\mathcal{E}\mathcal{C}\mathcal{O}$~~}			& {}	& 				 { $1.84$ }	& { $3.81$ }	& { $60.09$ } & {} & {}									\\	\hhline{-~---~~}
					\cellcolor[gray]{0.8}	{~~LO       + $\mathcal{E}\mathcal{C}\mathcal{O}\mathcal{S}$~~}	& {}	& 	 			 {~$1.88$~}	& { $3.90$ }	& { $44.51$ } & {} & {}									\\	\hhline{-~---~~}
								{~~~HOPE64  + $\mathcal{E}\mathcal{C}\mathcal{O}\mathcal{S}$~~}	& {}	& 				 {~$4.35$~}	& { $7.11$ }	& { $58.61$ } & {} & {}									\\	\hhline{-~---~~}
								{~~~HOPE128 + $\mathcal{E}\mathcal{C}\mathcal{O}\mathcal{S}$~~}	& {}	& 				 { $4.73$ }	& { $7.46$ }	& { $78.65$ } & {} & {}									\\	\hhline{-~---~-}
				\end{tabular}
			\end{center}
		\end{table}

		Table~\ref{fig:QResults} shows a quantitative evaluation of the algorithm performance
		with respect to the used visual features. 
		It can be noted that each feature contributes to the accuracy
		of the algorithm and 
		that the salient points $\mathcal{S}$ clearly boost its performance. 
		\reviewChangeA{		
		The benefit of the salient points is larger than for the RGB-D sequences since the localization of the finger tips from several high-resolution RGB cameras is more accurate than from a monocular depth camera with lower resolution. This is also indicated by the precision-recall curves in Figure \ref{fig:prc_OUR_only}. 
		}

		We also compared with~\citep{OikonomidisBMVC} on the synthetic data where we used an own implementation since the publicly available source code requires a single RGB-D sequence.  
		We also added the salient points term and used two settings,  
		namely $64$ and $128$ particles over $40$ generations. The results in Table~\ref{fig:QResults} show that our approach estimates the pose with a lower error and confirm the results for the RGB-D sequences reported in Table~\ref{table:ComparisonFORTH}.

		\begin{figure}[t]
		\begin{center}
		\begin{tabular}{c c}			
		\hspace*{-07.5mm}	\includegraphics[trim=00mm 000mm 00mm 00mm, clip=true, height=0.20\textwidth]{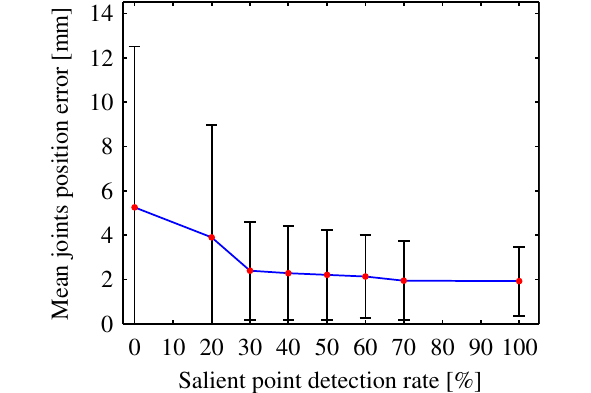}		&  
		\hspace*{-10.5mm}	\includegraphics[trim=08mm 000mm 00mm 00mm, clip=true, height=0.20\textwidth]{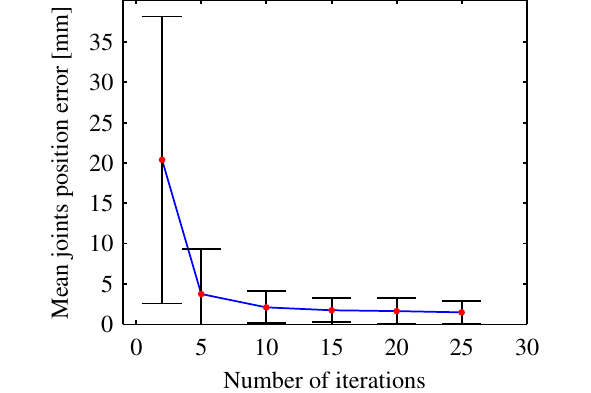}		\\
		\hspace*{+00.0mm}	{(a)} 													&
		\hspace*{-10.5mm}	{(b)} 	
		\end{tabular}
		\end{center}
		\vspace*{-4mm}
		\caption{Quantitative evaluation of the algorithm performance on noisy data, with respect to the
		salient point detection rate (a), and the number of iterations (b). Black bars indicate
		the standard deviation of the obtained error.}
		\label{fig:DetectionGraph}
		\end{figure}

		In order to make the synthetic experiments 
		as realistic as possible, 
		we simulated noise in all of the visual features.
		More precisely, edge detection errors were introduced by adding structural
		noise to the images, i.e. by adding and subtracting at random positions in each image $100$
		circles of radius varying between $10$ and $30$ pixels.
		The optical flow features corresponding to those circles were also not considered.
		Errors in the salient point detector were simulated by randomly deleting
		detections as well as by randomly adding outliers in a radius of $200$ pixels
		around the actual features. Gaussian noise of $5$ pixels was
		further introduced on the coordinates of the resulting salient points.
		Figure~\ref{fig:DetectionGraph}(a) shows the influence of the
		salient point detector on the accuracy of the pose estimation in case
		of noisy data. This experiment was run with a salient point false positive rate
		of $10\%$, and with varying detection rates.
		It is visible that the error quickly drops very close to its
		minimum even with a detection rate of only $30\%$.

		Figure~\ref{fig:DetectionGraph}(b)
		shows the convergence rate for different numbers of iterations.
		It can be noted that the algorithm accuracy becomes quite reasonable after just $10-15$ iterations, which is the same as for the RGB-D sequences.
		
		We also annotated one of the captured sequences for evaluation. 
		Since annotating joints in multiple RGB cameras is more time consuming than annotating joints in a single RGB-D camera, 
		we manually labeled only three points 
		on the hands in all camera views of the sequence \textit{``Holding and Passing a Ball''}. Since we obtain 3D points by
		triangulation, we therefore use the 3D distance between these points and the corresponding vertices in
		the hand model as error metric.
		Table~\ref{fig:TResults} shows the tracking accuracy obtained in this experiment.
		Overall, the median of the tracking error is at maximum $1$cm.

		\begin{table}[t]
			\footnotesize 
			\begin{center}
				\caption{	Results obtained on the manually marked data for the multicamera RGB sequences.
						The table reports the distance in mm between the manually tracked 3D points and the
						corresponding vertices on the hand model. The figure 
						shows the positions of the tracked points on the hand.
				}
				\vspace*{2mm}
				\label{fig:TResults}
				\setlength{\tabcolsep}{1pt}	
				\begin{tabular}{|l|c|c|c|c|c|cc}											\hhline{-|~|----~~}
					\multicolumn{1}{|c|}{~~Points~~} & 
					\multicolumn{1}{ c|}{} 	 & ~median~ & ~~mean~~ & ~~std~~ & ~~max~~& \multicolumn{1}{ c}{} 	
					& \multirow{4}{*}[3.8mm]{\includegraphics[width=2cm]{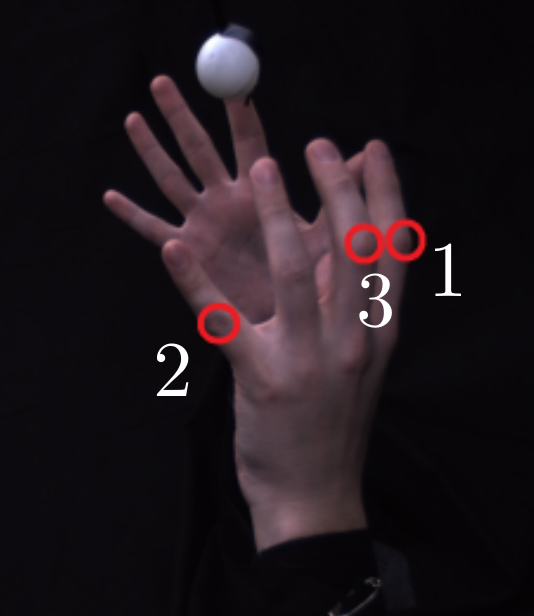}} 			\\	\hhline{-|~|----~~}
					\noalign{\smallskip}												\hhline{-|~|----~~}
					{~~point 1~~}	& {}	& { $06.98$ }	& { $07.98$ }	& {~$3.54$~} & {~$20.53$~}	& {}	& {} 	\\	\hhline{-|~|----~~}
					{~~point 2~~}	& {}	& {~$11.14$~}	& {~$12.28$~}	& {~$5.22$~} & {~$23.48$~}	& {}	& {} 	\\	\hhline{-|~|----~~}
					{~~point 3~~}	& {}	& { $10.91$ }	& { $10.72$ }	& { $4.13$ } & { $24.68$ }	& {}	& {} 	\\	\hhline{-|~|----~~}				
				\end{tabular}
			\end{center}
		\end{table}

	\newcommand{\ResultsWidth}{0.23\textwidth}
	\newcommand{\ResultsSkip}{0.7mm}
	\newcommand{\ResultsSquize}{5.2mm}

	\begin{figure*}[t]
	\begin{center}
	\tiny
	\begin{tabular}{c c c c}			
	\hspace*{+\ResultsSquize}	\includegraphics[trim=020mm 029mm 020mm 017mm, clip=true, width=\ResultsWidth]{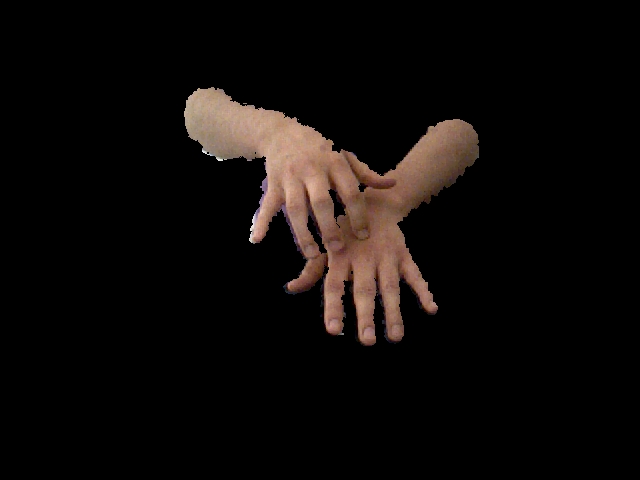}			&	
	\hspace*{-\ResultsSquize}	\includegraphics[trim=020mm 029mm 020mm 017mm, clip=true, width=\ResultsWidth]{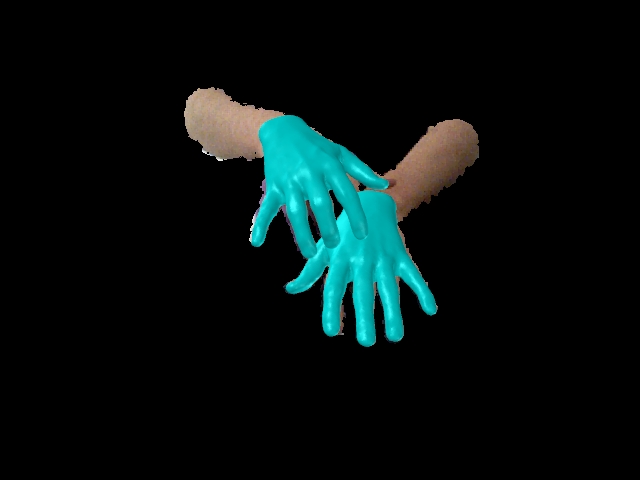}		&	
	\hspace*{-\ResultsSquize}	\includegraphics[trim=020mm 029mm 020mm 017mm, clip=true, width=\ResultsWidth]{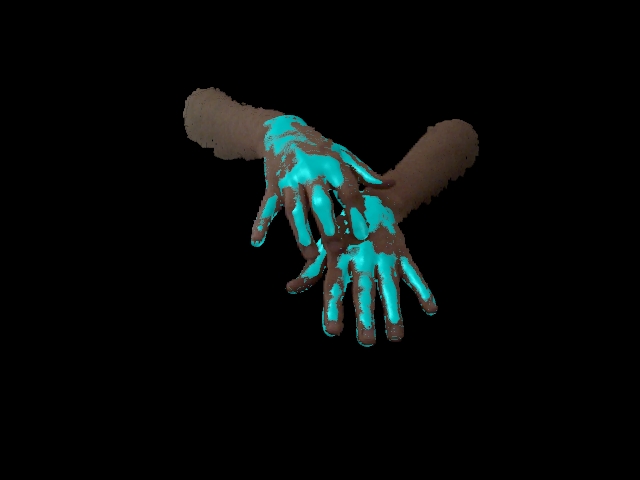}			&	
	\hspace*{-\ResultsSquize}	\includegraphics[trim=020mm 026mm 007mm 012mm, clip=true, width=\ResultsWidth]{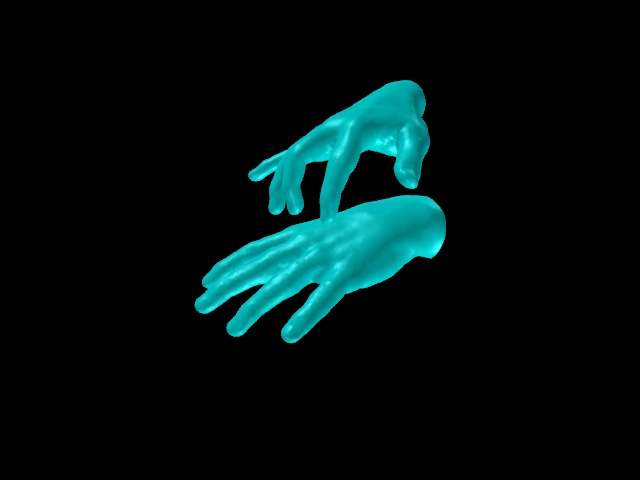}	\\	& \hspace*{+\ResultsSquize}	(a) Fingers Walking 			 \hspace*{-3\ResultsSquize}	& \vspace*{\ResultsSkip} \\

	\hspace*{+\ResultsSquize}	\includegraphics[trim=010mm 043mm 010mm 014mm, clip=true, width=\ResultsWidth]{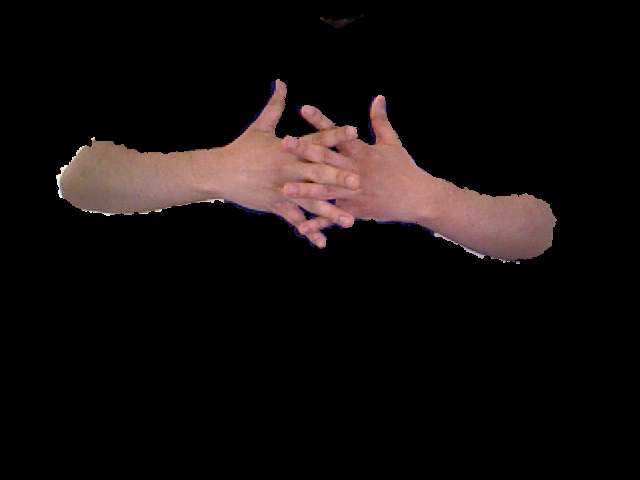}			&	
	\hspace*{-\ResultsSquize}	\includegraphics[trim=010mm 043mm 010mm 014mm, clip=true, width=\ResultsWidth]{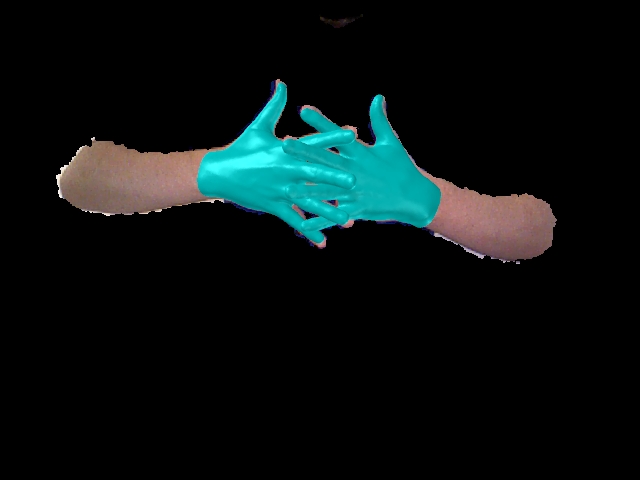}		&	
	\hspace*{-\ResultsSquize}	\includegraphics[trim=010mm 043mm 010mm 014mm, clip=true, width=\ResultsWidth]{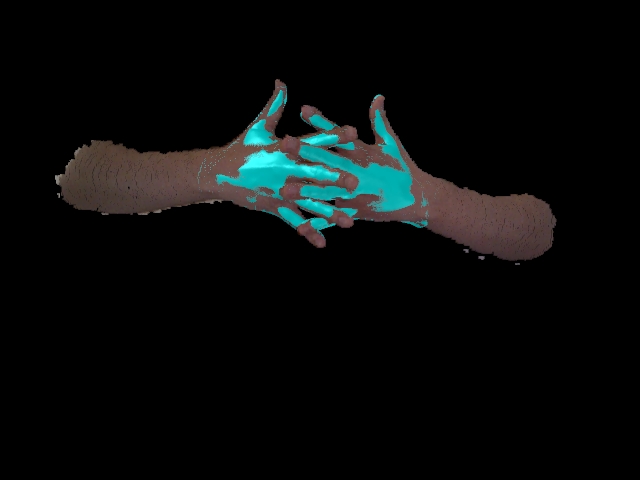}			&	
	\hspace*{-\ResultsSquize}	\includegraphics[trim=026mm 040mm 026mm 032mm, clip=true, width=\ResultsWidth]{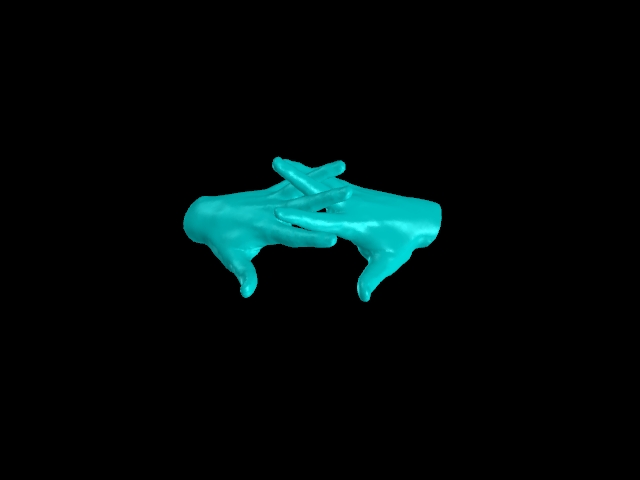}	\\	& \hspace*{+\ResultsSquize}	(b) Fingers Crossing 		 	\hspace*{-3\ResultsSquize}	& \vspace*{\ResultsSkip} \\

	\hspace*{+\ResultsSquize}	\includegraphics[trim=009mm 044mm 014mm 014mm, clip=true, width=\ResultsWidth]{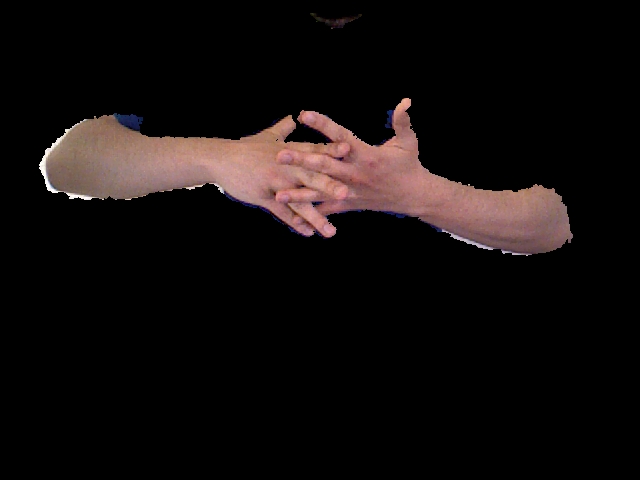}			&	
	\hspace*{-\ResultsSquize}	\includegraphics[trim=009mm 044mm 014mm 014mm, clip=true, width=\ResultsWidth]{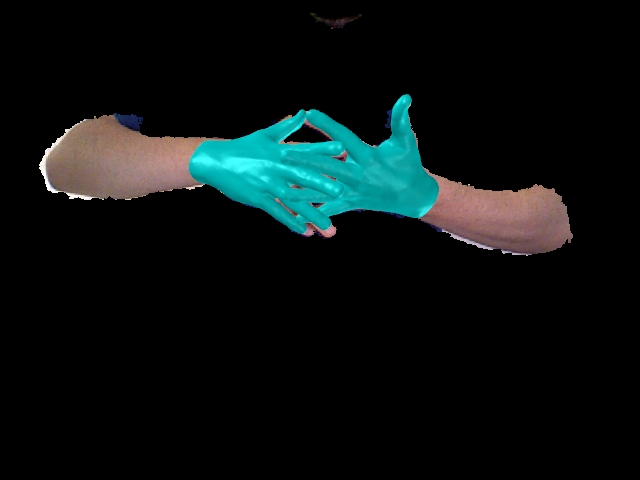}		&	
	\hspace*{-\ResultsSquize}	\includegraphics[trim=009mm 044mm 014mm 014mm, clip=true, width=\ResultsWidth]{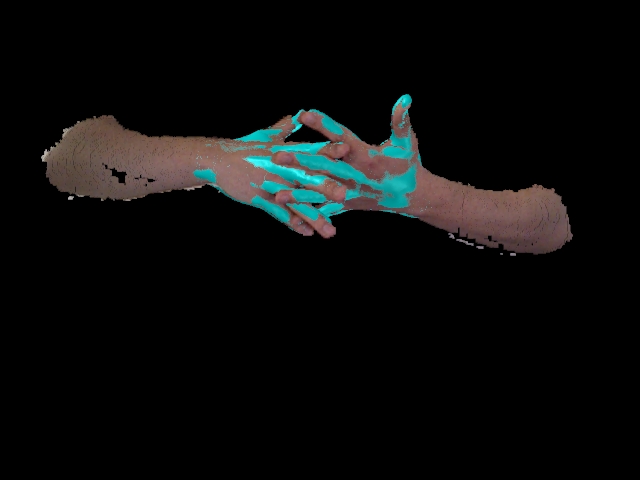}			&	
	\hspace*{-\ResultsSquize}	\includegraphics[trim=000mm 023mm 000mm 024mm, clip=true, width=\ResultsWidth]{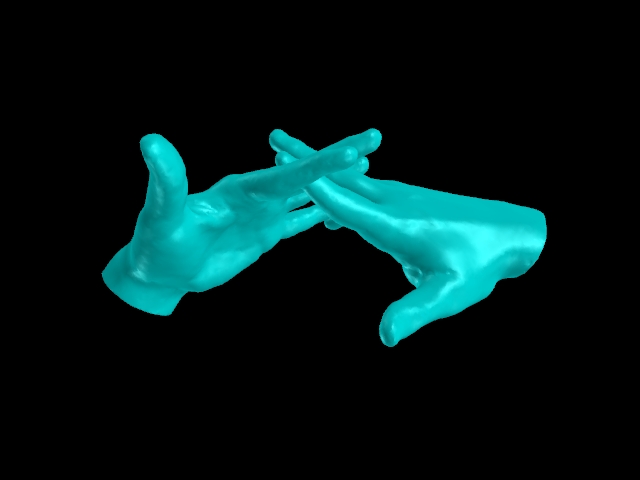}	\\	& \hspace*{+\ResultsSquize}	(c) Fingers Crossing and Twisting	 \hspace*{-3\ResultsSquize}	& \vspace*{\ResultsSkip} \\

	\hspace*{+\ResultsSquize}	\includegraphics[trim=050mm 027mm 008mm 032mm, clip=true, width=\ResultsWidth]{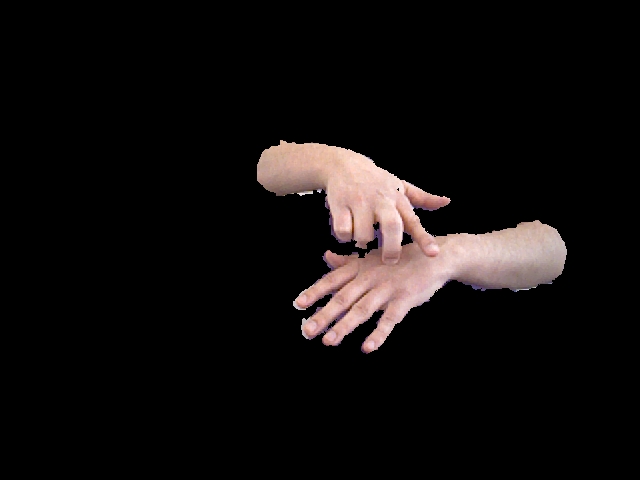}			&	
	\hspace*{-\ResultsSquize}	\includegraphics[trim=050mm 027mm 008mm 032mm, clip=true, width=\ResultsWidth]{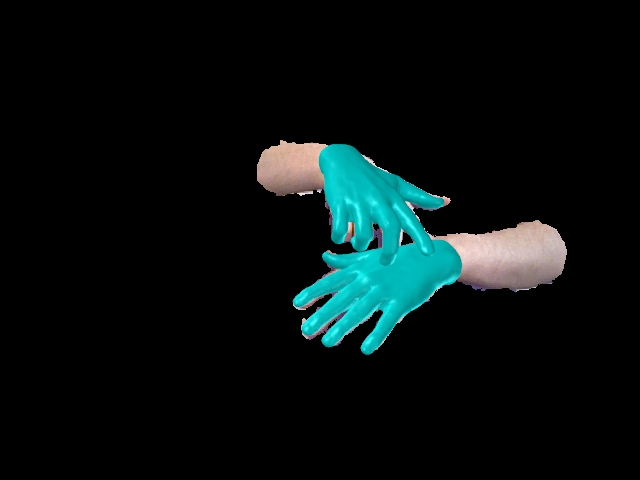}		&	
	\hspace*{-\ResultsSquize}	\includegraphics[trim=050mm 027mm 008mm 032mm, clip=true, width=\ResultsWidth]{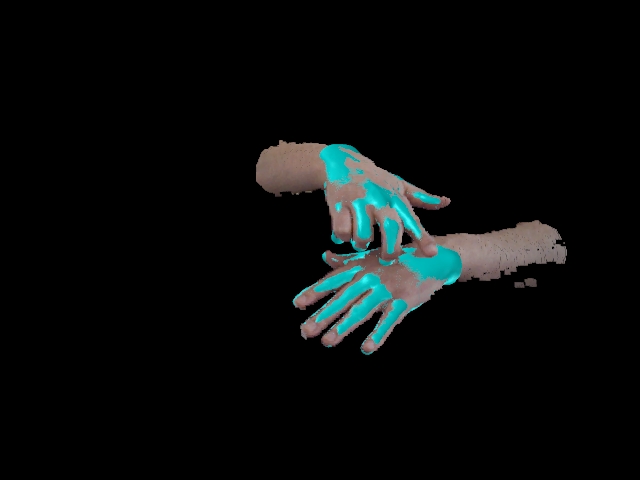}			&	
	\hspace*{-\ResultsSquize}	\includegraphics[trim=016mm 028mm 042mm 031mm, clip=true, width=\ResultsWidth]{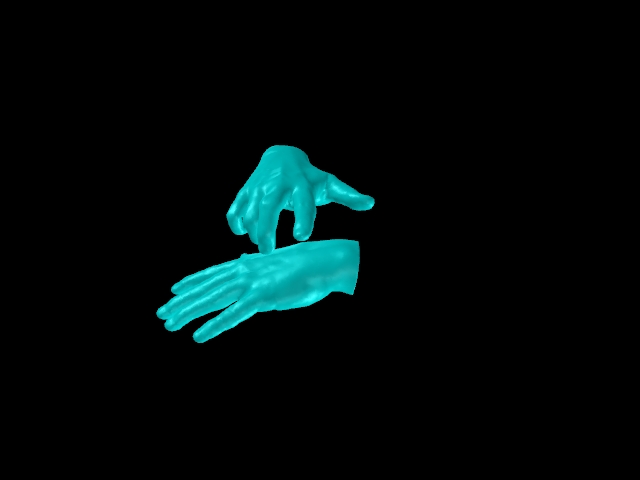}	\\	& \hspace*{+\ResultsSquize}	(d) Fingers Dancing	 		\hspace*{-3\ResultsSquize}	& \vspace*{\ResultsSkip} \\

	\hspace*{+\ResultsSquize}	\includegraphics[trim=065mm 040mm 005mm 037mm, clip=true, width=\ResultsWidth]{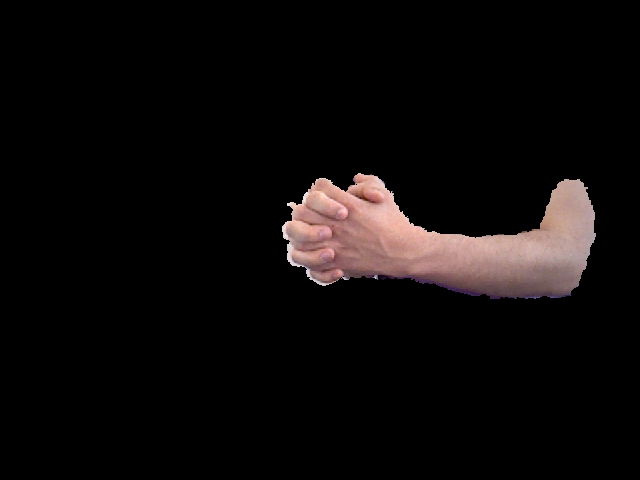}			&	
	\hspace*{-\ResultsSquize}	\includegraphics[trim=065mm 040mm 005mm 037mm, clip=true, width=\ResultsWidth]{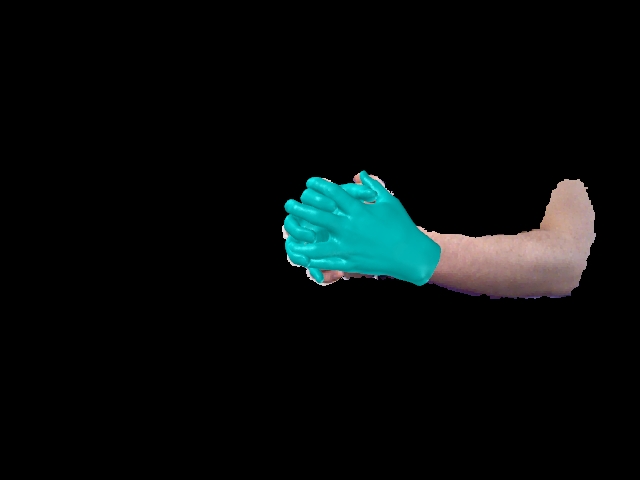}		&	
	\hspace*{-\ResultsSquize}	\includegraphics[trim=065mm 040mm 005mm 037mm, clip=true, width=\ResultsWidth]{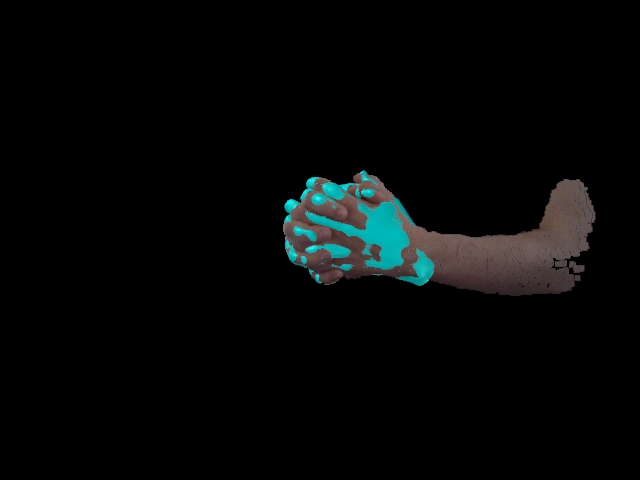}			&	
	\hspace*{-\ResultsSquize}	\includegraphics[trim=007mm 022mm 006mm 026mm, clip=true, width=\ResultsWidth]{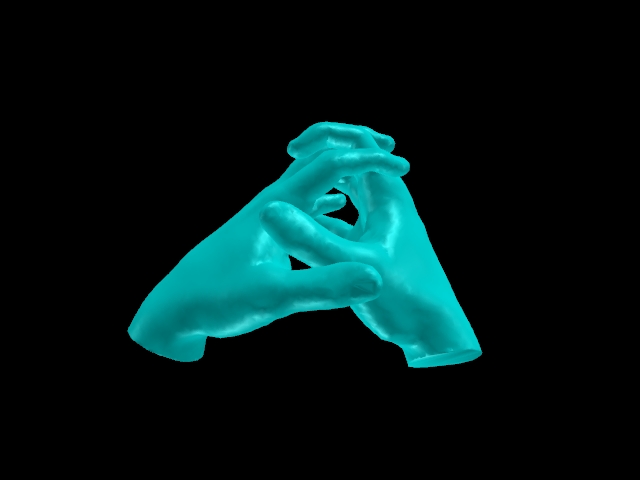}	\\	& \hspace*{+\ResultsSquize}	(d) Fingers Hugging			\hspace*{-3\ResultsSquize}	& \vspace*{\ResultsSkip} \\

	\hspace*{+\ResultsSquize}	\includegraphics[trim=018mm 071mm 088mm 024mm, clip=true, width=\ResultsWidth]{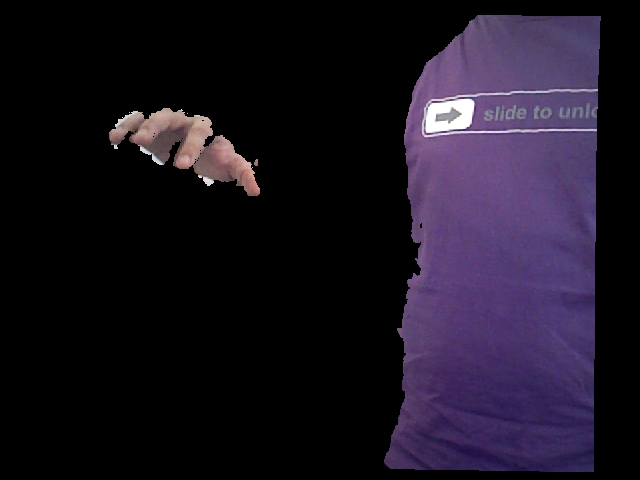}			&	
	\hspace*{-\ResultsSquize}	\includegraphics[trim=018mm 071mm 088mm 024mm, clip=true, width=\ResultsWidth]{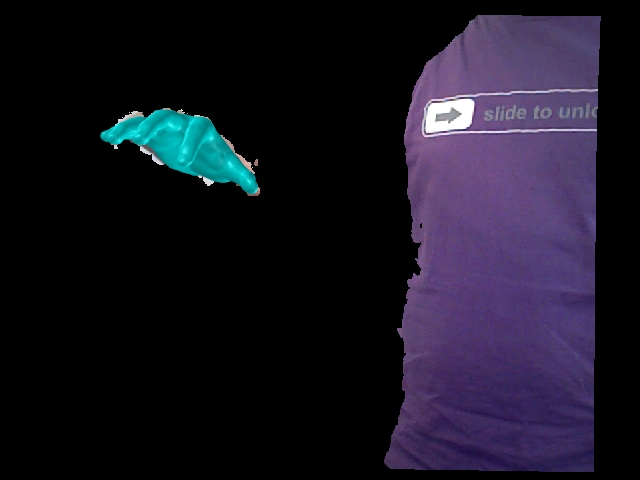}		&	
	\hspace*{-\ResultsSquize}	\includegraphics[trim=018mm 071mm 088mm 024mm, clip=true, width=\ResultsWidth]{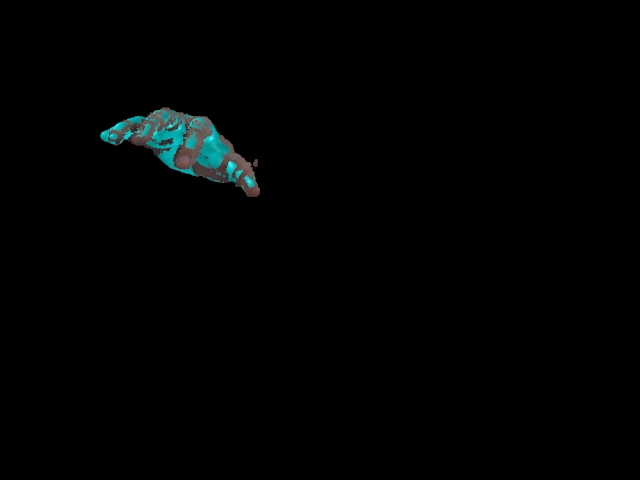}			&	
	\hspace*{-\ResultsSquize}	\includegraphics[trim=001mm 018mm 000mm 024mm, clip=true, width=\ResultsWidth]{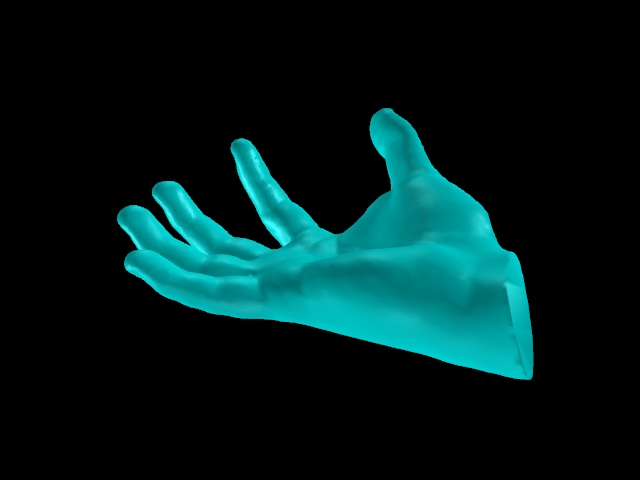}	\\	& \hspace*{+\ResultsSquize}	(d) Fingers Grasping			\hspace*{-3\ResultsSquize}	& \vspace*{\ResultsSkip} \\

	\hspace*{+\ResultsSquize}	\includegraphics[trim=000mm 053mm 070mm 017mm, clip=true, width=\ResultsWidth]{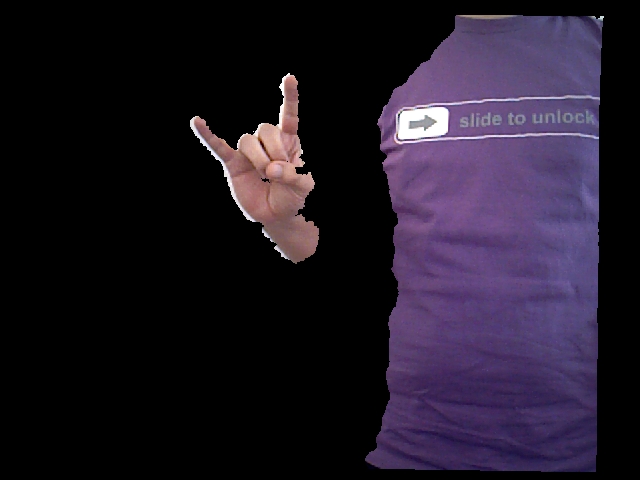}			&	
	\hspace*{-\ResultsSquize}	\includegraphics[trim=000mm 053mm 070mm 017mm, clip=true, width=\ResultsWidth]{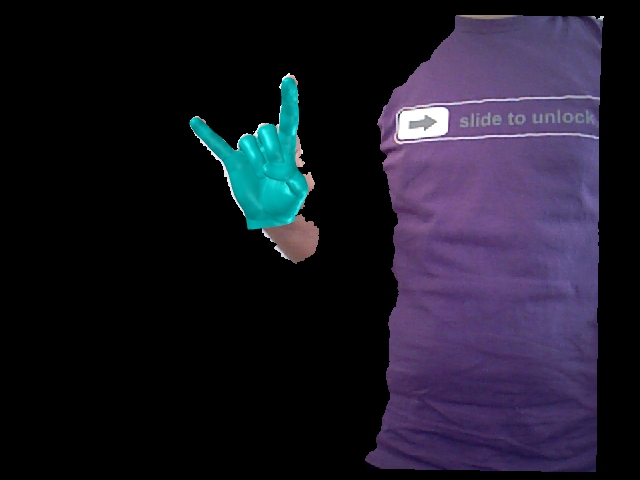}		&	
	\hspace*{-\ResultsSquize}	\includegraphics[trim=000mm 053mm 070mm 017mm, clip=true, width=\ResultsWidth]{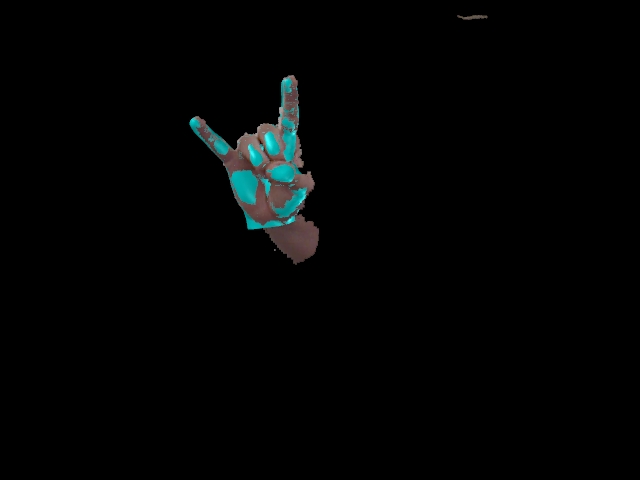}			&
	\hspace*{-\ResultsSquize}	\includegraphics[trim=013mm 023mm 015mm 023mm, clip=true, width=\ResultsWidth]{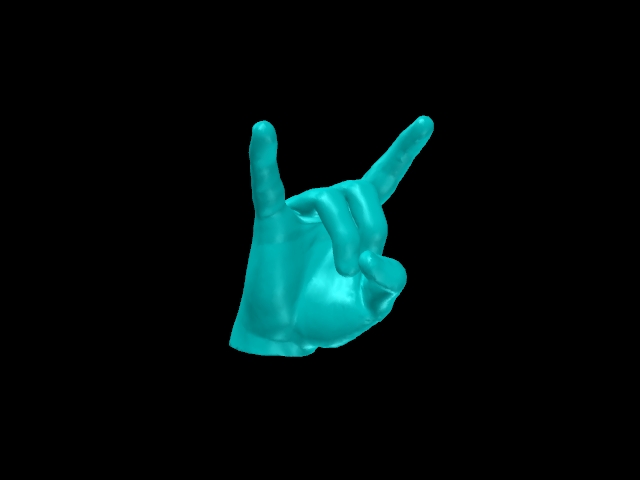}	\\	& \hspace*{+\ResultsSquize}	(d) Rock Gesture			\hspace*{-3\ResultsSquize}	& \vspace*{\ResultsSkip} \\

	\hspace*{+\ResultsSquize}	\includegraphics[trim=002mm 065mm 065mm 007mm, clip=true, width=\ResultsWidth]{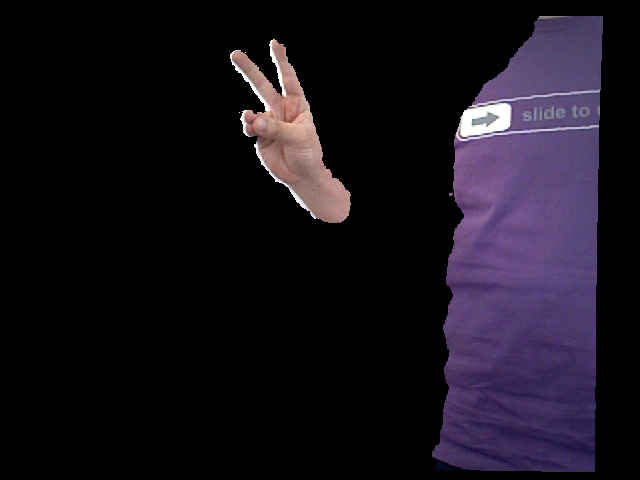}			&	
	\hspace*{-\ResultsSquize}	\includegraphics[trim=002mm 065mm 065mm 007mm, clip=true, width=\ResultsWidth]{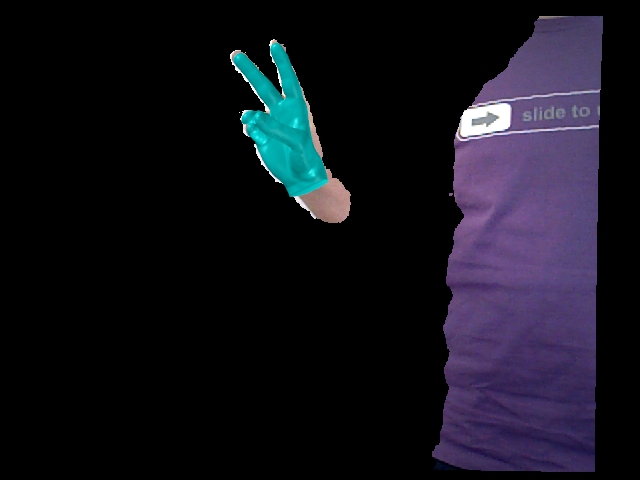}		&	
	\hspace*{-\ResultsSquize}	\includegraphics[trim=002mm 065mm 065mm 007mm, clip=true, width=\ResultsWidth]{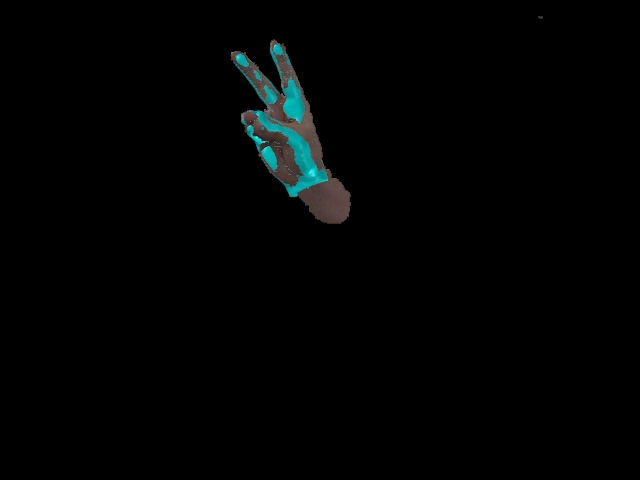}			&	
	\hspace*{-\ResultsSquize}	\includegraphics[trim=014mm 029mm 012mm 021mm, clip=true, width=\ResultsWidth]{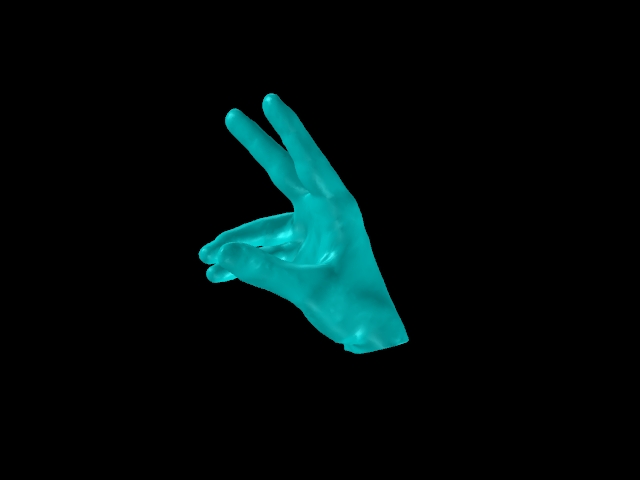}	\\	& \hspace*{+\ResultsSquize}	(d) Bunny Gesture			\hspace*{-3\ResultsSquize}	& \vspace*{\ResultsSkip} \\

	\end{tabular}					
	\normalsize
	\caption{	Some of the obtained results. 
			(Left) Input RGB-D image. 
			(Center-Left) Obtained results overlayed on the input image. 
			(Center-Right) Obtained results fitted in the input point cloud. 
			(Right) Obtained results from another viewpoint. 
	}
	\label{fig:My_Results_NoObject}
	\end{center}
	\end{figure*}
	
	\newcommand{\ResultsSquizZZ}{9.2mm}
	\newcommand{\ResultsSquizCC}{5.2mm}

	\begin{figure*}[t]
	\begin{center}
	\tiny
	\begin{tabular}{c c c c}			
	\hspace*{+2mm}			\includegraphics[trim=005mm 0mm 005mm 000mm, clip=true, width=\ResultsWidth]{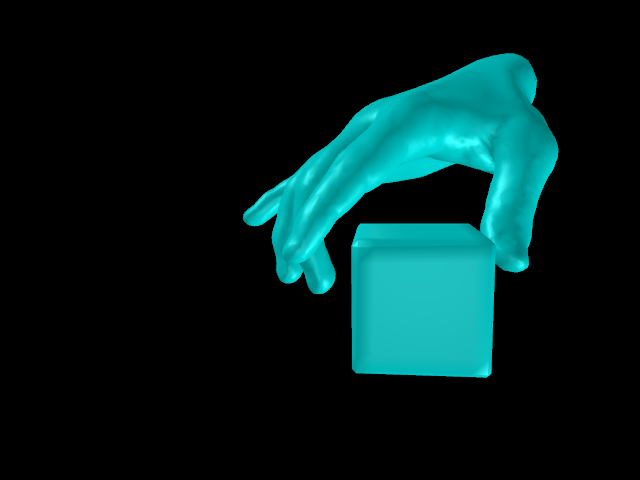}		&	
	\hspace*{-\ResultsSquizZZ}	\includegraphics[trim=005mm 0mm 005mm 000mm, clip=true, width=\ResultsWidth]{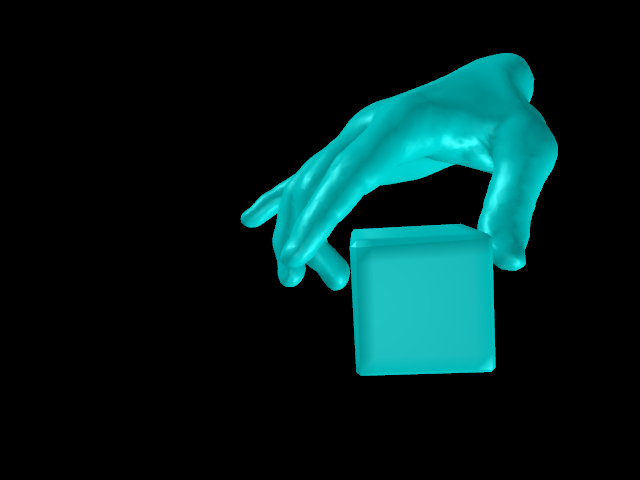}		&	
	\hspace*{-3mm}			\includegraphics[trim=005mm 0mm 005mm 000mm, clip=true, width=\ResultsWidth]{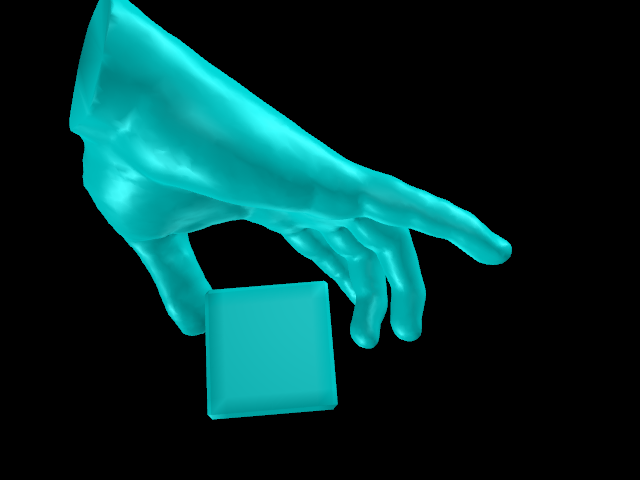}		&	
	\hspace*{-\ResultsSquizCC}	\includegraphics[trim=005mm 0mm 005mm 000mm, clip=true, width=\ResultsWidth]{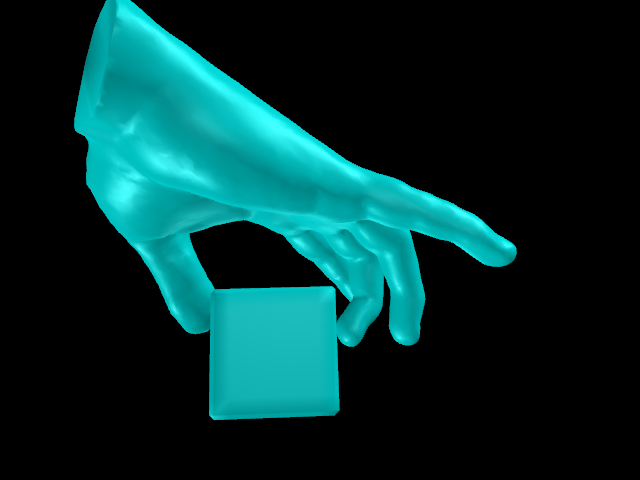}	\\	& \hspace*{+1mm}	(a) \textit{``Moving a Cube''} with occluded manipulating finger, Frame $083$						\hspace*{-3\ResultsSquizZZ}	& \vspace*{\ResultsSkip} \\

	\hspace*{+2mm}			\includegraphics[trim=005mm 0mm 005mm 000mm, clip=true, width=\ResultsWidth]{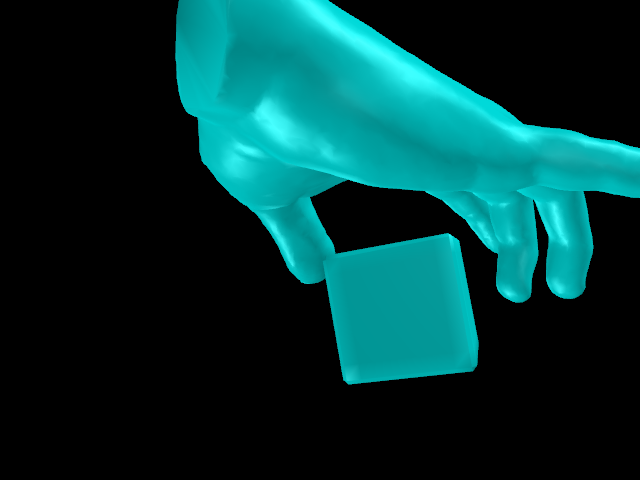}		&	
	\hspace*{-\ResultsSquizZZ}	\includegraphics[trim=005mm 0mm 005mm 000mm, clip=true, width=\ResultsWidth]{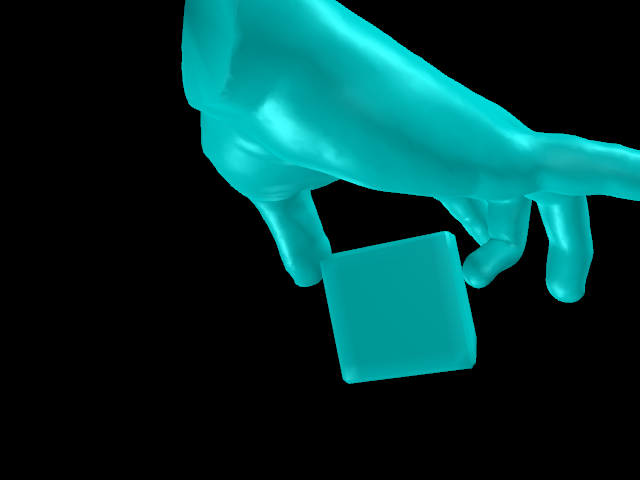}		&	
	\hspace*{-3mm}			\includegraphics[trim=005mm 0mm 005mm 000mm, clip=true, width=\ResultsWidth]{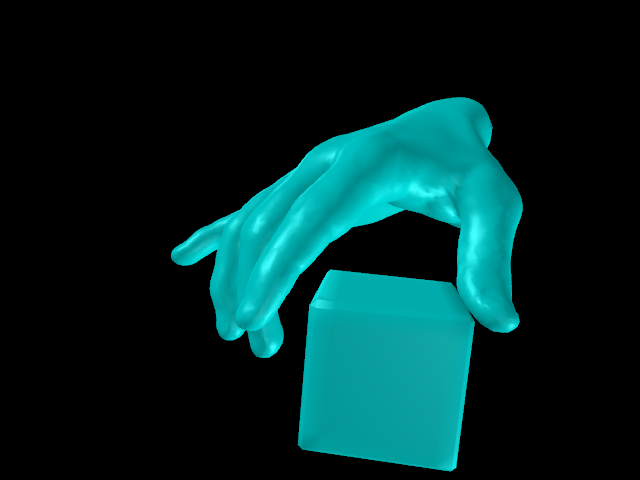}		&	
	\hspace*{-\ResultsSquizCC}	\includegraphics[trim=005mm 0mm 005mm 000mm, clip=true, width=\ResultsWidth]{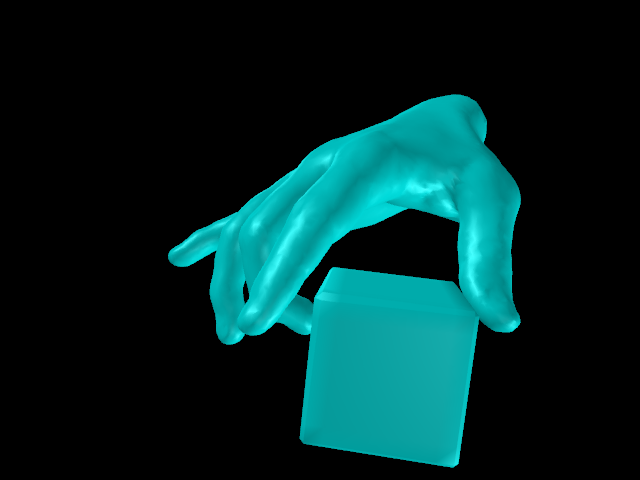}	\\	& \hspace*{+1mm}	(b) \textit{``Moving a Cube''} with occluded manipulating finger, Frame $106$		 				\hspace*{-3\ResultsSquizZZ}	& \vspace*{\ResultsSkip} \\

	\hspace*{+2mm}			\includegraphics[trim=005mm 0mm 005mm 000mm, clip=true, width=\ResultsWidth]{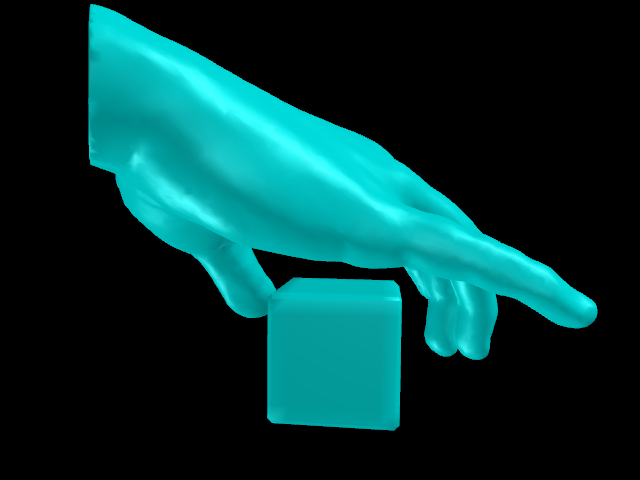}		&	
	\hspace*{-\ResultsSquizZZ}	\includegraphics[trim=005mm 0mm 005mm 000mm, clip=true, width=\ResultsWidth]{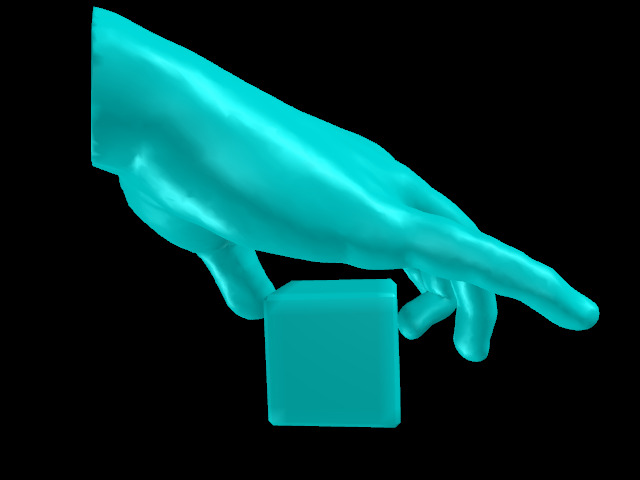}		&	
	\hspace*{-3mm}			\includegraphics[trim=005mm 0mm 005mm 000mm, clip=true, width=\ResultsWidth]{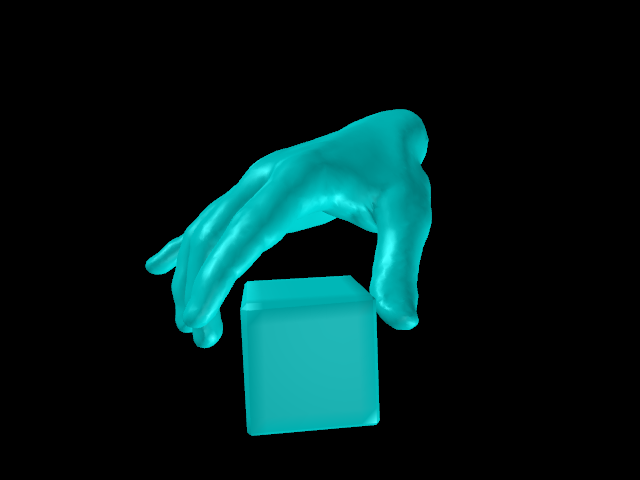}		&	
	\hspace*{-\ResultsSquizCC}	\includegraphics[trim=005mm 0mm 005mm 000mm, clip=true, width=\ResultsWidth]{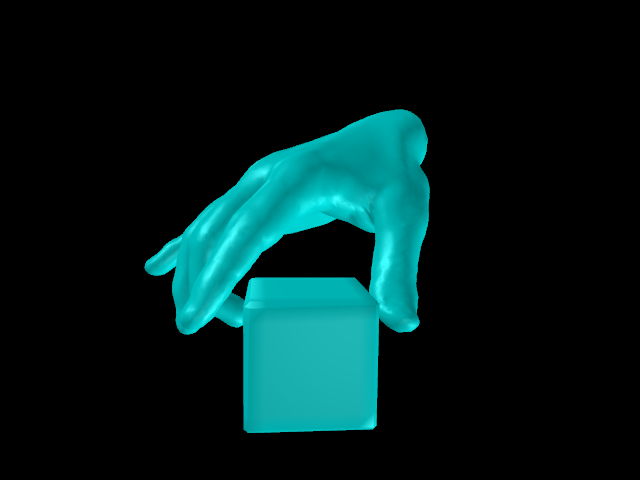}	\\	& \hspace*{+1mm}	(c) \textit{``Moving a Cube''} with occluded manipulating finger, Frame $125$		 				\hspace*{-3\ResultsSquizZZ}	& \vspace*{\ResultsSkip} \\

	\hspace*{+2mm}			\includegraphics[trim=005mm 0mm 005mm 000mm, clip=true, width=\ResultsWidth]{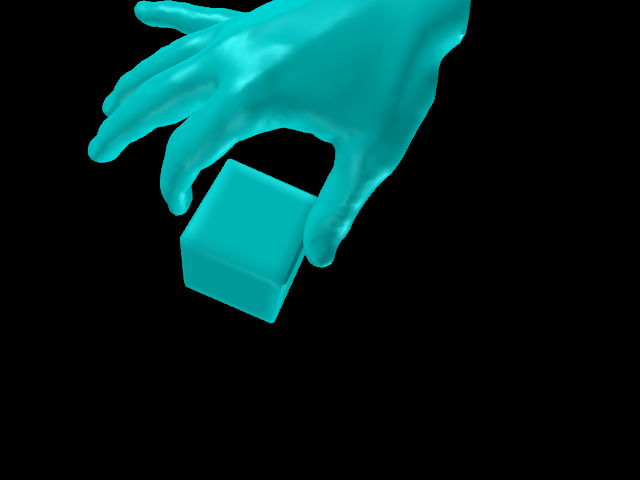}		&	
	\hspace*{-\ResultsSquizZZ}	\includegraphics[trim=005mm 0mm 005mm 000mm, clip=true, width=\ResultsWidth]{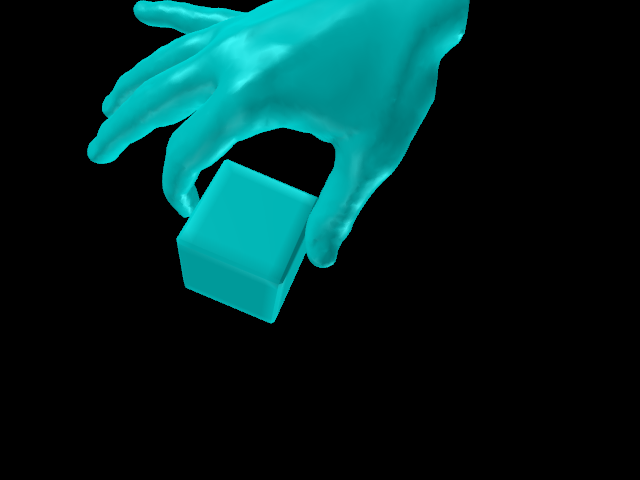}		&	
	\hspace*{-3mm}			\includegraphics[trim=005mm 0mm 005mm 000mm, clip=true, width=\ResultsWidth]{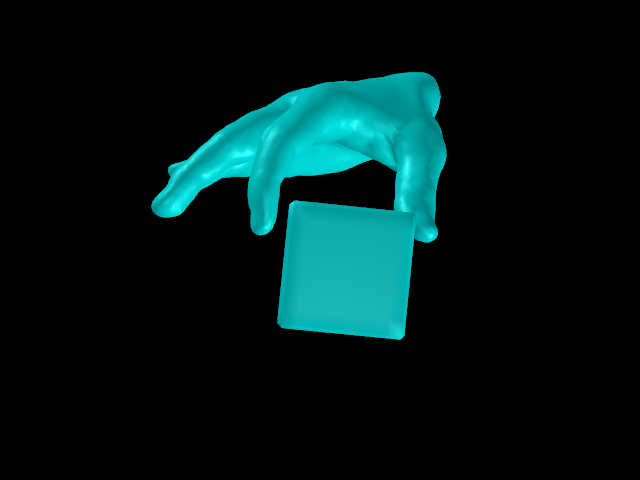}		&	
	\hspace*{-\ResultsSquizCC}	\includegraphics[trim=005mm 0mm 005mm 000mm, clip=true, width=\ResultsWidth]{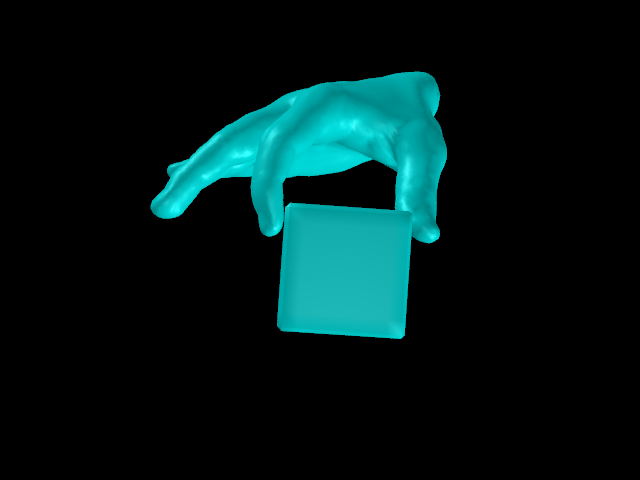}	\\	& \hspace*{+1mm}	(d) \textit{``Moving a Cube''}, Frame $085$		 								\hspace*{-3\ResultsSquizZZ}	& \vspace*{\ResultsSkip} \\

	\hspace*{+2mm}			\includegraphics[trim=005mm 0mm 005mm 000mm, clip=true, width=\ResultsWidth]{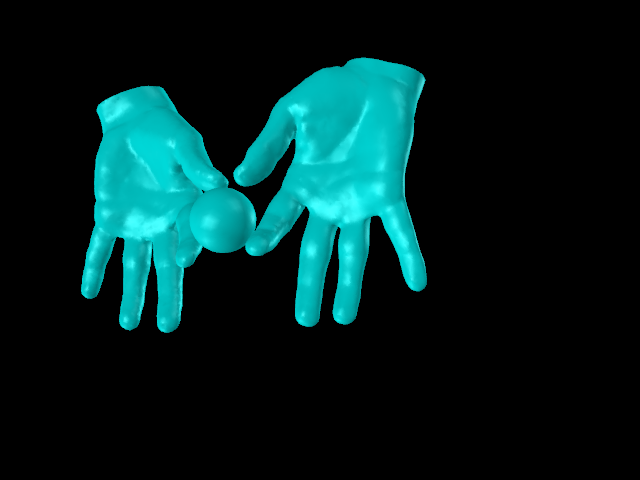}		&	
	\hspace*{-\ResultsSquizZZ}	\includegraphics[trim=005mm 0mm 005mm 000mm, clip=true, width=\ResultsWidth]{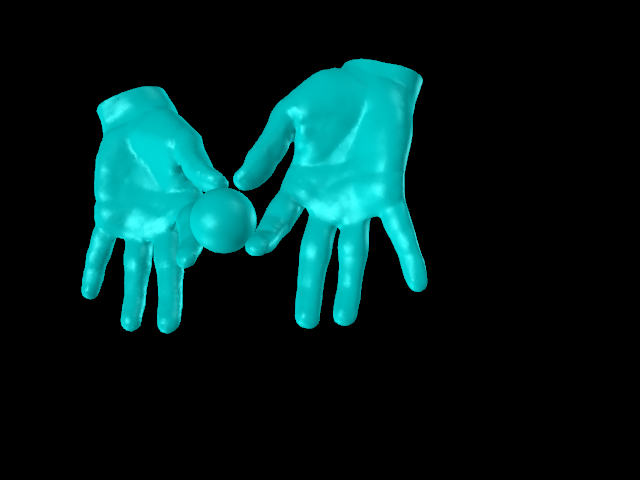}		&	
	\hspace*{-3mm}			\includegraphics[trim=005mm 0mm 005mm 000mm, clip=true, width=\ResultsWidth]{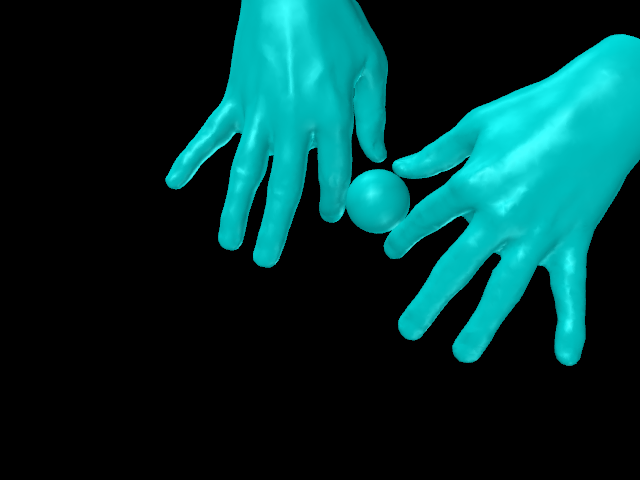}		&	
	\hspace*{-\ResultsSquizCC}	\includegraphics[trim=005mm 0mm 005mm 000mm, clip=true, width=\ResultsWidth]{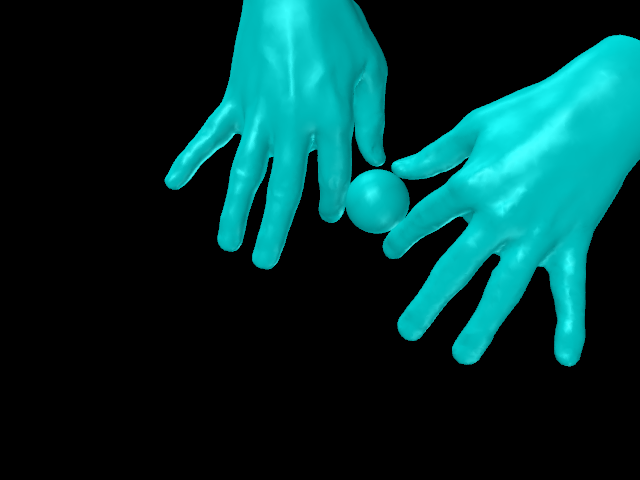}	\\	& \hspace*{+1mm}	(e) \textit{``Moving a Ball''} with $2$ hands, Frame $113$		 						\hspace*{-3\ResultsSquizZZ}	& \vspace*{\ResultsSkip} \\

	\hspace*{+2mm}			\includegraphics[trim=000mm 0mm 000mm 000mm, clip=true, width=\ResultsWidth]{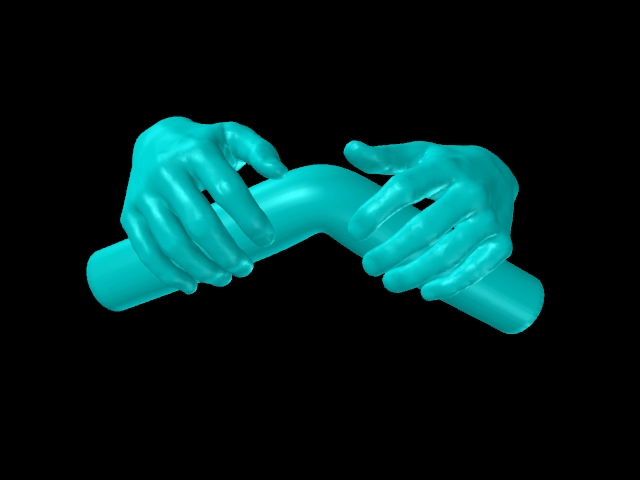}		&	
	\hspace*{-\ResultsSquizZZ}	\includegraphics[trim=000mm 0mm 000mm 000mm, clip=true, width=\ResultsWidth]{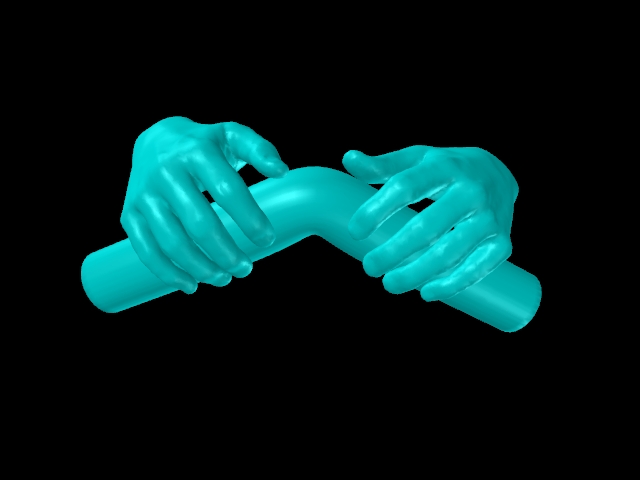}		&	
	\hspace*{-3mm}			\includegraphics[trim=000mm 0mm 000mm 000mm, clip=true, width=\ResultsWidth]{00159_SCx_camC.jpg}		&	
	\hspace*{-\ResultsSquizCC}	\includegraphics[trim=000mm 0mm 000mm 000mm, clip=true, width=\ResultsWidth]{00159_SCP_camC.jpg}	\\	& \hspace*{+1mm}	(f) (Left) \textit{``Bending a Pipe''}, Frame $159$. Right \textit{``Bending a Rope''}, Frame $159$	 		\hspace*{-3\ResultsSquizZZ}	& \vspace*{\ResultsSkip} \\

	\end{tabular}					
	\normalsize
	\caption{	The impact of the physics component. 
			For each image couple, the left image corresponds to {LO + $\mathcal{S}$$\mathcal{C}$x} and the right one to {LO + $\mathcal{S}$$\mathcal{C}$$\mathcal{P}$}. 
			In the case of missing or ambiguous input visual data, as in sequences with occluded manipulating finger, 
			the contribution of the physics component towards better physically plausible poses becomes more prominent
		}
	\label{fig:physicsDIFF}
	\end{center}
	\end{figure*}

	\newcommand{\ResultsSquizAA}{13.2mm}
	\newcommand{\ResultsSquizBB}{5.2mm}

	\begin{figure*}[t]
	\begin{center}
	\tiny
	\begin{tabular}{c c c c}			
	\hspace*{+2mm}			\includegraphics[trim=025mm 030mm 000mm 000mm, clip=true, width=\ResultsWidth]{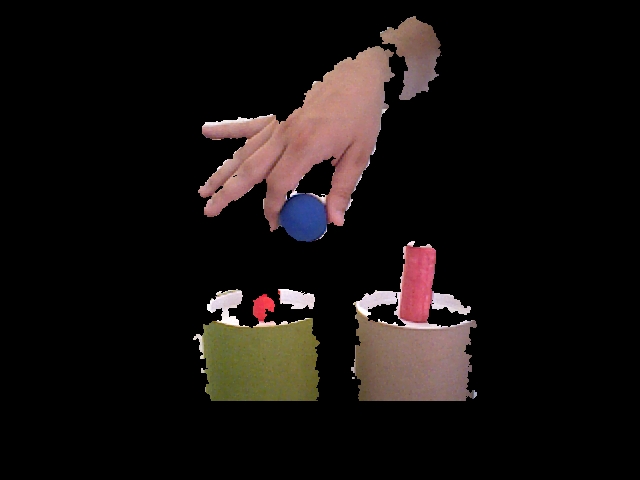}			&	
	\hspace*{-\ResultsSquizAA}	\includegraphics[trim=025mm 030mm 000mm 000mm, clip=true, width=\ResultsWidth]{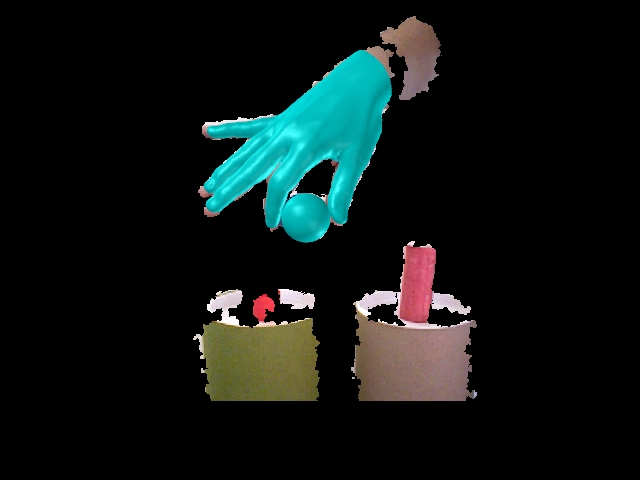}		&	
	\hspace*{-17mm}			\includegraphics[trim=025mm 030mm 000mm 000mm, clip=true, width=\ResultsWidth]{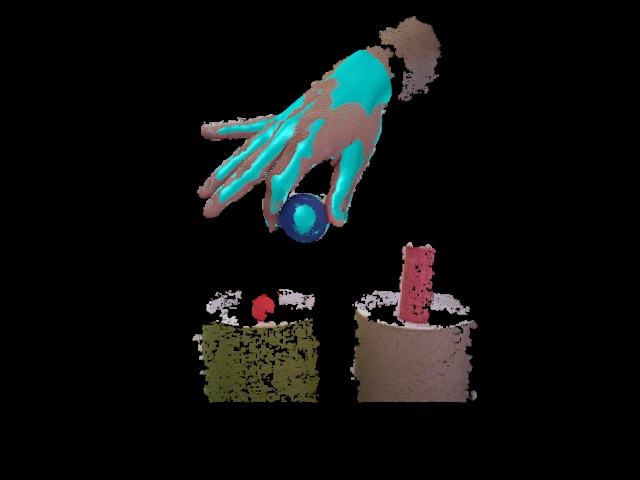}			&	
	\hspace*{-\ResultsSquizBB}	\includegraphics[trim=001mm 008mm 002mm 007mm, clip=true, width=\ResultsWidth]{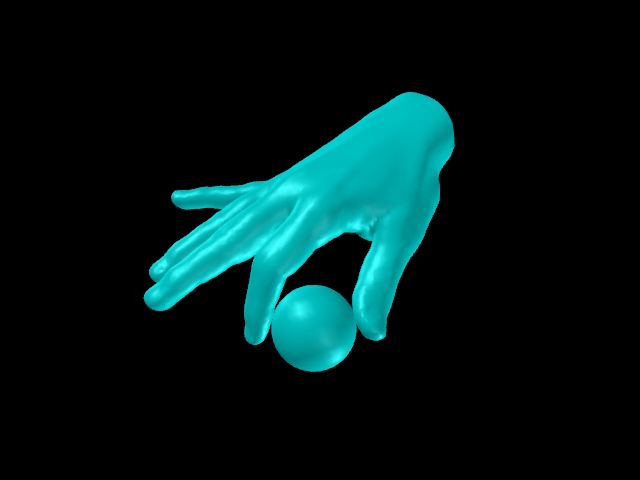}	\\	& \hspace*{+1mm}	(a) ``Moving a Ball'' with $1$ hand (new sequence) 				\hspace*{-3\ResultsSquize}	& \vspace*{\ResultsSkip} \\

	\hspace*{+2mm}			\includegraphics[trim=024mm 035mm 000mm 005mm, clip=true, width=\ResultsWidth]{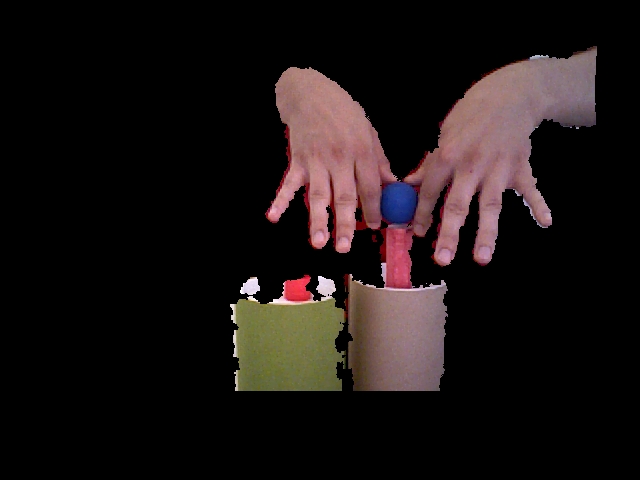}			&	
	\hspace*{-\ResultsSquizAA}	\includegraphics[trim=024mm 035mm 000mm 005mm, clip=true, width=\ResultsWidth]{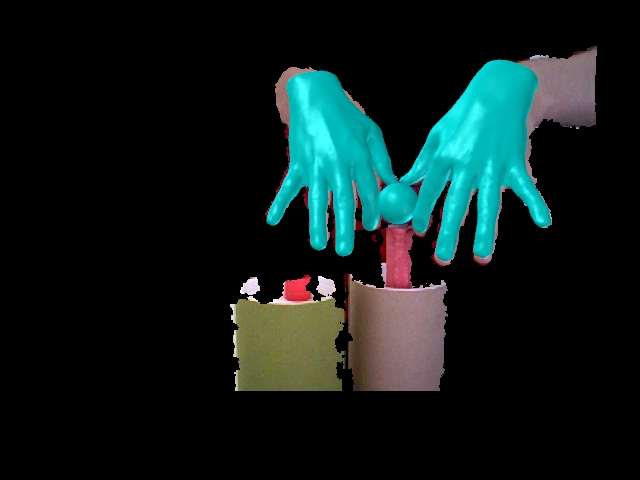}		&	
	\hspace*{-17mm}			\includegraphics[trim=024mm 035mm 000mm 005mm, clip=true, width=\ResultsWidth]{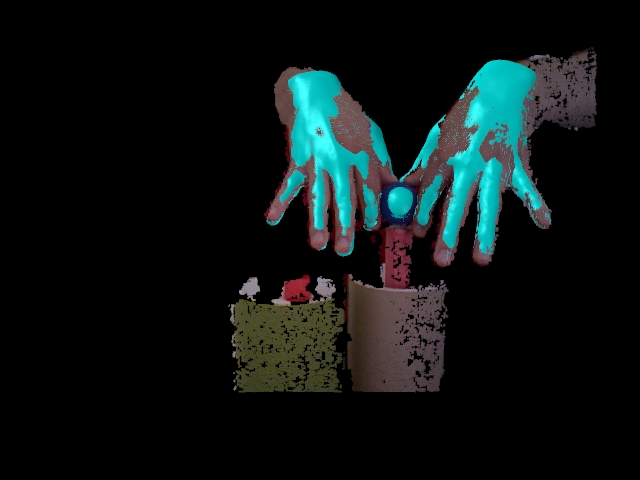}			&	
	\hspace*{-\ResultsSquizBB}	\includegraphics[trim=004mm 018mm 002mm 011mm, clip=true, width=\ResultsWidth]{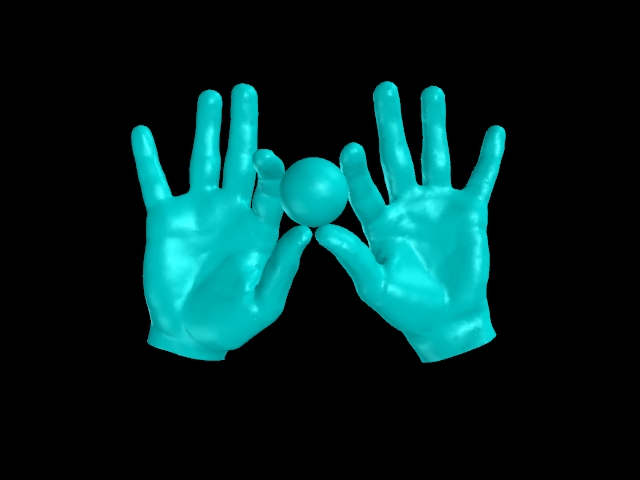}	\\	& \hspace*{+1mm}	(b) ``Moving a Ball'' with $2$ hands (new sequence) 				\hspace*{-3\ResultsSquize}	& \vspace*{\ResultsSkip} \\

	\hspace*{+2mm}			\includegraphics[trim=034mm 020mm 012mm 035mm, clip=true, width=\ResultsWidth]{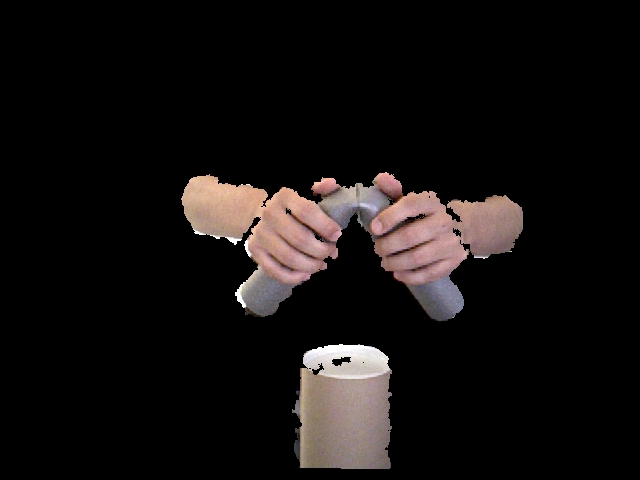}			&	
	\hspace*{-\ResultsSquizAA}	\includegraphics[trim=034mm 020mm 012mm 035mm, clip=true, width=\ResultsWidth]{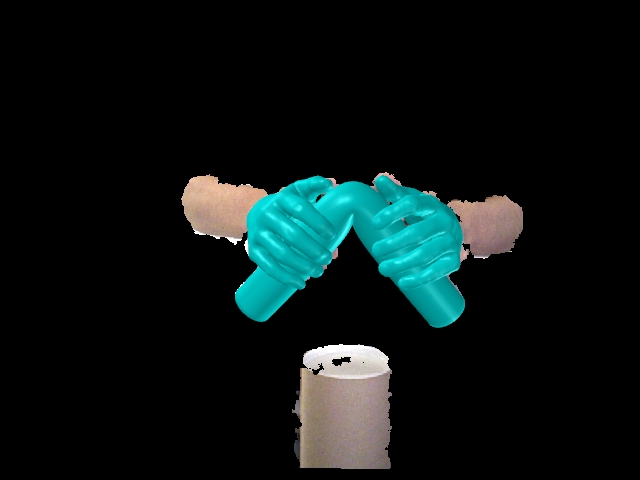}		&	
	\hspace*{-17mm}			\includegraphics[trim=034mm 020mm 012mm 035mm, clip=true, width=\ResultsWidth]{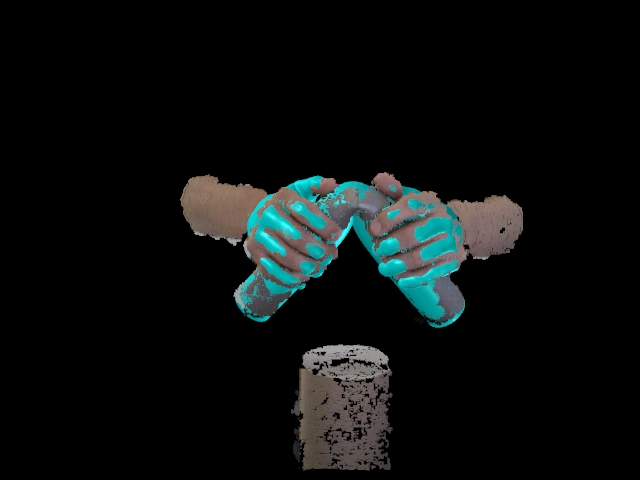}			&	
	\hspace*{-\ResultsSquizBB}	\includegraphics[trim=004mm 021mm 003mm 011mm, clip=true, width=\ResultsWidth]{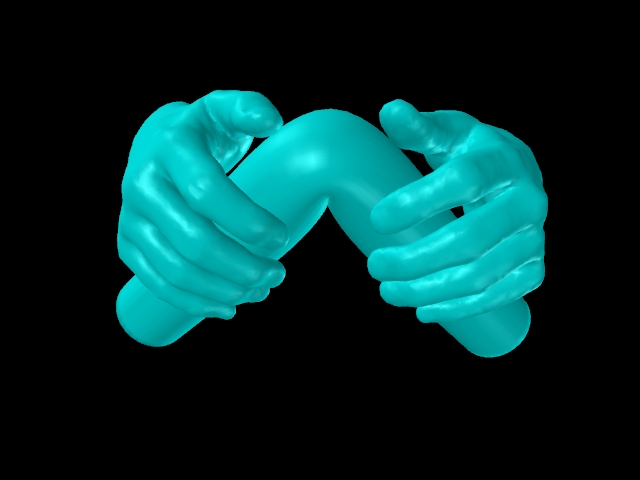}	\\	& \hspace*{+1mm}	(c) ``Bending a Pipe'' (new sequence) 						\hspace*{-3\ResultsSquize}	& \vspace*{\ResultsSkip} \\

	\hspace*{+2mm}			\includegraphics[trim=026mm 035mm 015mm 020mm, clip=true, width=\ResultsWidth]{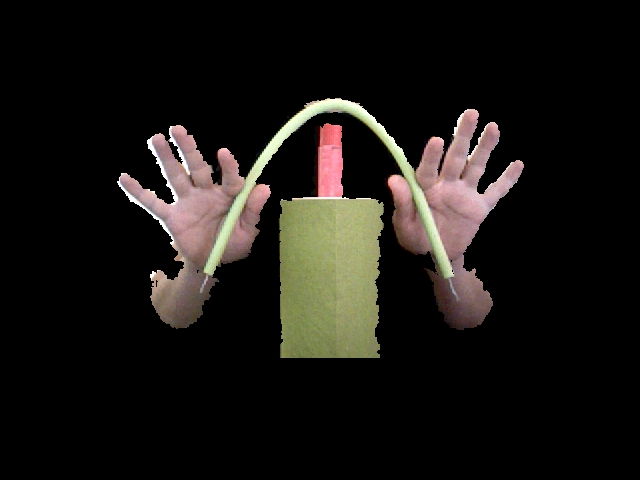}			&	
	\hspace*{-\ResultsSquizAA}	\includegraphics[trim=026mm 035mm 015mm 020mm, clip=true, width=\ResultsWidth]{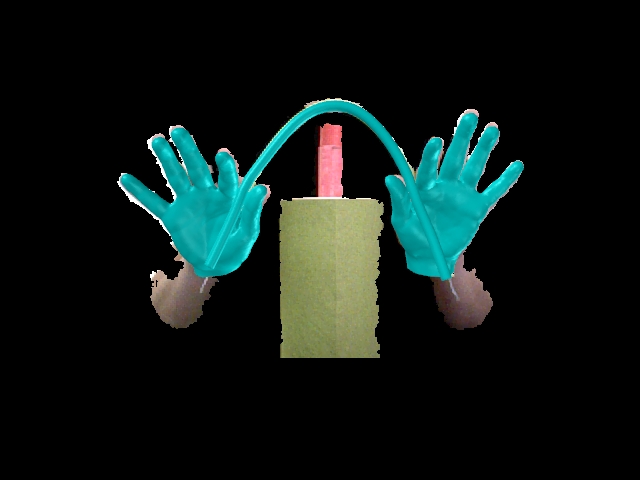}		&	
	\hspace*{-17mm}			\includegraphics[trim=026mm 035mm 015mm 020mm, clip=true, width=\ResultsWidth]{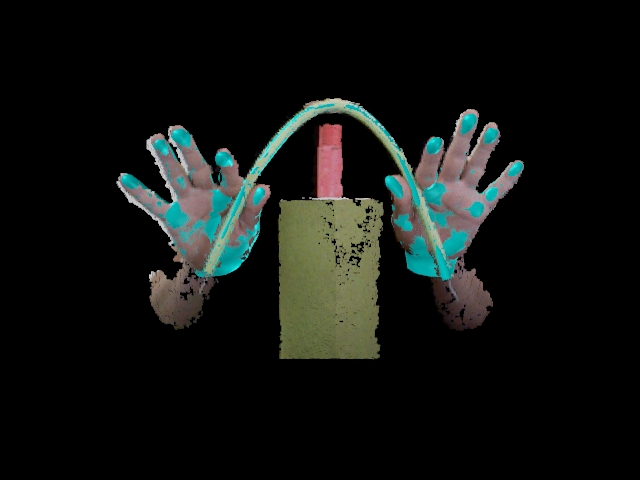}			&	
	\hspace*{-\ResultsSquizBB}	\includegraphics[trim=000mm 020mm 000mm 012mm, clip=true, width=\ResultsWidth]{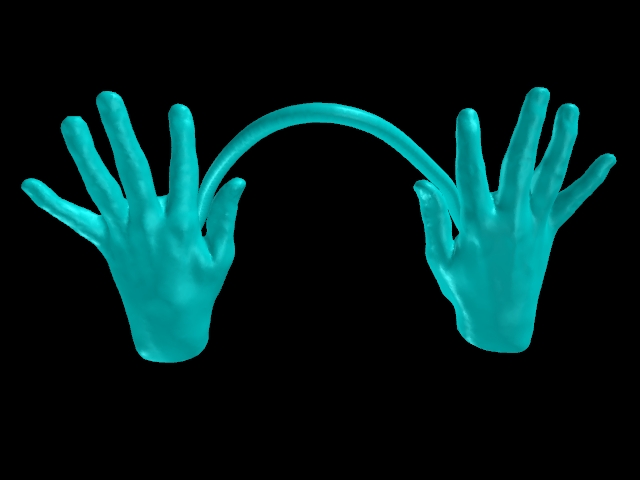}	\\	& \hspace*{+1mm}	(d) ``Bending a Rope'' (new sequence) 						\hspace*{-3\ResultsSquize}	& \vspace*{\ResultsSkip} \\

	\hspace*{+2mm}			\includegraphics[trim=040mm 010mm 015mm 030mm, clip=true, width=\ResultsWidth]{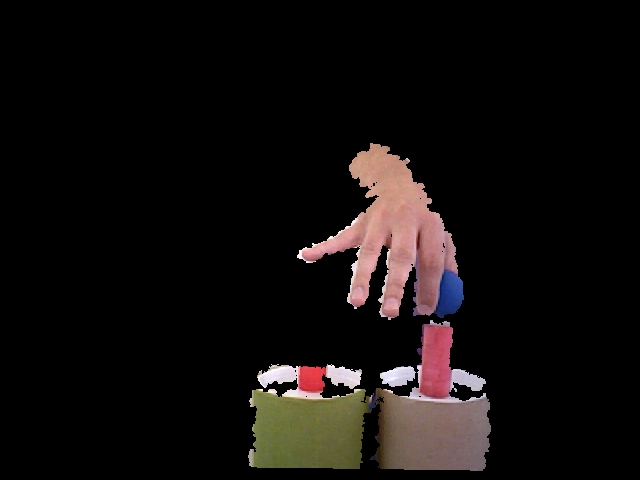}			&	
	\hspace*{-\ResultsSquizAA}	\includegraphics[trim=040mm 010mm 015mm 030mm, clip=true, width=\ResultsWidth]{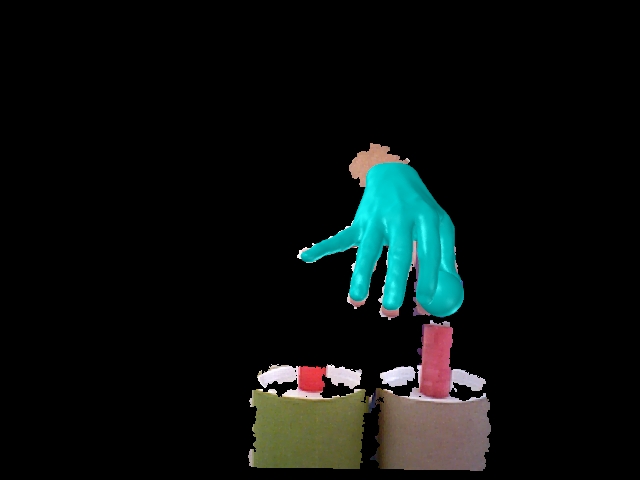}		&	
	\hspace*{-17mm}			\includegraphics[trim=040mm 010mm 015mm 030mm, clip=true, width=\ResultsWidth]{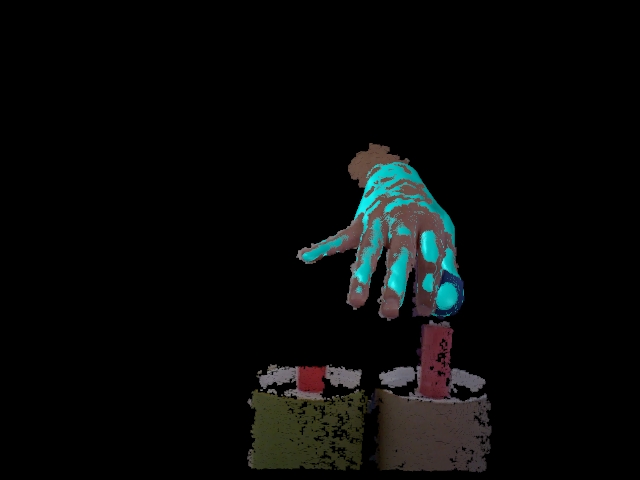}			&	
	\hspace*{-\ResultsSquizBB}	\includegraphics[trim=006mm 004mm 007mm 004mm, clip=true, width=\ResultsWidth]{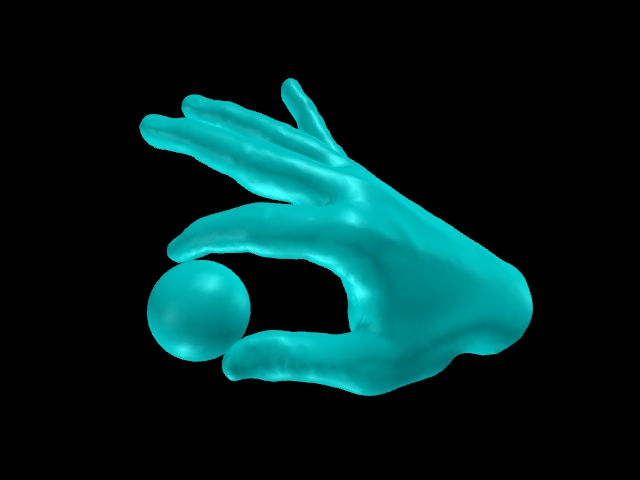}	\\	& \hspace*{+1mm}	(e) ``Moving a Ball'' with occluded manipulating finger (new sequence) 	\hspace*{-3\ResultsSquize}	& \vspace*{\ResultsSkip} \\

	\hspace*{+2mm}			\includegraphics[trim=027mm 020mm 028mm 030mm, clip=true, width=\ResultsWidth]{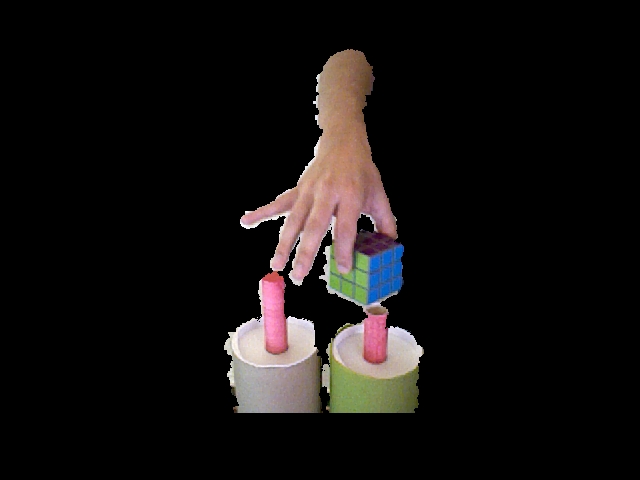}		&	
	\hspace*{-\ResultsSquizAA}	\includegraphics[trim=027mm 020mm 028mm 030mm, clip=true, width=\ResultsWidth]{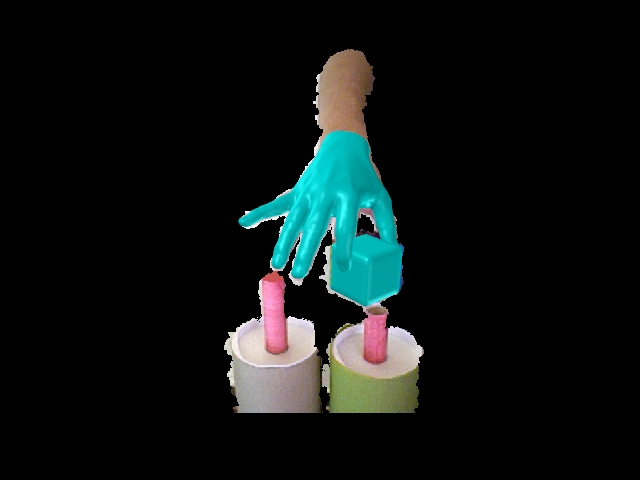}	&	
	\hspace*{-17mm}			\includegraphics[trim=027mm 020mm 028mm 030mm, clip=true, width=\ResultsWidth]{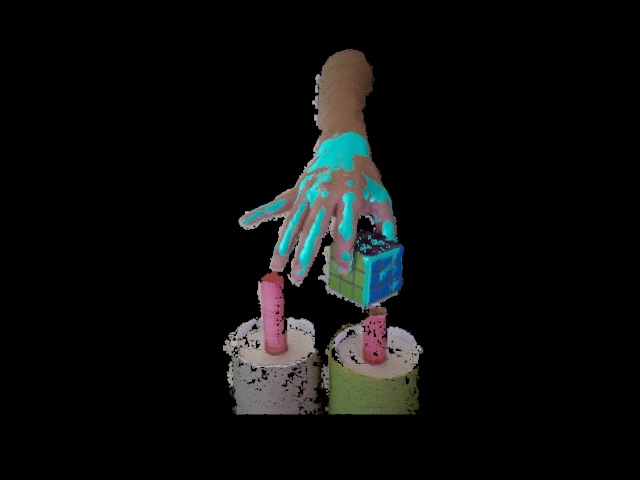}		&	
	\hspace*{-\ResultsSquizBB}	\includegraphics[trim=000mm 008mm 000mm 005mm, clip=true, width=\ResultsWidth]{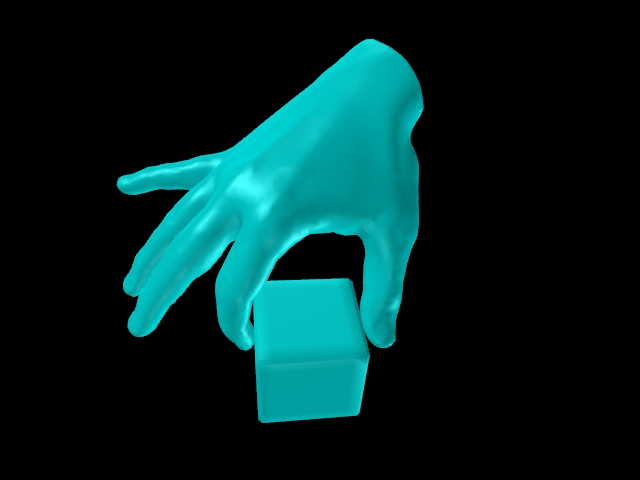}\\	& \hspace*{+1mm}	(f) ``Moving a Cube'' (new sequence) 				 		\hspace*{-3\ResultsSquize}	& \vspace*{\ResultsSkip} \\

	\hspace*{+2mm}			\includegraphics[trim=027mm 020mm 028mm 030mm, clip=true, width=\ResultsWidth]{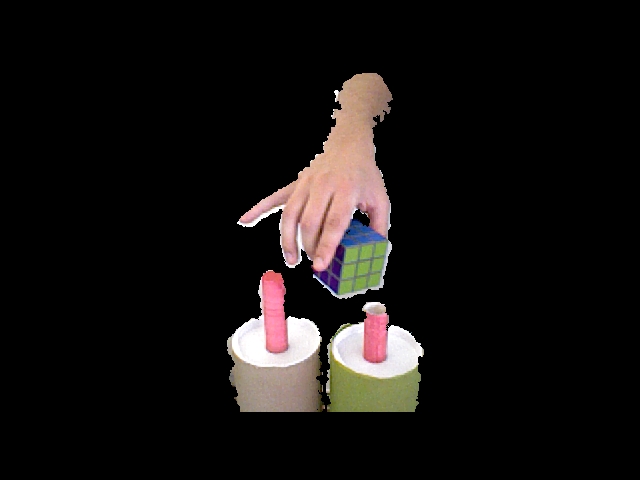}		&	
	\hspace*{-\ResultsSquizAA}	\includegraphics[trim=027mm 020mm 028mm 030mm, clip=true, width=\ResultsWidth]{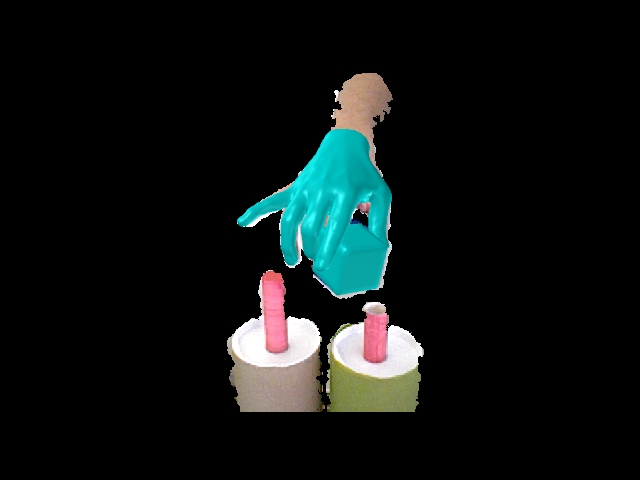}	&	
	\hspace*{-17mm}			\includegraphics[trim=027mm 020mm 028mm 030mm, clip=true, width=\ResultsWidth]{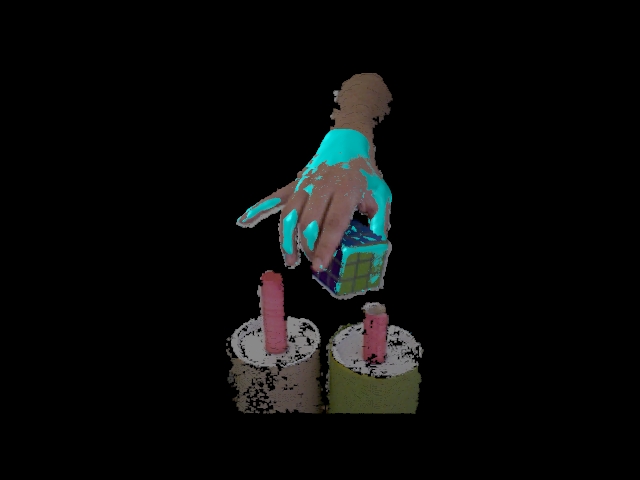}		&	
	\hspace*{-\ResultsSquizBB}	\includegraphics[trim=000mm 008mm 000mm 005mm, clip=true, width=\ResultsWidth]{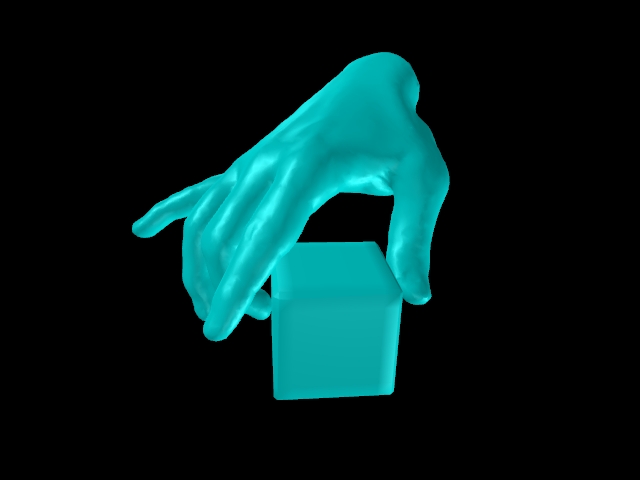}\\	& \hspace*{+1mm}	(g) ``Moving a Cube'' with occluded manipulating finger (new sequence) 	\hspace*{-3\ResultsSquize}	& \vspace*{\ResultsSkip} \\

	\end{tabular}					
	\normalsize
	\caption{	Some of the obtained results. 
			(Left) Input RGB-D image. 
			(Center-Left) Obtained results overlayed on the input image. 
			(Center-Right) Obtained results fitted in the input point cloud. 
			(Right) Obtained results from another viewpoint. 
	}
	\label{fig:My_Results_WithObject}
	\end{center}
	\end{figure*}

	\newcommand{\ResultsSquizeLUCA}{8.5mm}
	
	\begin{figure*}[t]
	\begin{center}
	\tiny
	\begin{tabular}{c c c}
	\hspace*{+\ResultsSquizeLUCA}	\includegraphics[trim=000mm 000mm 000mm 000mm, clip=true, width=\ResultsWidth]{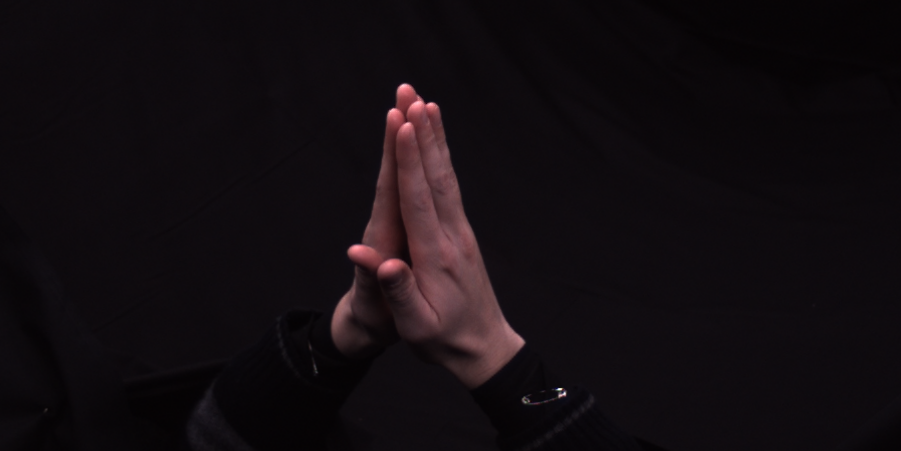}			&	\hspace*{-\ResultsSquizeLUCA}	\includegraphics[trim=000mm 000mm 000mm 000mm, clip=true, width=\ResultsWidth]{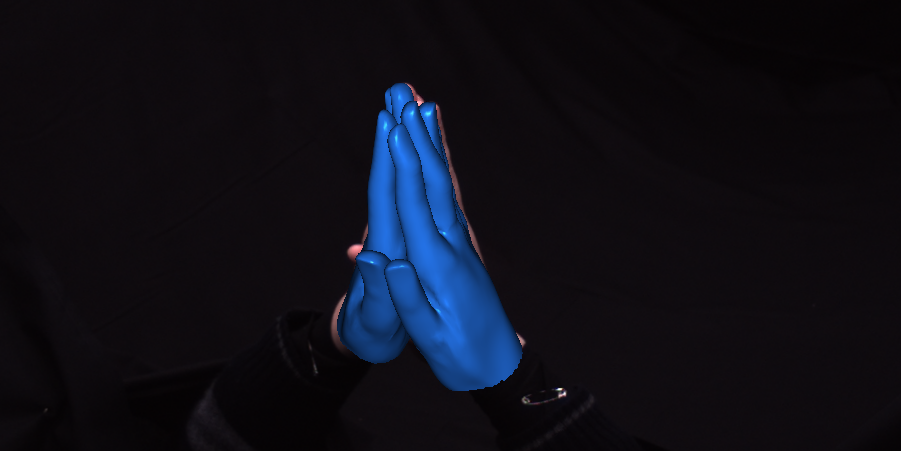}			&	\hspace*{-\ResultsSquizeLUCA}	\includegraphics[trim=000mm 000mm 000mm 000mm, clip=true, width=\ResultsWidth]{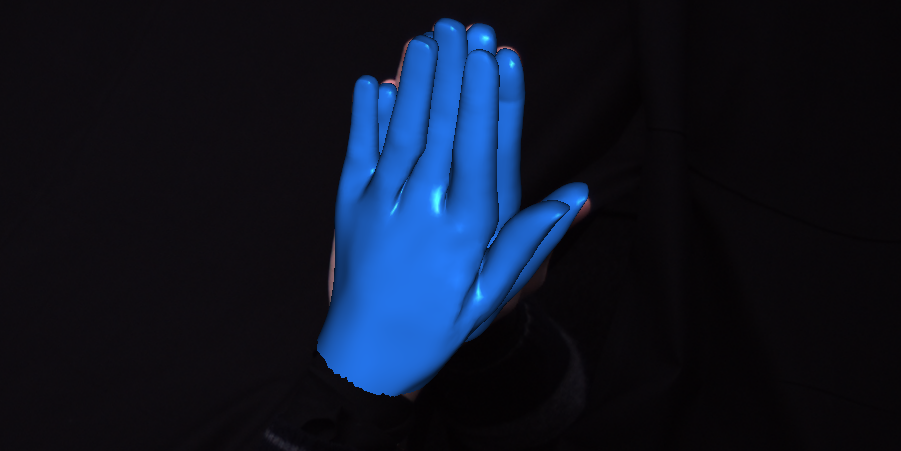}			\\	& (a) ``Praying'' 				& \vspace*{\ResultsSkip} \\
	\hspace*{+\ResultsSquizeLUCA}	\includegraphics[trim=000mm 000mm 000mm 000mm, clip=true, width=\ResultsWidth]{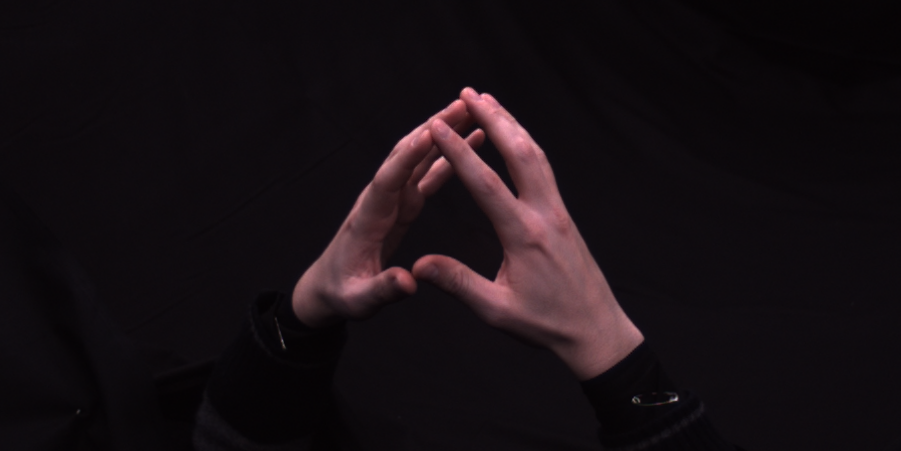}			&	\hspace*{-\ResultsSquizeLUCA}	\includegraphics[trim=000mm 000mm 000mm 000mm, clip=true, width=\ResultsWidth]{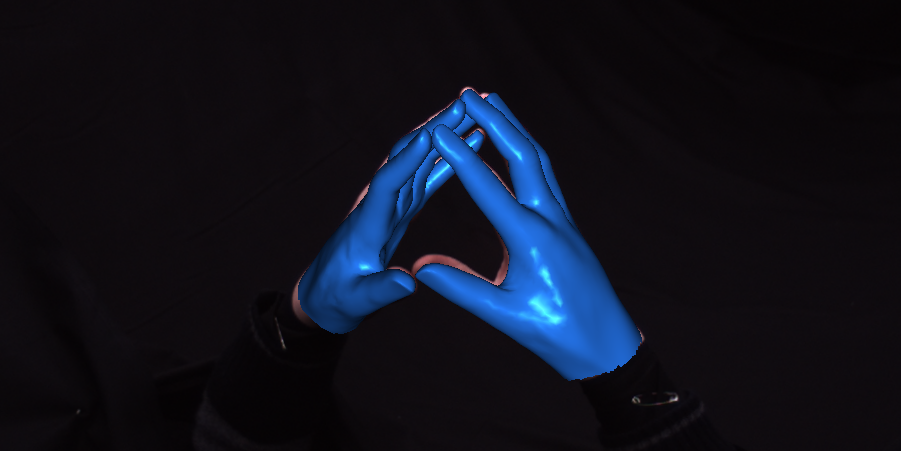}			&	\hspace*{-\ResultsSquizeLUCA}	\includegraphics[trim=000mm 000mm 000mm 000mm, clip=true, width=\ResultsWidth]{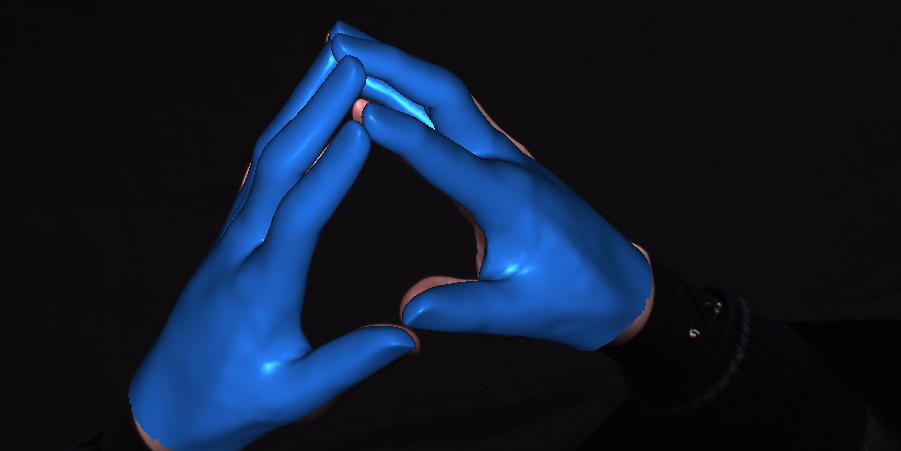}			\\	& (b) ``Finger Tips Touching'' 			& \vspace*{\ResultsSkip} \\
	\hspace*{+\ResultsSquizeLUCA}	\includegraphics[trim=000mm 000mm 000mm 000mm, clip=true, width=\ResultsWidth]{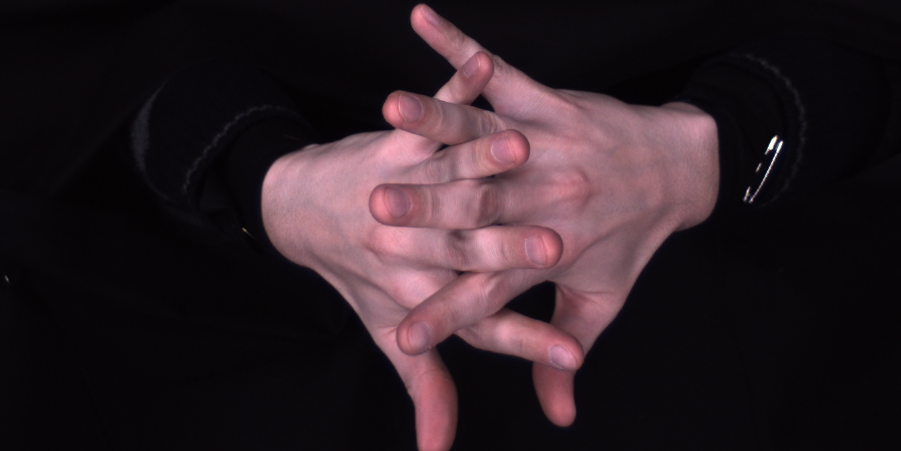}			&	\hspace*{-\ResultsSquizeLUCA}	\includegraphics[trim=000mm 000mm 000mm 000mm, clip=true, width=\ResultsWidth]{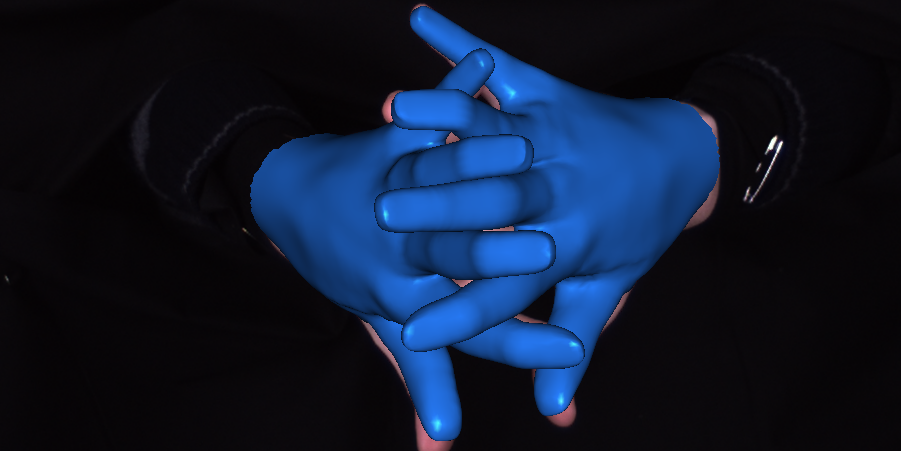}			&	\hspace*{-\ResultsSquizeLUCA}	\includegraphics[trim=000mm 000mm 000mm 000mm, clip=true, width=\ResultsWidth]{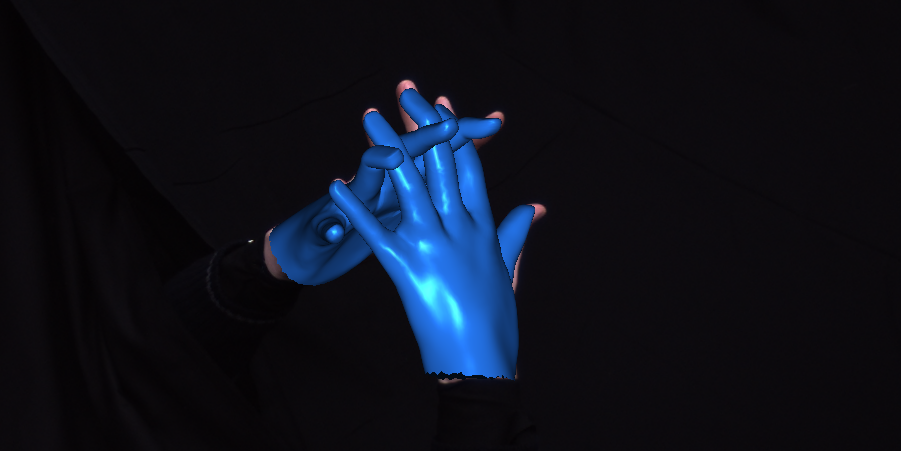}			\\	& (c) ``Fingers Crossing''			& \vspace*{\ResultsSkip} \\
	\hspace*{+\ResultsSquizeLUCA}	\includegraphics[trim=000mm 000mm 000mm 000mm, clip=true, width=\ResultsWidth]{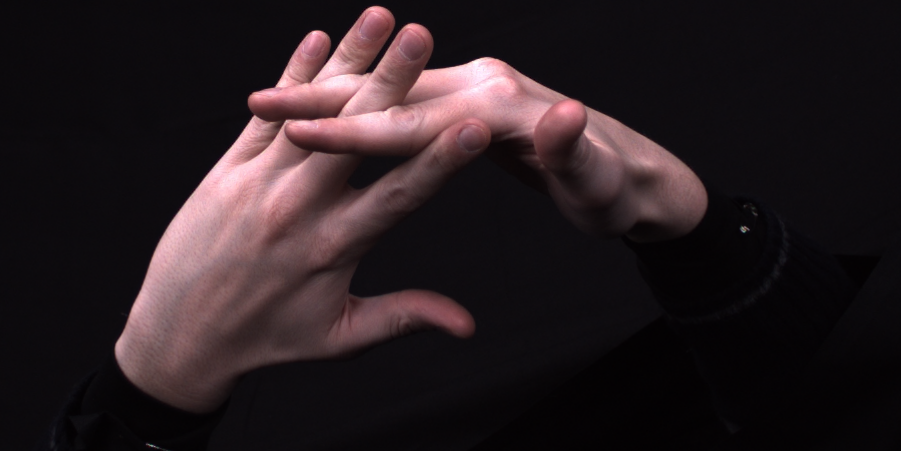}			&	\hspace*{-\ResultsSquizeLUCA}	\includegraphics[trim=000mm 000mm 000mm 000mm, clip=true, width=\ResultsWidth]{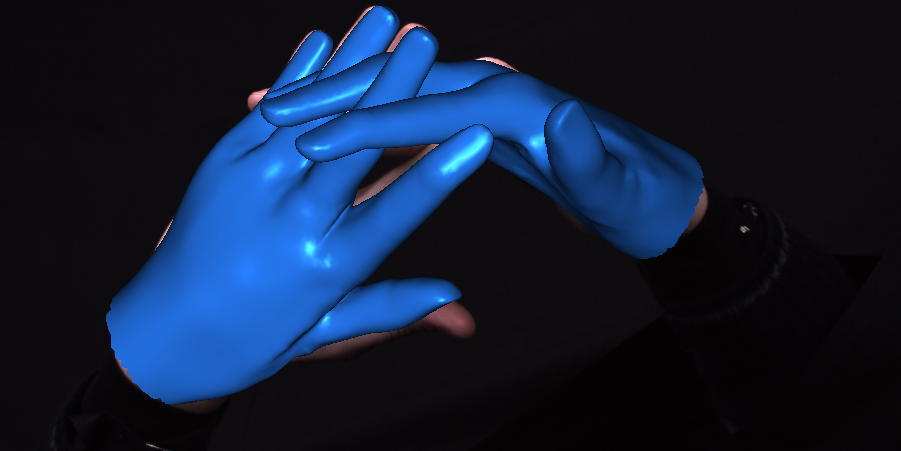}			&	\hspace*{-\ResultsSquizeLUCA}	\includegraphics[trim=000mm 000mm 000mm 000mm, clip=true, width=\ResultsWidth]{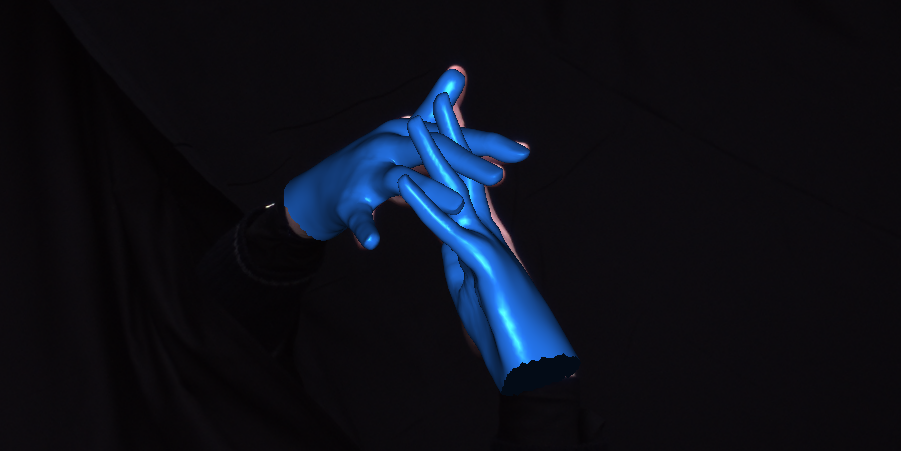}			\\	& (d) ``Fingers Crossing and Twisting'' 	& \vspace*{\ResultsSkip} \\
	\hspace*{+\ResultsSquizeLUCA}	\includegraphics[trim=000mm 000mm 000mm 000mm, clip=true, width=\ResultsWidth]{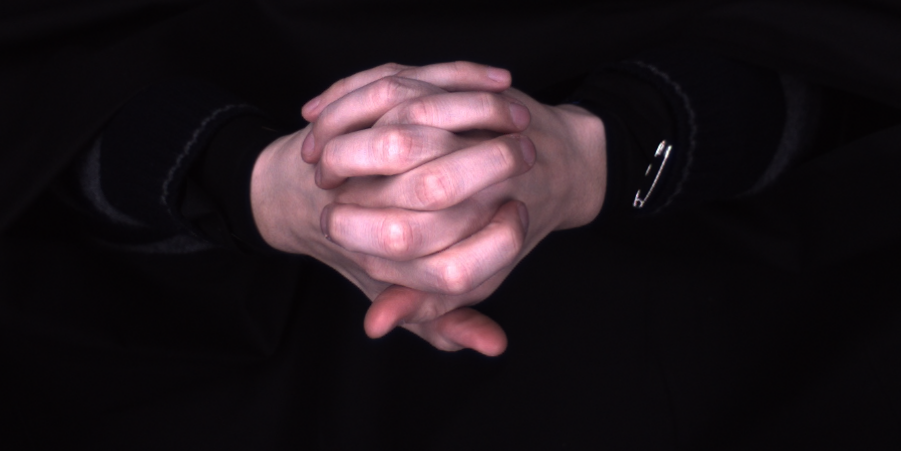}			&	\hspace*{-\ResultsSquizeLUCA}	\includegraphics[trim=000mm 000mm 000mm 000mm, clip=true, width=\ResultsWidth]{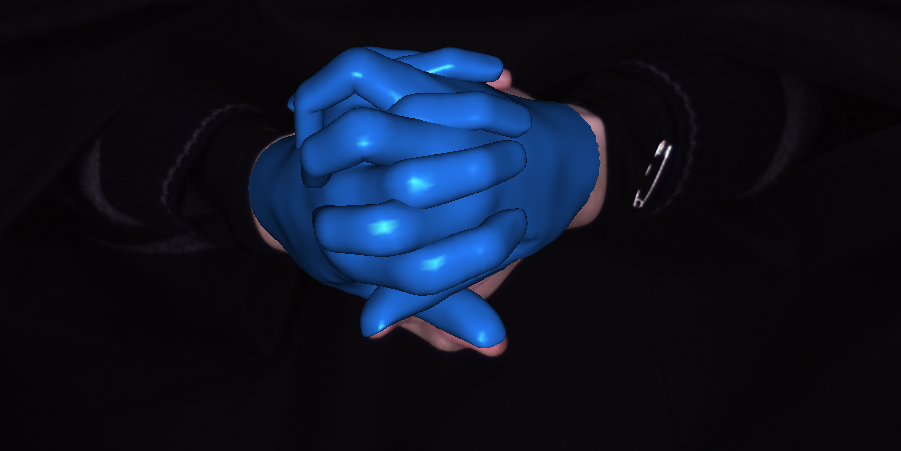}			&	\hspace*{-\ResultsSquizeLUCA}	\includegraphics[trim=000mm 000mm 000mm 000mm, clip=true, width=\ResultsWidth]{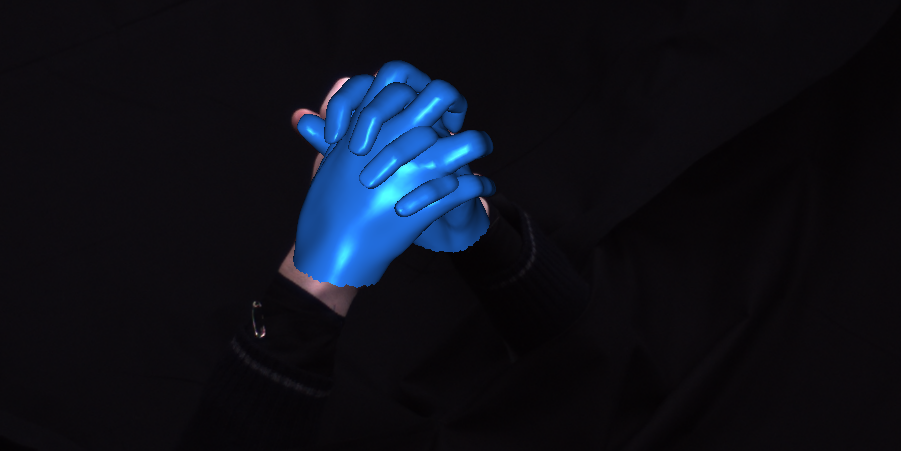}			\\	& (e) ``Fingers Folding'' 			& \vspace*{\ResultsSkip} \\
	\hspace*{+\ResultsSquizeLUCA}	\includegraphics[trim=000mm 000mm 000mm 000mm, clip=true, width=\ResultsWidth]{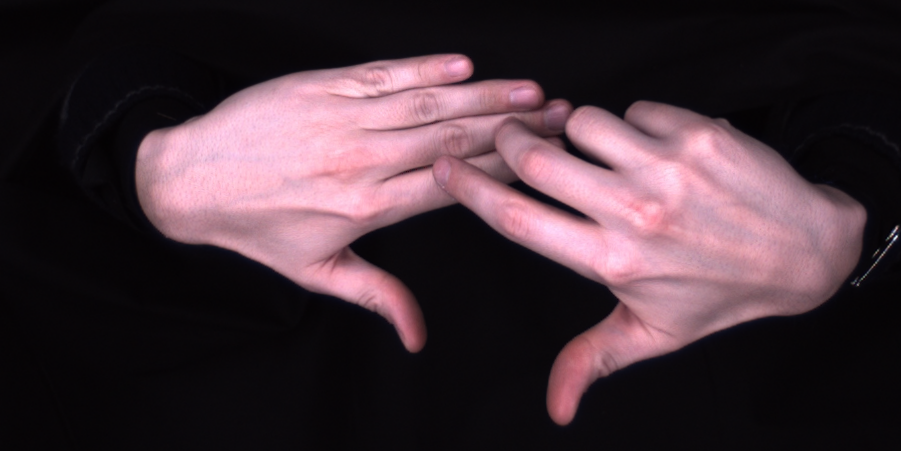}			&	\hspace*{-\ResultsSquizeLUCA}	\includegraphics[trim=000mm 000mm 000mm 000mm, clip=true, width=\ResultsWidth]{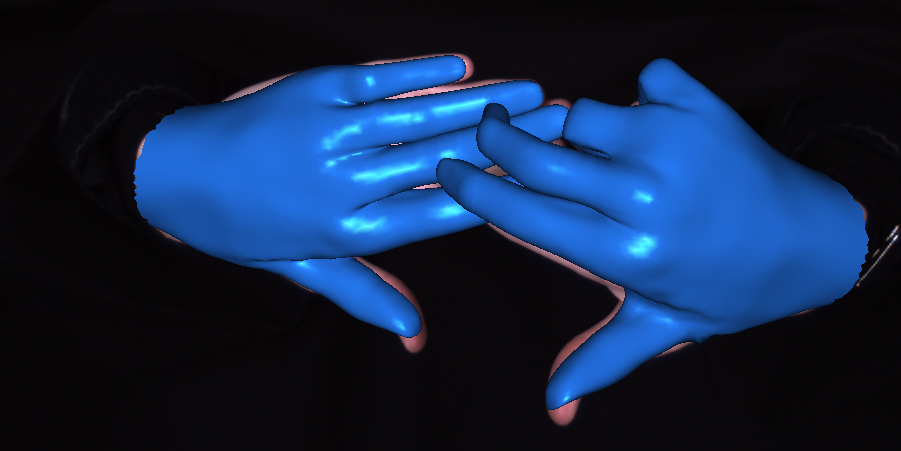}			&	\hspace*{-\ResultsSquizeLUCA}	\includegraphics[trim=000mm 000mm 000mm 000mm, clip=true, width=\ResultsWidth]{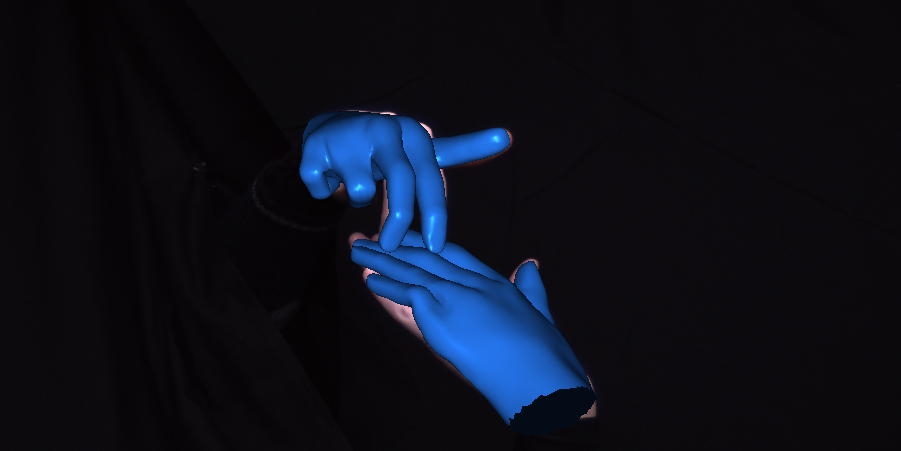}			\\	& (f) ``Fingers Walking''			& \vspace*{\ResultsSkip} \\
	\hspace*{+\ResultsSquizeLUCA}	\includegraphics[trim=000mm 000mm 000mm 000mm, clip=true, width=\ResultsWidth]{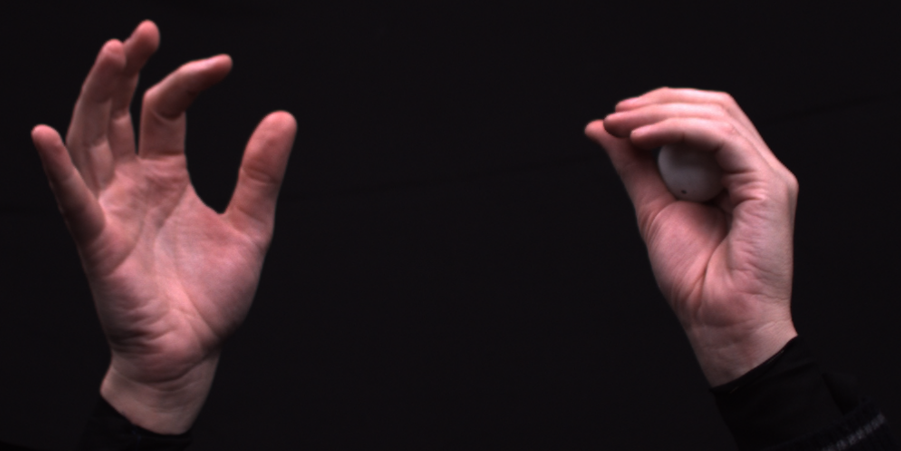}				&	\hspace*{-\ResultsSquizeLUCA}	\includegraphics[trim=000mm 000mm 000mm 000mm, clip=true, width=\ResultsWidth]{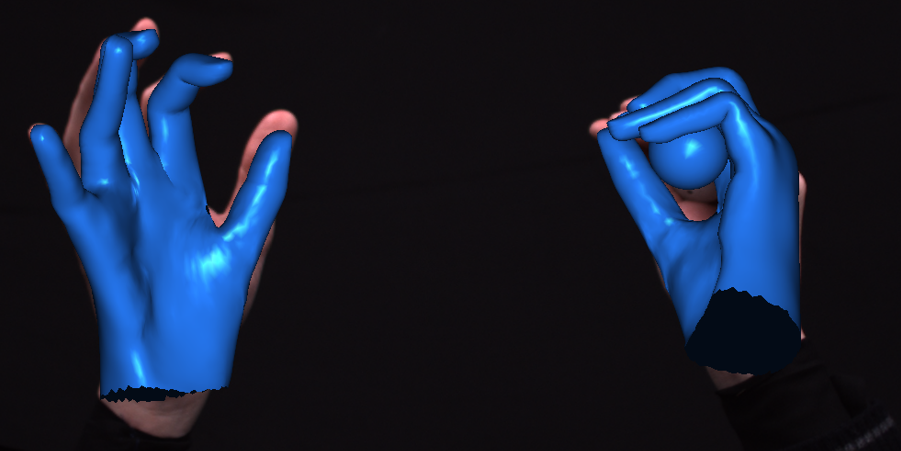}				&	\hspace*{-\ResultsSquizeLUCA}	\includegraphics[trim=000mm 000mm 000mm 000mm, clip=true, width=\ResultsWidth]{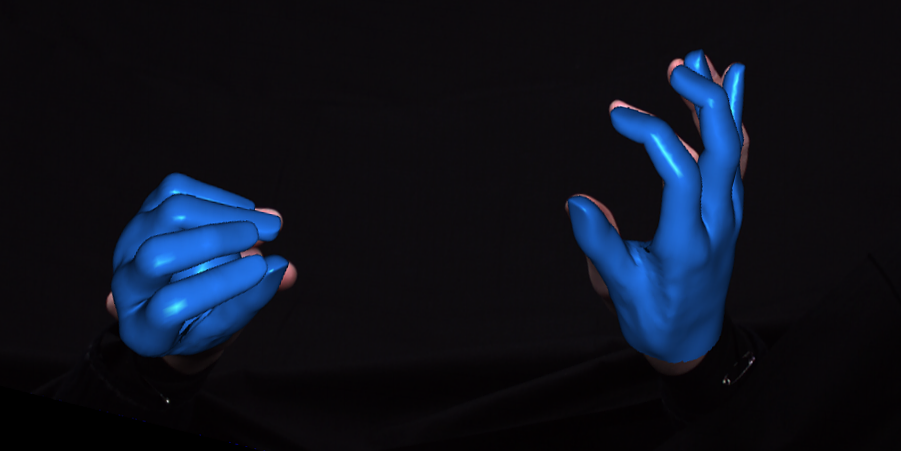}				\\	& (g) ``Holding and Passing a Ball'' 		& \vspace*{\ResultsSkip} \\
	\hspace*{+\ResultsSquizeLUCA}	\includegraphics[trim=190mm 060mm 150mm 105mm, clip=true, width=\ResultsWidth]{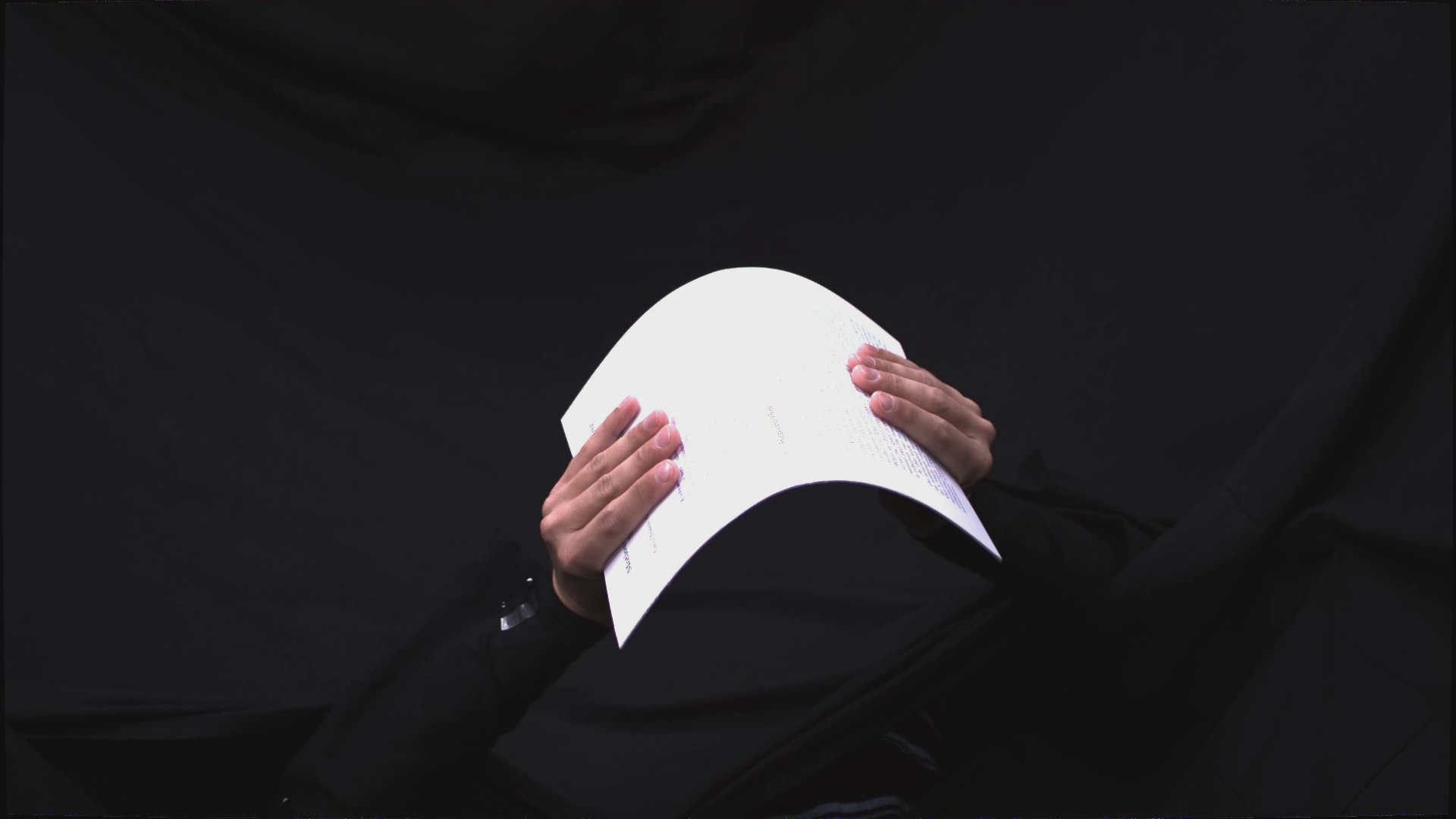}	&	\hspace*{-\ResultsSquizeLUCA}	\includegraphics[trim=190mm 060mm 150mm 105mm, clip=true, width=\ResultsWidth]{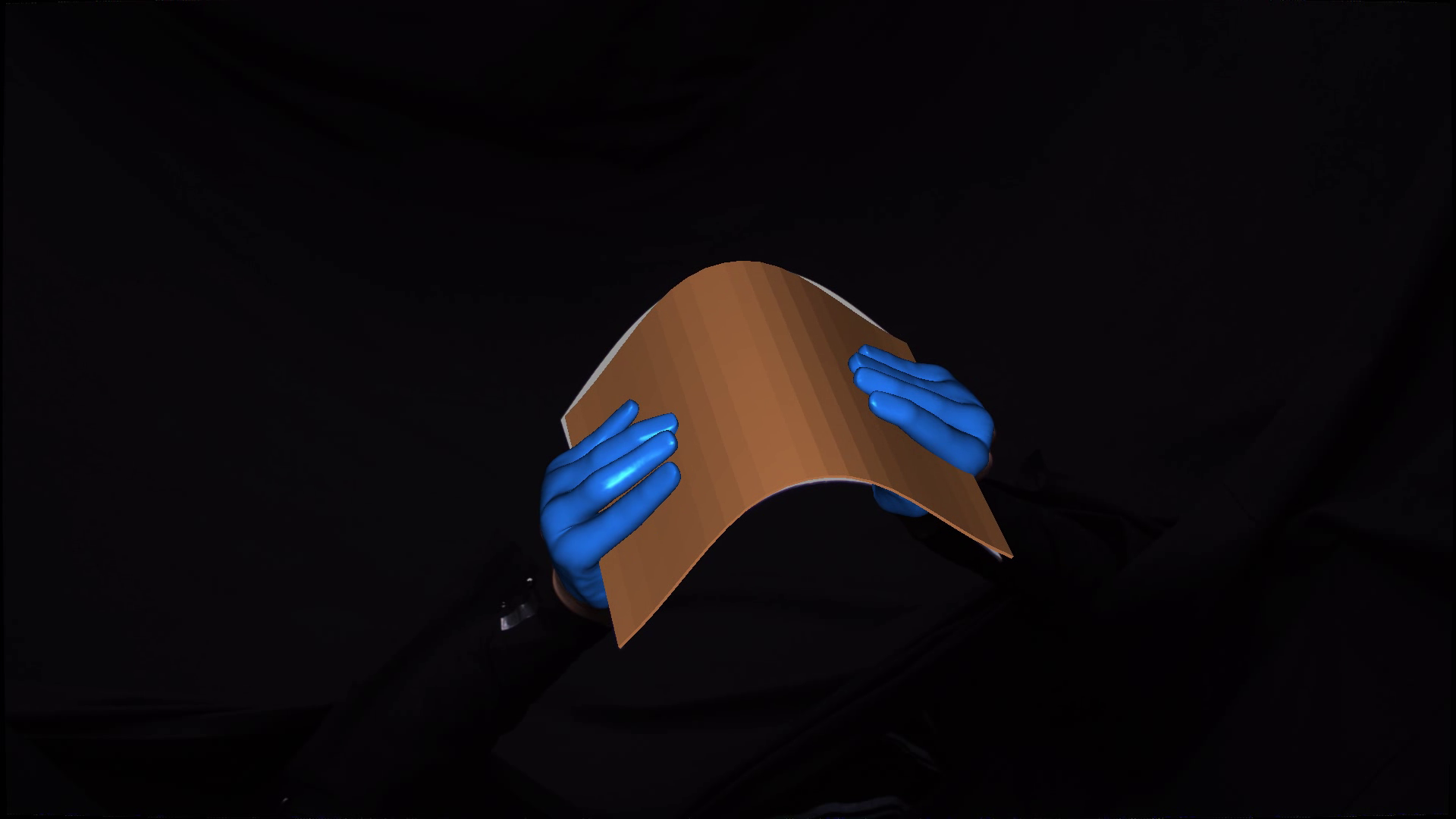}	&	\hspace*{-\ResultsSquizeLUCA}	\includegraphics[trim=190mm 060mm 150mm 105mm, clip=true, width=\ResultsWidth]{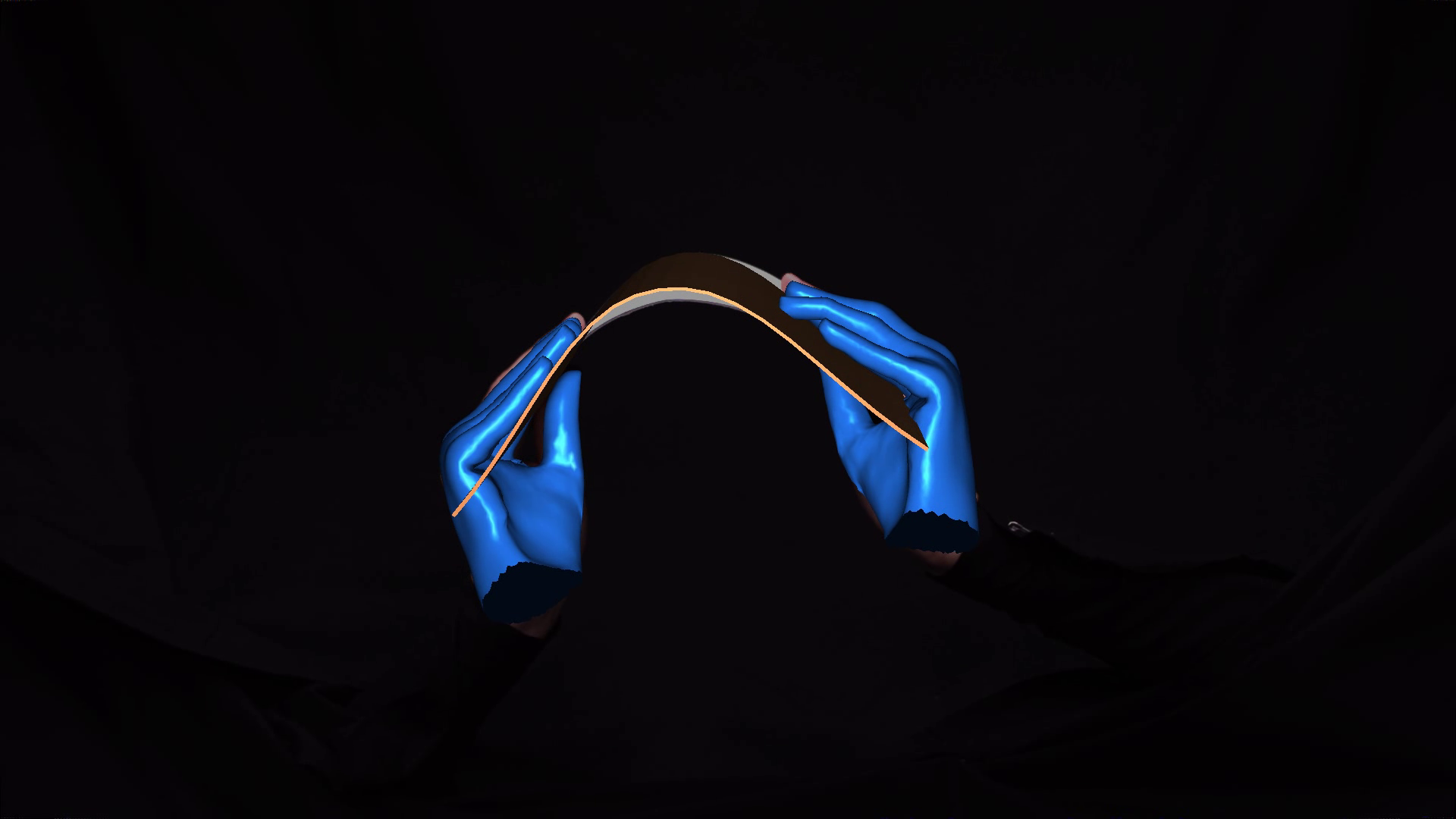}	\\	& (h) ``Paper Folding'' (new sequence)		& \vspace*{\ResultsSkip} \\
	\hspace*{+\ResultsSquizeLUCA}	\includegraphics[trim=120mm 100mm 060mm 005mm, clip=true, width=\ResultsWidth]{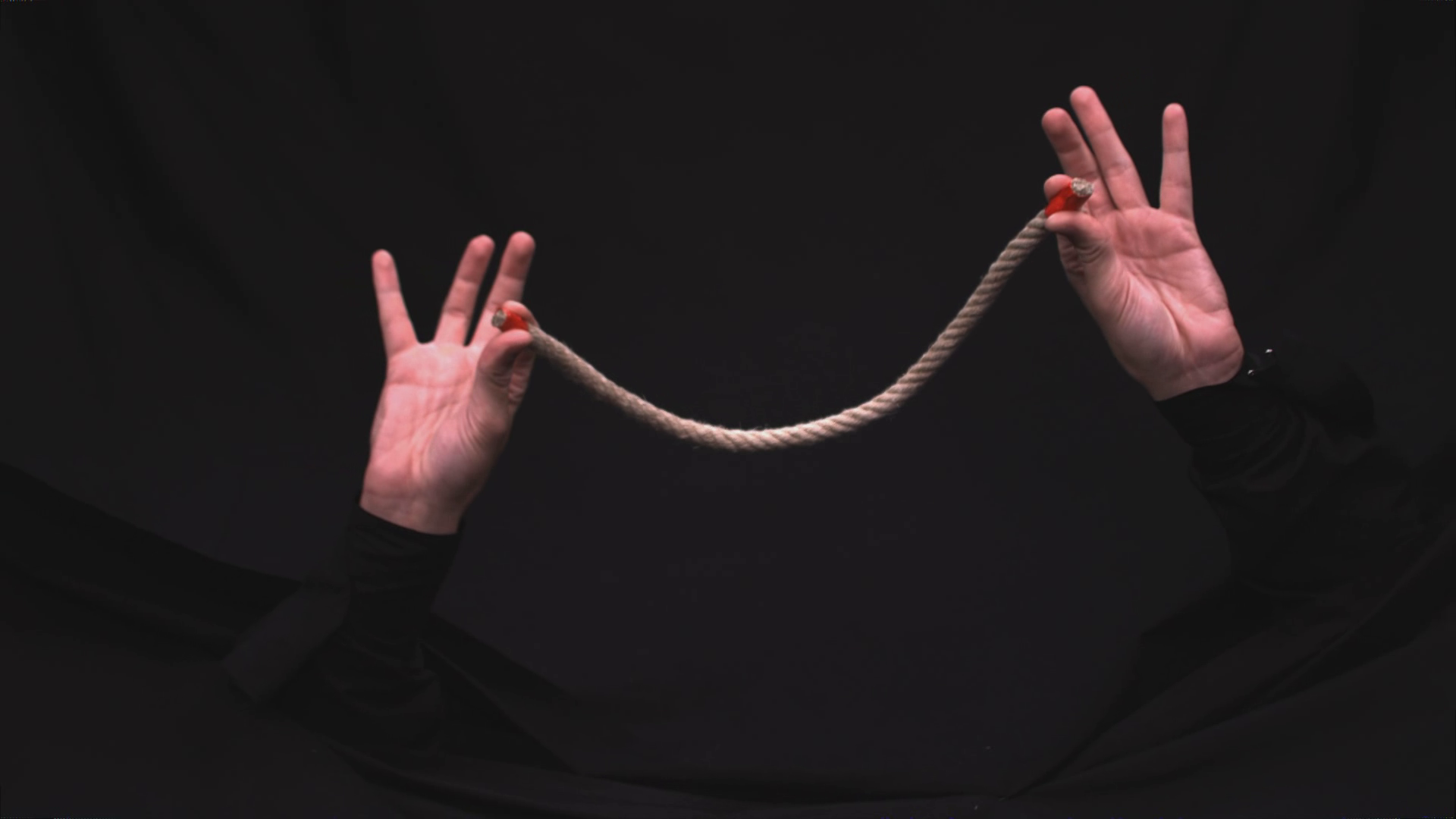}	&	\hspace*{-\ResultsSquizeLUCA}	\includegraphics[trim=120mm 100mm 060mm 005mm, clip=true, width=\ResultsWidth]{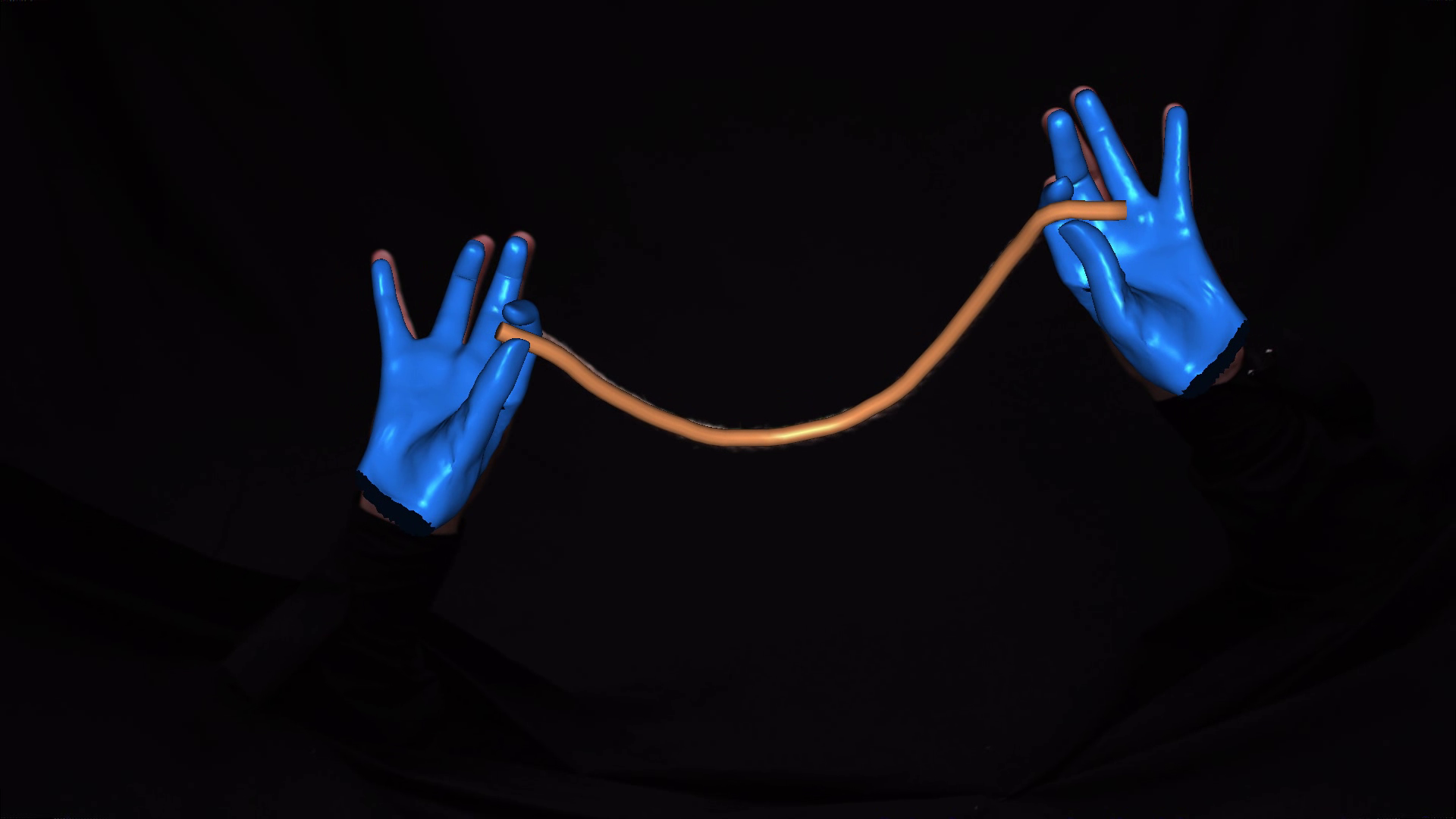}	&	\hspace*{-\ResultsSquizeLUCA}	\includegraphics[trim=100mm 094mm 060mm 000mm, clip=true, width=\ResultsWidth]{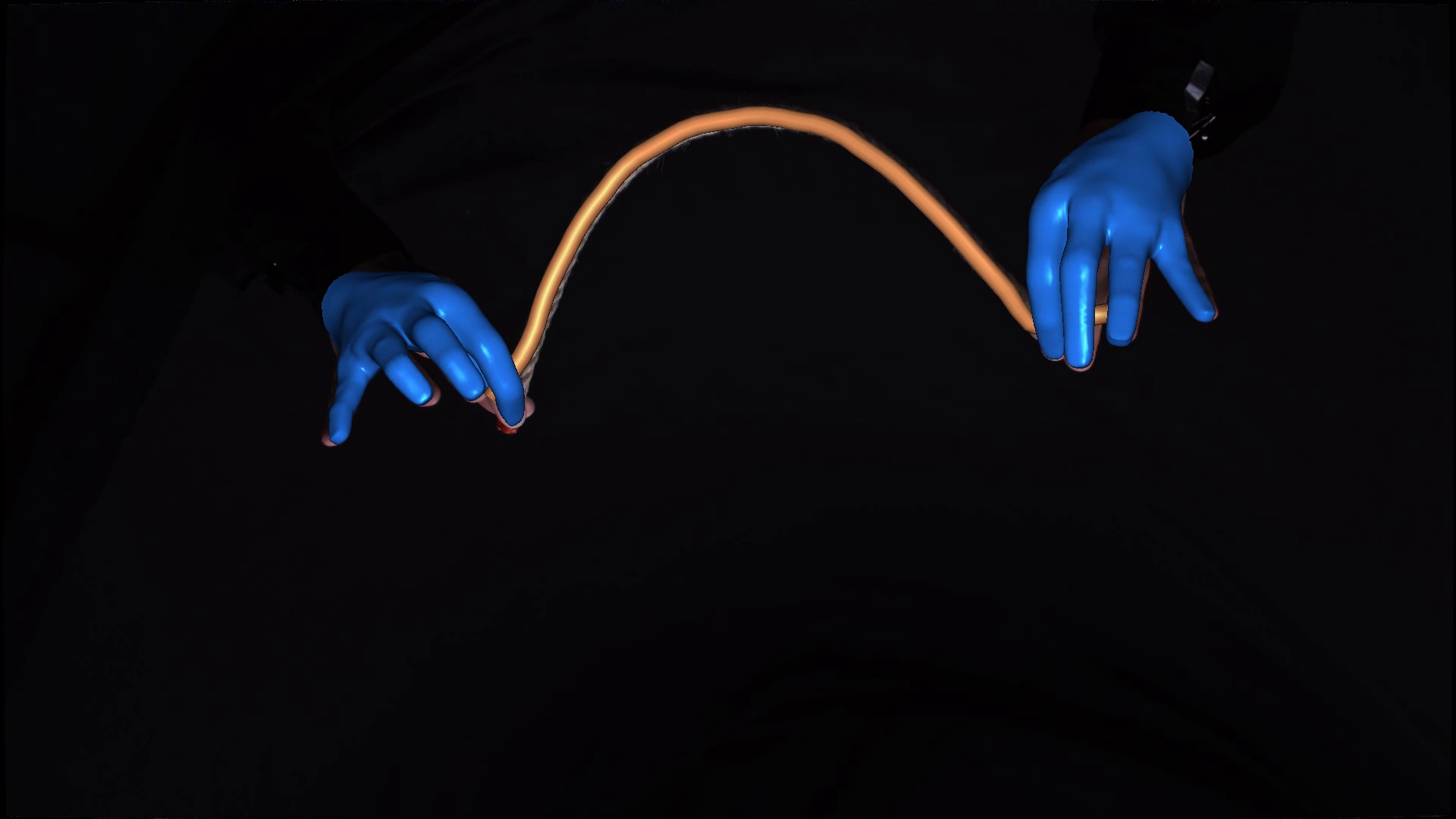}		\\	& (i) ``Rope Folding'' (new sequence)		&

	\end{tabular}					
	\normalsize
	\caption{	Some of the obtained results. 
			(Left) One of the input RGB images. 
			(Center) Obtained results overlayed on the input image. 
			(Right) Obtained results from another viewpoint. 
	}
	\label{fig:Luca_Results}
	\end{center}
	\end{figure*}

\section{Conclusion}\label{sec:conclusion}

	In this paper we have presented a framework that captures articulated motion of hands and manipulated objects from monocular RGB-D videos as well as multiple synchronized RGB videos.
	Contrary to works that focus on gestures and single hands, we focus on the more difficult case of intense hand-hand and hand-object interactions.
	To address the difficulties, we have proposed an approach that combines in a single objective function a generative model with discriminatively trained salient points, collision detection and physics simulation. 
	Although the collision and physics term reduce the pose estimation only slightly, they increase the realism of the captured motion, especially under occlusions and missing visual data.
	We performed qualitative and quantitative evaluations on $8$ sequences captured with multiple RGB cameras and on $21$ sequences captured with a single RGB-D camera. 
	Comparisons with an approach based on particle swarm optimization~\citep{OikonomidisBMVC} for both camera systems revealed the our model achieves a higher accuracy for hand pose estimation.	
	For the first time, we present successful tracking results of hands interacting with highly articulated objects.

\section{Acknowledgements}\label{sec:acknowledgements}

	Financial support was provided by the DFG Emmy Noether program (GA 1927/1-1).

\newpage

\bibliographystyle{spbasic}
\bibliography{IJCV_2014_Submission___FINAL_CLEAN___highlightedChangesOFF___numberingOFF}

\begin{thebibliography}{72}
\providecommand{\natexlab}[1]{#1}
\providecommand{\url}[1]{{#1}}
\providecommand{\urlprefix}{URL }
\expandafter\ifx\csname urlstyle\endcsname\relax
  \providecommand{\doi}[1]{DOI~\discretionary{}{}{}#1}\else
  \providecommand{\doi}{DOI~\discretionary{}{}{}\begingroup
  \urlstyle{rm}\Url}\fi
\providecommand{\eprint}[2][]{\url{#2}}

\bibitem[{Aggarwal et~al(1987)Aggarwal, Klawe, Moran, Shor, and
  Wilber}]{CGAL_distance1}
Aggarwal A, Klawe MM, Moran S, Shor P, Wilber R (1987) Geometric applications
  of a matrix-searching algorithm. Algorithmica 2(1-4):195--208

\bibitem[{Albrecht et~al(2003)Albrecht, Haber, and Seidel}]{AnatomicalModel}
Albrecht I, Haber J, Seidel HP (2003) Construction and animation of
  anatomically based human hand models. In: SCA, pp 98--109

\bibitem[{Athitsos and Sclaroff(2003)}]{AthitsosCluttered2003}
Athitsos V, Sclaroff S (2003) Estimating 3d hand pose from a cluttered image.
  In: CVPR, pp 432--439

\bibitem[{Ballan and Cortelazzo(2008)}]{Luca3DPVT08}
Ballan L, Cortelazzo GM (2008) Marker-less motion capture of skinned models in
  a four camera set-up using optical flow and silhouettes. In: {3DPVT}

\bibitem[{Ballan et~al(2012)Ballan, Taneja, Gall, Van~Gool, and
  Pollefeys}]{LucaHands}
Ballan L, Taneja A, Gall J, Van~Gool L, Pollefeys M (2012) Motion capture of
  hands in action using discriminative salient points. In: ECCV, pp 640--653

\bibitem[{Baran and Popovi\'{c}(2007)}]{pinocchio}
Baran I, Popovi\'{c} J (2007) Automatic rigging and animation of 3d characters.
  TOG 26(3)

\bibitem[{Belongie et~al(2002)Belongie, Malik, and Puzicha}]{Malik_Bipartite}
Belongie S, Malik J, Puzicha J (2002) Shape matching and object recognition
  using shape contexts. PAMI 24(4):509--522

\bibitem[{Bray et~al(2007)Bray, Koller-Meier, and
  Van~Gool}]{vanGool_smartParticle}
Bray M, Koller-Meier E, Van~Gool L (2007) Smart particle filtering for
  high-dimensional tracking. CVIU 106(1):116--129

\bibitem[{Bregler et~al(2004)Bregler, Malik, and Pullen}]{Malik_Twist}
Bregler C, Malik J, Pullen K (2004) Twist based acquisition and tracking of
  animal and human kinematics. IJCV 56(3):179--194

\bibitem[{Brox et~al(2010)Brox, Rosenhahn, Gall, and Cremers}]{Bro10}
Brox T, Rosenhahn B, Gall J, Cremers D (2010) Combined region- and motion-based
  3d tracking of rigid and articulated objects. PAMI 32(3):402--415

\bibitem[{de~Campos and Murray(2006)}]{Murray_handRegression2006}
de~Campos T, Murray D (2006) Regression-based hand pose estimation from
  multiple cameras. In: CVPR

\bibitem[{Canny(1986)}]{canny}
Canny J (1986) A computational approach to edge detection. PAMI 8(6):679--698

\bibitem[{Chen and Medioni(1991)}]{p2pl}
Chen Y, Medioni G (1991) Object modeling by registration of multiple range
  images. In: ICRA

\bibitem[{Coumans(2013)}]{BulletPhysics}
Coumans E (2013) {Bullet} real-time physics simulation.
  \urlprefix\url{http://bulletphysics.org}

\bibitem[{Delamarre and Faugeras(2001)}]{Faugeras_PhysicalForces}
Delamarre Q, Faugeras OD (2001) 3d articulated models and multiview tracking
  with physical forces. CVIU 81(3):328--357

\bibitem[{Ekvall and Kragic(2005)}]{Kragic_dataglove}
Ekvall S, Kragic D (2005) Grasp recognition for programming by demonstration.
  In: ICRA, pp 748--753

\bibitem[{Erol et~al(2007)Erol, Bebis, Nicolescu, Boyle, and
  Twombly}]{Review_Erol_HandPose}
Erol A, Bebis G, Nicolescu M, Boyle RD, Twombly X (2007) Vision-based hand pose
  estimation: A review. CVIU 108(1-2):52--73

\bibitem[{Everingham et~al(2010)Everingham, Van~Gool, Williams, Winn, and
  Zisserman}]{pascalChallengeVOC}
Everingham M, Van~Gool L, Williams C, Winn J, Zisserman A (2010) The pascal
  visual object classes (voc) challenge. IJCV 88(2):303--338

\bibitem[{Felzenszwalb and Huttenlocher(2004)}]{DT_Felzenszwalb}
Felzenszwalb PF, Huttenlocher DP (2004) Distance transforms of sampled
  functions. Tech. rep., Cornell Computing and Information Science

\bibitem[{Gall et~al(2011{\natexlab{a}})Gall, Fossati, and
  Van~Gool}]{functionalCategorization}
Gall J, Fossati A, Van~Gool L (2011{\natexlab{a}}) Functional categorization of
  objects using real-time markerless motion capture. In: CVPR, pp 1969--1976

\bibitem[{Gall et~al(2011{\natexlab{b}})Gall, Yao, Razavi, Van~Gool, and
  Lempitsky}]{juergen_Hough}
Gall J, Yao A, Razavi N, Van~Gool L, Lempitsky V (2011{\natexlab{b}}) Hough
  forests for object detection, tracking, and action recognition. PAMI
  33(11):2188--2202

\bibitem[{G\"{a}rtner and Sch\"{o}nherr(2000)}]{CGAL_distance2}
G\"{a}rtner B, Sch\"{o}nherr S (2000) An efficient, exact, and generic
  quadratic programming solver for geometric optimization. SCG '00, pp 110--118

\bibitem[{Hamer et~al(2009)Hamer, Schindler, Koller-Meier, and
  Van~Gool}]{Hamer_Hand_Manipulating}
Hamer H, Schindler K, Koller-Meier E, Van~Gool L (2009) Tracking a hand
  manipulating an object. In: ICCV, pp 1475--1482

\bibitem[{Hamer et~al(2010)Hamer, Gall, Weise, and {Van
  Gool}}]{Hamer_ObjectPrior}
Hamer H, Gall J, Weise T, {Van Gool} L (2010) An object-dependent hand pose
  prior from sparse training data. In: CVPR, pp 671--678

\bibitem[{Heap and Hogg(1996)}]{HoggHand96}
Heap T, Hogg D (1996) Towards 3d hand tracking using a deformable model. In:
  FG, pp 140--145

\bibitem[{Holzer et~al(2012)Holzer, Rusu, Dixon, Gedikli, and
  Navab}]{normals_integralImages_Holzer}
Holzer S, Rusu R, Dixon M, Gedikli S, Navab N (2012) Adaptive neighborhood
  selection for real-time surface normal estimation from organized point cloud
  data using integral images. In: IROS, pp 2684--2689

\bibitem[{Jones and Rehg(2002)}]{skinnColorGMM}
Jones MJ, Rehg JM (2002) Statistical color models with application to skin
  detection. IJCV 46(1):81--96

\bibitem[{Keskin et~al(2012)Keskin, Kıraç, Kara, and Akarun}]{KeskinECCV12}
Keskin C, Kıraç F, Kara Y, Akarun L (2012) Hand pose estimation and hand
  shape classification using multi-layered randomized decision forests. In:
  ECCV

\bibitem[{Kim et~al(2012)Kim, Hilliges, Izadi, Butler, Chen, Oikonomidis, and
  Olivier}]{Oikonomidis_MSR_Digits}
Kim D, Hilliges O, Izadi S, Butler AD, Chen J, Oikonomidis I, Olivier P (2012)
  Digits: freehand 3d interactions anywhere using a wrist-worn gloveless
  sensor. In: UIST, pp 167--176

\bibitem[{Kyriazis and Argyros(2013)}]{kyriazis2013}
Kyriazis N, Argyros A (2013) Physically plausible 3d scene tracking: The single
  actor hypothesis. In: CVPR, pp 9--16

\bibitem[{Kyriazis and Argyros(2014)}]{kyriazis2014}
Kyriazis N, Argyros A (2014) Scalable 3d tracking of multiple interacting
  objects. In: CVPR

\bibitem[{de~La~Gorce et~al(2011)de~La~Gorce, Fleet, and
  Paragios}]{ParagiosHandMonocular2011}
de~La~Gorce M, Fleet DJ, Paragios N (2011) Model-based 3d hand pose estimation
  from monocular video. PAMI 33(9):1793--1805

\bibitem[{Lewis et~al(2000)Lewis, Cordner, and Fong}]{LBS_PoseSpace}
Lewis JP, Cordner M, Fong N (2000) Pose space deformation: A unified approach
  to shape interpolation and skeleton-driven deformation. In: SIGGRAPH

\bibitem[{Lu et~al(2003)Lu, Metaxas, Samaras, and
  Oliensis}]{metaxasSamarasHand_multipleCues}
Lu S, Metaxas D, Samaras D, Oliensis J (2003) Using multiple cues for hand
  tracking and model refinement. In: CVPR, pp 443--450

\bibitem[{MacCormick and Isard(2000)}]{Isard2000}
MacCormick J, Isard M (2000) {Partitioned sampling, articulated objects, and
  interface-quality hand tracking}. In: ECCV, pp 3--19

\bibitem[{Murray et~al(1994)Murray, Sastry, and Zexiang}]{Murray_MathInRob}
Murray RM, Sastry SS, Zexiang L (1994) A Mathematical Introduction to Robotic
  Manipulation

\bibitem[{Oikonomidis et~al(2011{\natexlab{a}})Oikonomidis, Kyriazis, and
  Argyros}]{OikonomidisBMVC}
Oikonomidis I, Kyriazis N, Argyros A (2011{\natexlab{a}}) Efficient model-based
  3d tracking of hand articulations using kinect. In: BMVC, pp 101.1--101.11

\bibitem[{Oikonomidis et~al(2011{\natexlab{b}})Oikonomidis, Kyriazis, and
  Argyros}]{Oikonomidis_1hand_object}
Oikonomidis I, Kyriazis N, Argyros A (2011{\natexlab{b}}) Full dof tracking of
  a hand interacting with an object by modeling occlusions and physical
  constraints. In: ICCV

\bibitem[{Oikonomidis et~al(2012)Oikonomidis, Kyriazis, and
  Argyros}]{Oikonomidis_2hands}
Oikonomidis I, Kyriazis N, Argyros AA (2012) Tracking the articulated motion of
  two strongly interacting hands. In: CVPR, pp 1862--1869

\bibitem[{Oikonomidis et~al(2014)Oikonomidis, Lourakis, and
  Argyros}]{Oikonomidis14}
Oikonomidis I, Lourakis MI, Argyros AA (2014) Evolutionary quasi-random search
  for hand articulations tracking. In: CVPR

\bibitem[{Paris and Durand(2009)}]{bilateralFAST}
Paris S, Durand F (2009) A fast approximation of the bilateral filter using a
  signal processing approach. IJCV 81(1):24--52

\bibitem[{Pons-Moll and Rosenhahn(2011)}]{PonsModelBased}
Pons-Moll G, Rosenhahn B (2011) Model-Based Pose Estimation, pp 139--170

\bibitem[{Qian et~al(2014)Qian, Sun, Wei, Tang, and
  Sun}]{MSR_ASIA_handTracking}
Qian C, Sun X, Wei Y, Tang X, Sun J (2014) Realtime and robust hand tracking
  from depth. In: CVPR

\bibitem[{Rehg and Kanade(1995)}]{Kanade95}
Rehg J, Kanade T (1995) Model-based tracking of self-occluding articulated
  objects. In: ICCV, pp 612--617

\bibitem[{Rehg and Kanade(1994)}]{Kanade94}
Rehg JM, Kanade T (1994) {Visual tracking of high dof articulated structures:
  an application to human hand tracking}. In: ECCV, pp 35--46

\bibitem[{Romero et~al(2009)Romero, Kjellstr\"om, and Kragic}]{Romero09}
Romero J, Kjellstr\"om H, Kragic D (2009) Monocular real-time 3d articulated
  hand pose estimation. In: HUMANOIDS, pp 87--92

\bibitem[{Romero et~al(2010)Romero, Kjellstr\"om, and
  Kragic}]{JavierHandsInAction}
Romero J, Kjellstr\"om H, Kragic D (2010) Hands in action: real-time 3d
  reconstruction of hands in interaction with objects. In: ICRA, pp 458--463

\bibitem[{Rosales et~al(2001)Rosales, Athitsos, Sigal, and
  Sclaroff}]{Athitsos_HandSpecializedMappings2001}
Rosales R, Athitsos V, Sigal L, Sclaroff S (2001) 3d hand pose reconstruction
  using specialized mappings. In: ICCV, pp 378--387

\bibitem[{Rosenhahn et~al(2007)Rosenhahn, Brox, and
  Weickert}]{Rosenh_Plucker_IJCV}
Rosenhahn B, Brox T, Weickert J (2007) Three-dimensional shape knowledge for
  joint image segmentation and pose tracking. IJCV 73(3):243--262

\bibitem[{Rusinkiewicz and Levoy(2001)}]{ICP_EfficientVariants}
Rusinkiewicz S, Levoy M (2001) Efficient variants of the icp algorithm. In:
  3DIM, pp 145--152

\bibitem[{Rusinkiewicz et~al(2002)Rusinkiewicz, Hall-Holt, and
  Levoy}]{RusinkiewiczRealTimeINHAND}
Rusinkiewicz S, Hall-Holt O, Levoy M (2002) Real-time 3d model acquisition. TOG
  21(3):438--446

\bibitem[{Schmidt et~al(2014)Schmidt, Newcombe, and Fox}]{DART-Schmidt-RSS-14}
Schmidt T, Newcombe R, Fox D (2014) Dart: Dense articulated real-time tracking.
  In: Proceedings of Robotics: Science and Systems, Berkeley, USA

\bibitem[{Sharp et~al(2015)Sharp, Keskin, Robertson, Taylor, Shotton, Kim,
  Rhemann, Leichter, Vinnikov, Wei, Freedman, Kohli, Krupka, Fitzgibbon, and
  Izadi}]{sharp_chi2015}
Sharp T, Keskin C, Robertson D, Taylor J, Shotton J, Kim D, Rhemann C, Leichter
  I, Vinnikov A, Wei Y, Freedman D, Kohli P, Krupka E, Fitzgibbon A, Izadi S
  (2015) Accurate, robust, and flexible real-time hand tracking. In: CHI

\bibitem[{Shotton et~al(2011)Shotton, Fitzgibbon, Cook, Sharp, Finocchio,
  Moore, Kipman, and Blake}]{kinect_Paper}
Shotton J, Fitzgibbon A, Cook M, Sharp T, Finocchio M, Moore R, Kipman A, Blake
  A (2011) Real-time human pose recognition in parts from single depth images.
  In: CVPR, pp 1297--1304

\bibitem[{Sridhar et~al(2013)Sridhar, Oulasvirta, and
  Theobalt}]{srinath_iccv2013}
Sridhar S, Oulasvirta A, Theobalt C (2013) Interactive markerless articulated
  hand motion tracking using rgb and depth data. In: ICCV, pp 2456--2463

\bibitem[{Sridhar et~al(2014)Sridhar, Rhodin, Seidel, Oulasvirta, and
  Theobalt}]{srinath_3dv2014}
Sridhar S, Rhodin H, Seidel HP, Oulasvirta A, Theobalt C (2014) Real-time hand
  tracking using a sum of anisotropic gaussians model. In: 3DV

\bibitem[{Sridhar et~al(2015)Sridhar, Mueller, Oulasvirta, and
  Theobalt}]{srinath_cvpr2015}
Sridhar S, Mueller F, Oulasvirta A, Theobalt C (2015) Fast and robust hand
  tracking using detection-guided optimization. In: CVPR

\bibitem[{Stenger et~al(2001)Stenger, Mendonca, and
  Cipolla}]{Cipolla_ModelBased}
Stenger B, Mendonca P, Cipolla R (2001) {Model-based 3D tracking of an
  articulated hand}. In: CVPR

\bibitem[{Stolfi(1991)}]{Stolfi91}
Stolfi J (1991) Oriented Projective Geometry: A Framework for Geometric
  Computation. Boston: Academic Press

\bibitem[{Sudderth et~al(2004)Sudderth, Mandel, Freeman, and
  Willsky}]{SudderthNonparamBeliefPropag}
Sudderth E, Mandel M, Freeman W, Willsky A (2004) {Visual Hand Tracking Using
  Nonparametric Belief Propagation}. In: Workshop on Generative Model Based
  Vision, pp 189--189

\bibitem[{Tang et~al(2013)Tang, Yu, and
  Kim}]{TKKIM_ICCV13_Real_time_Articulated_Hand}
Tang D, Yu TH, Kim TK (2013) Real-time articulated hand pose estimation using
  semi-supervised transductive regression forests. In: ICCV, pp 3224--3231

\bibitem[{Tang et~al(2014)Tang, Chang, Tejani, and
  Kim}]{TKKIM_CVPR14_LatentRegressionForest}
Tang D, Chang HJ, Tejani A, Kim TK (2014) Latent regression forest: Structured
  estimation of 3d articulated hand posture. In: CVPR

\bibitem[{Taylor et~al(2014)Taylor, Stebbing, Ramakrishna, Keskin, Shotton,
  Izadi, Hertzmann, and Fitzgibbon}]{MSR_handShapeAdaptation}
Taylor J, Stebbing R, Ramakrishna V, Keskin C, Shotton J, Izadi S, Hertzmann A,
  Fitzgibbon A (2014) User-specific hand modeling from monocular depth
  sequences. CVPR

\bibitem[{Teschner et~al(2004)Teschner, Kimmerle, Heidelberger, Zachmann,
  Raghupathi, Fuhrmann, Cani, Faure, Magnetat-Thalmann, and
  Strasser}]{collisionDeformableObjects}
Teschner M, Kimmerle S, Heidelberger B, Zachmann G, Raghupathi L, Fuhrmann A,
  Cani MP, Faure F, Magnetat-Thalmann N, Strasser W (2004) Collision detection
  for deformable objects. In: Eurographics

\bibitem[{Thayananthan et~al(2003)Thayananthan, Stenger, Torr, and
  Cipolla}]{Cipolla_TORR}
Thayananthan A, Stenger B, Torr PHS, Cipolla R (2003) Shape context and chamfer
  matching in cluttered scenes. In: CVPR, pp 127--133

\bibitem[{Tompson et~al(2014)Tompson, Stein, Lecun, and
  Perlin}]{NYU_tracker_tompson14tog}
Tompson J, Stein M, Lecun Y, Perlin K (2014) Real-time continuous pose recovery
  of human hands using convolutional networks. TOG 33

\bibitem[{Tzionas and Gall(2013)}]{GCPR_2013_Tzionas_Gall}
Tzionas D, Gall J (2013) A comparison of directional distances for hand pose
  estimation. In: GCPR

\bibitem[{Tzionas et~al(2014)Tzionas, Srikantha, Aponte, and
  Gall}]{GCPR_2014_Tzionas_Gall}
Tzionas D, Srikantha A, Aponte P, Gall J (2014) Capturing hand motion with an
  rgb-d sensor, fusing a generative model with salient points. In: GCPR

\bibitem[{Vaezi and Nekouie(2011)}]{colorGlove}
Vaezi M, Nekouie MA (2011) 3d human hand posture reconstruction using a single
  2d image. IJHCI 1(4):83--94

\bibitem[{Wang and Popovi\'{c}(2009)}]{glove_MIT}
Wang RY, Popovi\'{c} J (2009) Real-time hand-tracking with a color glove. TOG
  28(3):63:1--63:8

\bibitem[{Wu et~al(2001)Wu, Lin, and
  Huang}]{Huang_capturingNaturalHandArtic2001}
Wu Y, Lin J, Huang T (2001) Capturing natural hand articulation. In: ICCV, pp
  426--432

\bibitem[{Ye et~al(2013)Ye, Zhang, Wang, Zhu, Yang, and Gall}]{survey13}
Ye M, Zhang Q, Wang L, Zhu J, Yang R, Gall J (2013) A survey on human motion
  analysis from depth data. In: Time-of-Flight and Depth Imaging. Sensors,
  Algorithms, and Applications, pp 149--187

\end{thebibliography}

\end{document}